\documentclass[a4paper,fleqn]{cas-sc}
\usepackage[numbers]{natbib}
\usepackage{algorithm}
\usepackage{algorithmic}
\usepackage{amsmath, amsthm, amsfonts}
\usepackage{booktabs}
\usepackage{array}
\usepackage{float}
\usepackage{stfloats}
\usepackage{multirow}
\usepackage{graphicx}
\usepackage{chemfig}
\usepackage{ulem}
\usepackage{url}
\usepackage{pifont}
\usepackage{longtable}
\usepackage{geometry}
\usepackage{makecell}
\usepackage{tablefootnote}
\usepackage[caption=false,font=footnotesize,labelfont=sf,textfont=sf]{subfig}

\def\tsc#1{\csdef{#1}{\textsc{\lowercase{#1}}\xspace}}
\tsc{WGM}
\tsc{QE}
\tsc{EP}
\tsc{PMS}
\tsc{BEC}
\tsc{DE}

\begin{document}
\let\WriteBookmarks\relax
\def\floatpagepagefraction{1}
\def\textpagefraction{.001}
\shorttitle{Orientation-Aware Sparse Tensor PCA for Efficient Unsupervised Feature Selection}
\shortauthors{Junjing Zheng et~al.}

\title [mode = title]{Orientation-Aware Sparse Tensor PCA for Efficient Unsupervised Feature Selection}                      
\tnotemark[1,2]

\tnotetext[1]{This document is the results of the research
   project funded by the National Key Research and Development Program of China No. 2021YFB3100800, the
National Science Foundation of China under Grants 61025006, 60872134, 62376283 and
61901482 and the China Postdoctoral Science Foundation under Grant
2018M633667.}


\author[1]{Junjing Zheng}[style=chinese,type=editor]

\ead{zjj20212035@163.com}

\credit{Conceptualization, Methodology, Formal analysis, Investigation, Software, Validation, Writing – original draft, Data curation.}

\affiliation[1]{organization={College of Electronic Science and Technology, National University of Defense Technology},
                city={Changsha},
                postcode={410073}, 
                state={Hunan},
                country={China}}

\author[1]{Xinyu Zhang}[style=chinese]

\ead{zhangxinyu90111@163.com}
\credit{Conceptualization, Formal analysis, Supervision, Writing – review \& editing.}

\author[1]{Weidong Jiang}[style=chinese]
\cormark[1]
\ead{jwd2232@vip.163.com}

\credit{Conceptualization, Methodology, Resources, Supervision, Writing – review \& editing.}

\author[1]{Xiangfeng Qiu}[style=chinese]
\ead{qxf1981993100@163.com}

\credit{Resources, Supervision, Writing – review \& editing.}

\author[1]{Mingjian Ren}[style=chinese]
\ead{renmingjian97@163.com}
\credit{Software, Data curation, Supervision, Writing – review \& editing.}


\cortext[cor1]{Corresponding author}


\begin{abstract}
Recently, introducing Tensor Decomposition (TD) techniques into unsupervised feature selection (UFS) has been an emerging research topic. A tensor structure is beneficial for mining the relations between different modes and helps relieve the computation burden. However, while existing methods exploit TD to preserve the data tensor structure, they do not consider the influence of data orientation and thus have difficulty in handling orientation-specific data such as time series. To solve the above problem, we utilize the orientation-dependent tensor-tensor product from Tensor Singular Value Decomposition based on $\star_{\mathbf{M}}$-product (T-SVDM) and extend the one-dimensional Sparse Principal Component Analysis (SPCA) to a tensor form. The proposed sparse tensor PCA model can constrain sparsity at the specified mode and yield sparse tensor principal components, which enhances flexibility and accuracy in learning feature relations. To ensure fast convergence and a flexible description of feature correlation, we develop a convex version specially designed for general UFS tasks and propose an efficient slice-by-slice algorithm that performs dual optimization in the transform domain. Experimental results on real-world datasets demonstrate the effectiveness and remarkable computational efficiency of the proposed method for tensor data of diverse structures over the state-of-the-art. When transform axes align with feature distribution patterns, our method is promising for various applications. The codes related to our proposed method and the experiments are available at \url{https://github.com/zjj20212035/STPCA.git}.


\end{abstract}


\maketitle

\section{Introduction}

Feature selection (FS) aims to select a minimal subset of features (variables) sufficient for a given task \cite{A_book_about_FS, Pattern}. Typically employed as a pre-processing step, FS reduces computational complexity and enhances interpretability by providing only essential features.
FS methods are commonly categorized based on the level of supervision during training: supervised \cite{supervised-FS}, semi-supervised \cite{Semi-supervised-FS}, and unsupervised \cite{unsupervised-FS}. Among these, Unsupervised Feature Selection (UFS) presents a greater challenge due to the absence of labels. To address this, most UFS methods leverage learning techniques, including but not limited to principal component analysis (PCA) \cite{SPCAFS}, spectral clustering \cite{SOGFS}, and graph-based learning \cite{CPUFS_2023}. This paper specifically focuses on PCA-based UFS methods.

A common limitation in many UFS approaches is the vectorization of data samples, which can lead to the curse of dimensionality \cite{DynamicProgramming} and loss of inherent data structure, particularly when handling tensors. Tensors, as natural multi-way extensions of matrices for representing data, are prevalent in diverse fields like image processing \cite{HOSVD_Dynamic_Texture}, signal processing \cite{HOSVD_MIMO_Coarrays_AE}, and econometric analysis \cite{HOSVD_Time_Series_Eco}. An $n$-th order tensor consists of $n$ modes, each with multiple dimensions, inherently encoding rich relational information between these modes and dimensions that is valuable for feature analysis. This paper focuses on exploiting this tensor structure to reduce computational complexity without compromising UFS performance.
 performance.

To leverage structural information, tensor-based UFS methods employ various Tensor Decomposition (TD) techniques \cite{CPUFS_2023, liang_multi-view_TRPCA_graph_2023, UFS_Tucker_vehicle_detection_2019}.
Tensor decomposition factorizes a tensor into a product of several subtensors. This area has been extensively studied since the seminal Tucker decomposition \cite{Tucker} was introduced in 1964. The establishment of different tensor rank definitions has led to the development of diverse TD methods, such as Higher-Order Singular Value Decomposition (HOSVD) \cite{HOSVD}, CANDECOMP/PARAFAC decomposition (CPD) \cite{CP1, CP2}, and Tensor Singular Value Decomposition (T-SVD) \cite{T-SVD}.

A key purpose of TD is to uncover the relationships between modes and dimensions, which are intrinsically linked to UFS. TD methods preserve the tensor structure crucial for UFS, while UFS provides a framework to validate the accuracy and interpretability of the relationships learned through TD. By minimizing the reconstruction error between the original tensor and its decomposed representation, tensor-based UFS methods effectively preserve data structure information. However, most existing methods do not inherently impose sparsity constraints within the TD process itself. Consequently, they often rely on a separate classifier matrix to perform feature selection, which can obscure the interpretability and limit the direct utilization of the learned element relationships.
To address this limitation, we propose seeking compact representations of the factor matrices. It is well-established that TD methods can also be interpreted as achieving Tensor Principal Component Analysis (TPCA) \cite{HOSVD, TensorFaces}. Principal components capture maximal data variance, and their directions indicate the significance of features during projection onto the underlying manifold. Therefore, the key to selecting interpretable and discriminative features likely lies in TPCA.

Recent studies in domains such as electroencephalogram (EEG) analysis \cite{egg_review} underscore the necessity of orientation-aware approaches. \cite{egg_example} demonstrates that spatial-only feature representations, which neglect time-frequency couplings, fail to capture discriminative features in multi-channel EEG signals with temporal variability, resulting in 15\%-26\% silhouette score degradation for feature selection and clustering tasks. Consequently, the authors implement feature selection separately across time, frequency, and time-frequency domains, significantly enhancing clustering accuracy.
In broader applications, TPCA methods can be categorized into three primary types: Tucker-based TPCA \cite{HOSVD, MPCA, Multiview_PCA}, CPD-based TPCA, and T-SVDM-based TPCA \cite{Facial_recognition_T-SVD, 2DTPCA, pose_estimation_TPCA_t_svd_2011}. Key distinctions are summarized in Table \ref{tab:decomp_comparison}.
While Tucker-based and CPD-based methods model potential mode correlations, their reliance on vector outer product principles renders them inherently orientation-insensitive \cite{T-SVD}. This limitation manifests in orientation-specific data (e.g., temporal sequences), where Tucker's and CPD's decomposition results remain invariant to data orientation. In contrast, T-SVDM-based approaches leverage matrix outer products to adapt to orientation and transform domain variations, yielding direction-sensitive outcomes with broader applicability.
Furthermore, T-SVDM’s slice-by-slice processing (Section \ref{Tensor Operations}) enables slice-wise sparsity, allowing independent optimization of factor matrices. This offers greater flexibility for algorithmic design compared to the rank-constrained CPD and Tucker decomposition that produces dense core tensors. \cite{T-SVDM_T-SVDM2} further highlights T-SVDM’s reduced parameter sensitivity versus Tucker and CPD in image denoising.
Although T-SVDM-based methods incur higher computational complexity from factor tensor operations, they provide granular element-wise weighting that meticulously characterizes feature relationships. These attributes confer unique advantages for complex scenarios.
Motivated by these strengths, we propose integrating T-SVDM-based TPCA into UFS. Achieving this requires extending the framework with sparsity constraints to preserve only essential feature relations. To our knowledge, such a framework remains absent in the literature.

\begin{table}[!ht]
\centering
\caption{Comparative analysis of UFS methods based on different TD approaches.}
\label{tab:decomp_comparison}
\begin{tabular}{p{3.7cm}p{3.8cm}p{3.5cm}p{3.8cm}}
\toprule
\textbf{Property} & \textbf{Tucker-based} & \textbf{CPD-based} & \textbf{T-SVDM-based} \\
\midrule

\textbf{Basic product} & vector outer product & vector outer product & matrix outer product \\

\textbf{Computational Complexity} & 
$O(t(d_1d_2k_1k_2 + d_1d_3k_1k_3+d_2d_3k_2k_3))^{\ast}$ & 
$O(t(d_1d_2+d_1d_3+d_2d_3)k^2)$ & 
$O\big(d_3^2d_1d_2 +{}$\newline
$d_3\min(d_1d_2^2,d_1^2d_2)\big)$ \\

\textbf{Sensitivity to Orientation} & 
Isotropic & 
Permutation-invariant & 
Directionally aware \\

\textbf{Sparsity Handling} & 
Dense core & 
Rank-constrained & 
Slice-wise sparsity \\

\textbf{Domain Adaptability} & 
Original domain & 
Original domain & 
Transform domain \\

\bottomrule
\end{tabular}
\vspace{0.2cm}
\parbox{\textwidth}{
\footnotesize{
$^{\ast}$\textit{$t$: iterations. 
$d_i$: mode-$i$ dimensionality (original tensor). 
$k_i$: mode-$i$ dimensionality (core tensor) ($i=1,2,3$).}
}}
\end{table}

This paper proposes a novel sparse T-SVDM-based TPCA framework with a convex formulation for Unsupervised Feature Selection. We establish a methodology that leverages tensor structures to reduce computational complexity while maintaining competitive UFS performance. Our contributions are summarized as follows:
\begin{enumerate}[\textbullet]
    \item We put forward a novel Sparse Tensor PCA model based on $\star_{\mathbf{M}}$-Product (STPCA-MP) that extends the principle of one-dimensional Sparse PCA (SPCA) to a tensor form by applying a tensor decomposition technique called T-SVDM. By considering the orientation of the data tensor, the proposed model can better utilize the data organization and learn the feature correlations. 
    
    \item We deduce a convex version of STPCA-MP specially designed for unsupervised feature selection tasks by merging the encoder tensor and decoder tensor into the unified sparse reconstruction factor tensor. Our method not only preserves the global information by minimizing the reconstruction error but also demonstrates consistency between TD and UFS by utilizing the reconstruction tensor to perform UFS, fully leveraging learned latent feature relations and gaining more interpretability. We prove that the global optimal solution falls onto the Hermitian Positive Semidefinite Cone. 

    \item We design an optimization algorithm that exploits the transform domain to solve a dual problem in a slice-by-slice manner. By applying the HPSD projector, we can efficiently solve each subproblem and end up with a score map for feature selection. The convergence of the proposed algorithm is proven theoretically and practically. Compared to non-tensor-based methods, our proposed method reaches a lower computational complexity. 

    \item Real-world datasets are used to validate the effectiveness and efficiency of the proposed method, especially towards orientation-specific data like time series. 
\end{enumerate}

\begin{figure*}
    \centering
    \includegraphics[width=1\linewidth]{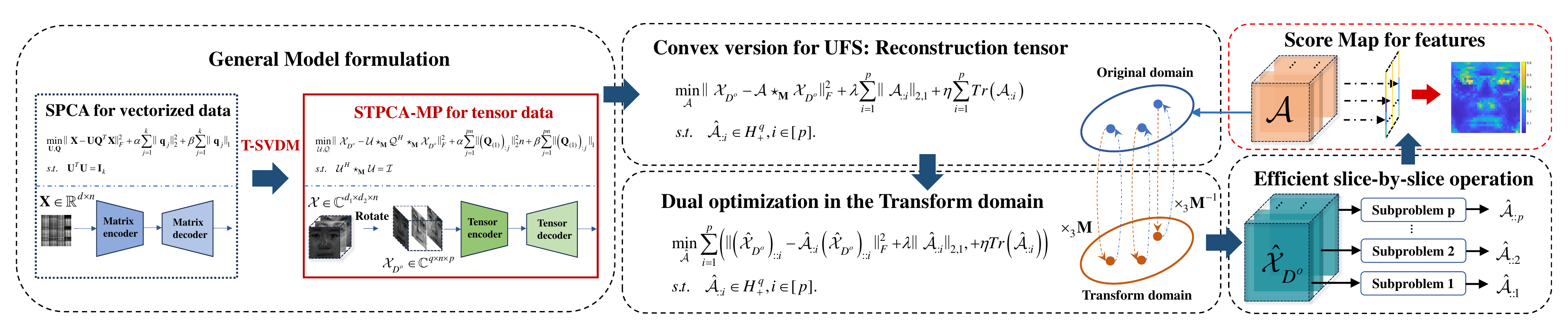}
    \caption{Overall framework of our work}
    \label{Overall framework of our work}
\end{figure*}

\section{Related Work}
This section introduces mathematical notations employed throughout the paper and reviews relevant research on tensor decomposition, tensor principal component analysis, and unsupervised feature selection.

\subsection{Notations}
\subsubsection{Basic Symbols}
Notations employed throughout the paper are summarized in Table \ref{tab:notations}. Specifically, fibers correspond to vector sections along individual tensor modes, while slices represent matrix sections defined by fixing all but two modes. 

\begin{table}[h]
\centering
\caption{Summary of mathematical notations. The following notations use the letter A as an example, and they are equally applicable to other letters.}
\label{tab:notations}
\begin{tabular}{@{}cll}
\toprule
\textbf{Symbol} & \textbf{Description} \\
\midrule
$A$ & Generic set &\\
$\mathbb{Z}, \mathbb{R}, \mathbb{C}$ & Integer, real, complex number sets \\
$H^d_{+}$ & Hermitian Positive Semidefinite (HPSD) cone ($d \times d$ matrices) \\
$\Phi$ & Empty set \\
$[n]$ & Ordered set $\{1, 2, \dots, n\}$ \\
$A^o$ & Ordered set (element order matters) \\
$\#(A)$ & Cardinality of set $A$ \\
$a$ & Scalar \\
$\mathbf{a}$ & Vector \\
$\mathbf{A}$ & Matrix \\
$\mathcal{A}$ & Third-order Tensor \\
$\mathcal{A}_{: i_2 i_3}$, $\mathcal{A}_{i_1 : i_3}$, $\mathcal{A}_{i_1 i_2 :}$ & Column (mode-1), Row (mode-2), and Tube (mode-3) fibers \\
$\mathcal{A}_{::i_3}$, $\mathcal{A}_{:i_2:}$, $\mathcal{A}_{i_1::}$ & Frontal, Lateral and Horizontal slices\\
$\mathcal{A}_{i_1\cdots i_n}$ & Tensor element at $(i_1,\dots,i_n)$ \\

$a_{i j}$ & Matrix element at $(i,j)$ \\
$\mathbf{A}_{j,:}, \mathbf{a}^j$ & $j$-th row vector of matrix \\
$\mathbf{A}_{:,j}, \mathbf{a}_j$ & $j$-th column vector of matrix \\
$\mathbf{A}^T, \mathbf{A}^*, \mathbf{A}^H$ & Transpose, conjugate, conjugate transpose \\
$\mathrm{Tr}(\mathbf{A})$ & Matrix trace \\
$\bigotimes$ & Kronecker product \\
$\|\mathcal{A}\|_F$ & Frobenius norm ($\sqrt{\sum_{i_1}\sum_{i_2}\sum_{i_3}\mathcal{A}^2_{i_1i_2i_3}}$) \\
$\|\mathbf{A}\|_*$ & Nuclear norm ($\sum \sigma_i^{\ast}$) \\
$\|\mathbf{A}\|_{2,1}$ & $\ell_{2,1}$ norm ($\sum_j \|\mathbf{A}_{:,j}\|_2$) \\
\bottomrule
\end{tabular}
\vspace{0.2cm}
\parbox{\textwidth}{
\footnotesize{
$^{\ast}$\textit{singular values}
}}
\end{table}

\subsubsection{Tensor Operations}
\label{Tensor Operations}




In this section, we introduce fundamental tensor operations.

\vspace{0.05cm}
\textbf{a) Tensor Unfolding and Tensor-Matrix Product} \label{Vectorization and Tensor unfolding}
\vspace{0.05cm}

Tensor unfolding (matricization) is performed via $n$-mode unfolding. For an $m$-th order tensor $\mathcal{X} \in \mathbb{R}^{d_1 \times d_2 \times \cdots \times d_m}$, its $n$-mode unfolding $\mathbf{X}_{(n)}$ is defined such that:
\begin{equation}
    \begin{aligned}
    \mathcal{X}(i_1,i_2,\dots,i_m)&=\mathbf{X}_{(n)}(i_n,j), \quad
    j=1+\sum^{m}_{\substack{k=1\\k\neq n}}(i_k-1)q_k \hspace{0.2cm} \text{with} \hspace{0.2cm} q_k=\prod^{k-1}_{\substack{p=1\\p\neq n}}d_p.
    \end{aligned}
\end{equation}

Based on this unfolding operation, the tensor-matrix product ($n$-mode product) is defined as:
\begin{equation}
    \mathcal{Y}=\mathcal{X}\times_{n}\mathbf{U}\iff \mathbf{Y}_{(n)}=\mathbf{U}\mathbf{X}_{(n)},
\end{equation}

\vspace{0.05cm}
\textbf{b) $\star_\mathbf{M}$-product} 
\vspace{0.05cm}

Within the T-SVDM framework, the $\star_\mathbf{M}$-product defines tensor operations for third-order tensors. Consider $\mathcal{A}\in\mathbb{C}^{d\times{m}\times{n}}$ and an invertible transformation matrix $\mathbf{M}\in\mathbb{C}^{n\times{n}}$. We first define the transformed tensor in the $\mathbf{M}$-domain as:
\begin{equation}                      
 \hat{\mathcal{A}}:=\mathcal{A}\times_{3}\mathbf{M}.
\end{equation}

For another tensor $\mathcal{B}\in\mathbb{C}^{m\times{l}\times{n}}$, the $\star_\mathbf{M}$-product $\mathcal{C} = \mathcal{A}\star_\mathbf{M}\mathcal{B}$ is given by:
\begin{equation} 
    \begin{aligned}
        \hat{\mathcal{C}}_{:: i}&=\hat{\mathcal{A}}_{::i}\hat{\mathcal{B}}_{::i}, \quad \forall i\in\{1,2,\cdots,n\}, \\
        \mathcal{C}&=\hat{\mathcal{C}}\times_{3}\mathbf{M}^{-1}\in{\mathbb{C}^{d\times{l}\times{n}}}.
    \end{aligned}
\end{equation}

Throughout this work, $\hat{\mathcal{A}}$ consistently denotes $\mathcal{A}\times_{3}\mathbf{M}$. The $\star_\mathbf{M}$-product performs slice-by-slice matrix multiplications to achieve tensor-tensor products in the transform domain, making it inherently orientation-sensitive. Consequently, different tensor orientations yield distinct results.

The identity tensor $\mathcal{I}$ under the $\star_{\mathbf{M}}$-product satisfies:
\begin{equation}
    \mathcal{A}\star_{\mathbf{M}}\mathcal{I} = \mathcal{I}\star_{\mathbf{M}}\mathcal{A}^H = \mathcal{A}.
\end{equation}
In the transform domain, each frontal slice of $\hat{\mathcal{I}}$ corresponds to an identity matrix $\mathbf{I}$.

The conjugate transpose of $\mathcal{A}\in\mathbb{C}^{d\times{m}\times{n}}$ under $\star_{\mathbf{M}}$ is defined as:
\begin{equation}
    \left(\hat{\mathcal{A}}^H\right)_{::i} = \left(\hat{\mathcal{A}}_{::i}\right)^H, \quad \forall i\in\{1,\cdots,n\},
\end{equation}
which ensures the fundamental property:
\begin{equation}
    \mathcal{A}^H\star_{\mathbf{M}}\mathcal{B}^H = \left(\mathcal{B}\star_{\mathbf{M}}\mathcal{A}\right)^H.
\end{equation}

A tensor $\mathcal{A}\in\mathbb{C}^{m\times{m}\times{n}}$ is $\star_{\mathbf{M}}$-orthogonal if it satisfies:
\begin{equation}
    \mathcal{A}^H\star_{\mathbf{M}}\mathcal{A} = \mathcal{A}\star_{\mathbf{M}}\mathcal{A}^H = \mathcal{I}.
\end{equation}




\subsection{T-SVDM}

Within the $\star_{\mathbf{M}}$-product framework, the T-SVDM decomposition \cite{T-SVDM_T-SVDM2} factorizes a third-order tensor $\mathcal{A}\in\mathbb{C}^{d_1\times d_2\times d_3}$ as:
\begin{equation}
    \mathcal{A}=\mathcal{U} \star_{\mathbf{M}} \mathcal{S} \star_{\mathbf{M}} \mathcal{V}^H, 
\label{T-SVDM}
\end{equation}
where $\mathcal{U}\in\mathbb{C}^{d_1\times d_1 \times d_3}$ and $\mathcal{V}\in\mathbb{C}^{d_2\times d_2 \times d_3}$ are $\star_{\mathbf{M}}$-unitary tensors. $\mathcal{S}\in\mathbb{C}^{d_1\times d_2\times d_3}$ is an \emph{f-diagonal} tensor with diagonal frontal slices.
This decomposition generalizes two important cases: \textbf{T-SVD} \cite{T-SVD}: Special case when $\mathbf{M}$ is the DFT matrix; \textbf{T-SVDM} \cite{T-SVDM_T-SVDM2}: General case with arbitrary invertible $\mathbf{M}$.

The connection between T-SVDM and TPCA was first established in \cite{Facial_recognition_T-SVD}, which applied T-SVD to facial image compression through data tensor pre-rotation. This work was later extended to T-SVDM in \cite{T-SVDM_T-SVDM2}. Notably, \cite{2DTPCA} demonstrated the critical role of orientation-dependence in T-SVD for multi-directional feature extraction. Building upon these foundations, we develop a novel sparse tensor model that integrates T-SVDM with sparse PCA.

\subsection{PCA and Sparse PCA (SPCA)}
Principal Component Analysis (PCA) \cite{PCA} is a fundamental unsupervised feature extraction method. It seeks an orthogonal linear transformation that projects high-dimensional data onto a lower-dimensional manifold while maximizing the variance of the projected samples. The PCA optimization problem can be expressed in the following self-contained regression form:
\begin{equation}
    \min\limits_{\mathbf{U}^T\mathbf{U}=\mathbf{I}_k} {\Vert{\mathbf{X}-\mathbf{U}\mathbf{U}^{T}\mathbf{X}}\Vert}^2_F, \label{PCA}
\end{equation}
where $\mathbf{X}=\left[\mathbf{x}_1,\mathbf{x}_2,\mathbf{x}_3,\cdots,\mathbf{x}_n\right]\in\mathbb{R}^{d\times{n}}$ is the data matrix, with each column representing a sample. $\mathbf{U}\in\mathbb{R}^{d\times{k}}$ ($k \ll d$) is an orthogonal transformation matrix. 

To enhance interpretability through sparse feature weights, Sparse PCA (SPCA) \cite{SPCA} introduces ridge and lasso constraints to the PCA framework:
\begin{equation}
\begin{aligned} \label{SPCA}
    \min\limits_{\mathbf{U},\mathbf{Q}} \quad &{\Vert {\mathbf{X}-\mathbf{U}\mathbf{Q}^T\mathbf{X}} \Vert}^2_F+\alpha\sum\limits_{j=1}^k{\Vert {\mathbf{q}_j}\Vert}^2_2+\beta\sum\limits_{j=1}^k{\Vert {\mathbf{q}_j}\Vert}_1\\
    s.t. \quad&{\mathbf{U}^T\mathbf{U}=\mathbf{I}_k},
\end{aligned}
\end{equation}
where $\alpha,\beta >0$ are regularization parameters. 

\begin{figure}
    \centering
    \includegraphics[width=5in]{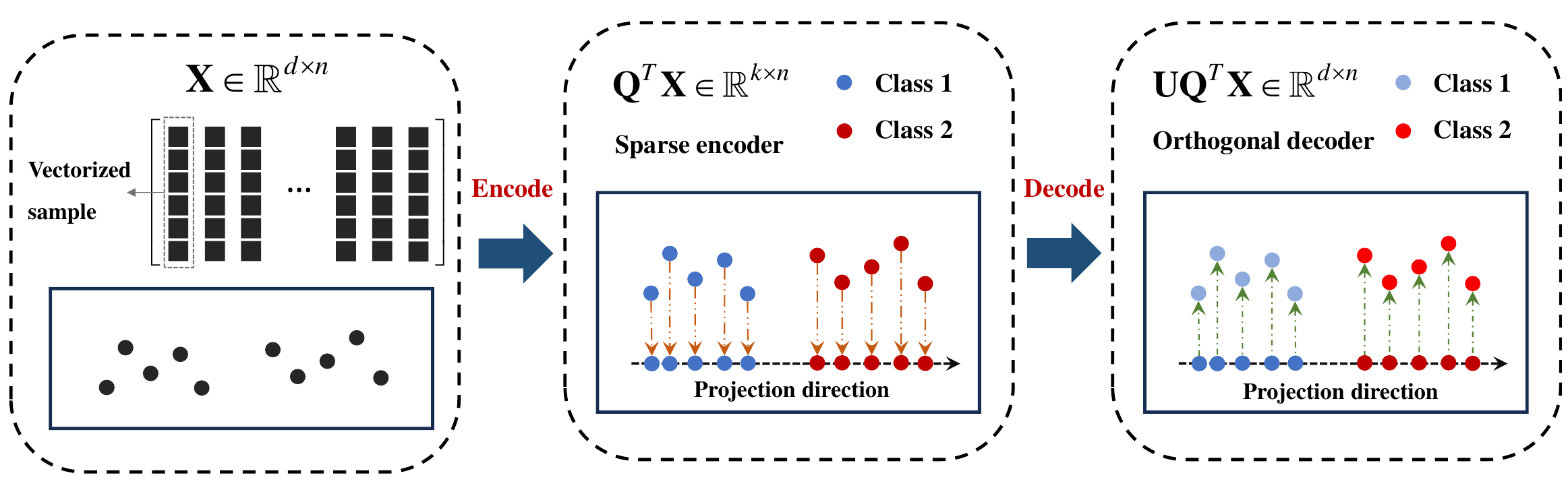}
    \caption{The encoder-decoder principle of one-dimensional SPCA}
    \label{The encoder-decoder principle of SPCA}
\end{figure}

The self-contained regression term in SPCA implements an encoder-decoder architecture: 
\begin{itemize}
    \item \textit{Encoder} ($\mathbf{Q}$): Selectively transforms a sparse subset of input features into latent representations
    \item \textit{Decoder} ($\mathbf{U}$): Reconstructs original data through an orthogonal transformation
\end{itemize}
as illustrated in Figure \ref{The encoder-decoder principle of SPCA}. This formulation enhances flexibility by decoupling the transformation matrix $\mathbf{Q}$ from $\mathbf{U}$. Notably, when $\mathbf{Q}=\mathbf{U}$, SPCA reduces to standard PCA. 

Through the minimization of reconstruction errors, SPCA preserves global data structures in an unsupervised manner while maintaining essential geometric relationships within the data manifold. The lasso constraint enables selective feature weighting by assigning non-zero weights only to necessary features. Simultaneously, the orthogonality constraint ensures complementarity of latent features and reduces redundancy. These properties make SPCA particularly well-suited for unsupervised feature selection (UFS), effectively balancing clustering performance and feature interpretability.

\subsection{Tensor-based unsupervised feature selection}

Tensor-based unsupervised feature selection (UFS) remains an emerging research domain where most existing methods rely heavily on domain-specific knowledge, limiting their applicability to particular scenarios. For instance, \cite{UFS_Tucker_vehicle_detection_2019} proposes an image-transformation-based feature selection method specialized for vehicle detection, while \cite{HOSVD_FS_multi_omics_bio} constructs multi-omics data tensors with feature selection criteria designed using bioinformatics knowledge. Although recent methods targeting general applications have emerged—such as GRLTR \cite{GRLTR_UFS_RTPCA_Graph_2018} which combines graph regularization with tensor tubal rank constraints to preserve global and local information—they still require vectorization. CPUFS \cite{CPUFS_2023} represents a promising tensor-based approach that preserves tensor structures via CP decomposition reconstruction, but it fails to derive sparse factor matrices. Instead, CPUFS applies spectral clustering and utilizes a sparse linear classifier matrix for feature selection, leading to insufficient utilization of factor matrix information and reduced interpretability. Furthermore, its graph construction requires prior knowledge of class numbers and remains vulnerable to noisy features.

Recently, sparse PCA models constitute a growing trend in UFS \cite{CSPCA, AW-SPCA, SPCAFS, SPCA-PSD} due to their ability to avoid graph construction dependencies, preserve data structures through global manifold learning, deliver superior average performance, and provide interpretable sparse weights via variance-maximizing factor matrices. However, existing PCA-based methods lack tensor formulations, limiting their applicability to higher-dimensional data. To bridge this gap, we develop a novel sparse TPCA-based methodology specifically designed for tensor UFS tasks.

\section{Methodology}

\label{Methodology}
This section extends the SPCA framework to tensor formulation by integrating T-SVDM techniques, establishing a novel \emph{Sparse Tensor PCA based on $\star_{\mathbf{M}}$ product (STPCA-MP)} model. We subsequently derive a convex variant of STPCA-MP for UFS applications by unifying the encoder and decoder tensors into a single sparse reconstruction tensor. To optimize this model efficiently, we develop a slice-by-slice algorithm that independently solves each dual subproblem in the transform domain. Finally, leveraging both the transformation matrix $\mathbf{M}$ and the learned reconstruction tensor, we compute feature score maps to perform feature selection. Figure \ref{Overall framework of our work} illustrates the complete framework, with key mathematical symbols summarized in Table \ref{symbols}.

\begin{table}[]
    \centering
    \caption{Symbols used throughout the deduction.}
    \label{symbols}
    \begin{tabular}{ccc}
    \toprule
        Notation &Description \\
    \midrule   
        $\mathcal{X}\in\mathbb{C}^{d_1\times d_2 \times n }$  &  Data tensor containing $n$ samples \\
        $\hat{\mathcal{X}}\in\mathbb{C}^{d_1\times d_2 \times n }$  &  Data tensor in the transform domain\\
        $\Tilde{\mathcal{X}}\in\mathbb{C}^{d_1\times d_2 \times n }$  &  Reconstructed data tensor\\
        $D^o\in \mathbb{Z}^3$    &     Order set defining the order of modes \\
        $\mathcal{X}_{D^o}\in\mathbb{C}^{q\times n \times p}$  & Data tensor permuted by $D^o$ \\
        $\mathcal{U}$, $\hat{\mathcal{U}}\in\mathbb{C}^{q\times k \times p}$  & Decoder tensor for $\mathcal{X}_{D^o}$ in the original and the transform domain\\
        $\mathcal{Q}$, $\hat{\mathcal{Q}}\in\mathbb{C}^{q\times k \times p}$ & Encoder tensor for $\mathcal{X}_{D^o}$ in the original and the transform domain\\
        $\mathcal{Q}_{::i}$ & The $i$-th frontal slice of $\mathcal{Q}$\\
        $\mathbf{Q}_{(1)}$ & 1-mode unfolding of $\mathcal{Q}$\\
        $ \left( \mathbf{Q}_{(1)}\right)_{:j}$ & The $j$-th column vector in $\mathbf{Q}_{(1)}$ \\
        $\mathcal{A}$, $\hat{\mathcal{A}}\in\mathbb{C}^{q\times q \times p}$ & Reconstruction tensor for $\mathcal{X}_{D^o}$ in the original and the transform domain \\
        $\mathcal{I}\in\mathbb{C}^{q\times q \times p}$ & Identity tensor under $\star_{\mathbf{M}}$ product \\
        $\alpha$    & Regularization parameter of lasso regression \\
        $\beta$     & Regularization parameter of ridge regression \\
        $\lambda$   &  Regularization parameter of $\ell_{2,1}$-norm \\ 
        $\eta$  & Regularization parameter of trace function \\
    \bottomrule
    \end{tabular}
\end{table}

\subsection{Model formulation of STPCA-MP} 
We formulate the model for third-order tensors to simplify deduction and algorithmic development, noting that the principle extends recursively to higher-order scenarios (see Section \ref{Discussion} for generalization details).

Consider a data tensor $\mathcal{X}\in\mathbb{C}^{d_1\times d_2 \times n}$ containing $n$ samples of dimensions $d_1\times d_2$. Through reorientation via an ordered set $D^o$ where the second mode corresponds to the sample dimension, we obtain the rotated tensor $\mathcal{X}_{D^o}\in\mathbb{C}^{q\times n \times p}$. Note that following rotation, each frontal slice of $\mathcal{X}_{D^o}$ undergoes sample-wise mean centering along its second mode (sample dimension). Building on the T-SVDM framework (\ref{T-SVDM}), we reduce row dimensionality of each frontal slice in $\mathcal{X}_{D^o}$ using $\mathcal{U}^H\in\mathbb{C}^{k\times q \times p}$ ($k<d$), followed by reconstruction with $\mathcal{U}$:
\begin{equation}
\begin{aligned}
    \mathcal{Y} &= \mathcal{U}^H \star_\mathbf{M} \mathcal{X}_{D^o}, \\
    \Tilde{\mathcal{X}}_{D^o} &= \mathcal{U} \star_\mathbf{M} \mathcal{Y},
\end{aligned}
\end{equation}
where $\mathcal{Y}\in\mathbb{C}^{k\times n \times p}$, and $\Tilde{\mathcal{X}}_{D^o}\in\mathbb{C}^{q\times n \times p}$ is the reconstructed tensor. 

Intuitively, we can first formulate a tensor PCA model inheriting the principle of data reconstruction and orthogonality constraint in PCA (Problem (\ref{PCA})):

\begin{equation}
\label{TPCA based on T-SVDM}
\begin{aligned}
    &\underset{\mathcal{U}}{\min} \quad {\Vert \mathcal{X}_{D^o} - \mathcal{U} \star_\mathbf{M} \mathcal{U}^H \star_\mathbf{M} \mathcal{X}_{D^o}\Vert^2_F}
    \\
    & \textit{s.t.} \quad  \mathcal{U}^H\star_\mathbf{M}\mathcal{U} = \mathcal{I},
\end{aligned}
\end{equation}
which minimizes the data reconstruction error without compromising the tensor structure. 

Building on the encoder-decoder architecture of SPCA \cite{SPCA}, we introduce a distinct transformation tensor $\mathcal{Q}\in\mathbb{C}^{q\times k \times p}$ to replace $\mathcal{U}$ for feature extraction. This yields a more flexible reconstruction framework that enhances latent feature representation:

\begin{equation}
\label{TPCA based on T-SVDM: Q}
\begin{aligned}
    &\underset{\mathcal{U}}{\min} \quad {\Vert \mathcal{X}_{D^o} - \mathcal{U} \star_\mathbf{M} \mathcal{Q}^H \star_\mathbf{M} \mathcal{X}_{D^o}\Vert^2_F}
    \\
    &\textit{s.t.}  \quad  \mathcal{U}^H\star_\mathbf{M}\mathcal{U} = \mathcal{I}.
\end{aligned}
\end{equation}
Similar to SPCA, Problem (\ref{TPCA based on T-SVDM: Q}) adopts an encoder-decoder framework:
\begin{itemize}
    \item \emph{Tensor encoder} ($\mathcal{Q}$): Projects input features into a latent space while preserving the intrinsic tensor structure and orientation relationships.
    \item \emph{Tensor decoder} ($\mathcal{U}$): Recovers the original tensor through a series of orthogonal transformations, ensuring minimal reconstruction error.
\end{itemize}

To enforce feature selection during encoding, we impose sparsity constraints on the transformation tensor $\mathcal{Q}$, ensuring only discriminative features are retained in the latent representation. For highly correlated features, particularly in high-dimensional scenarios ($d_m \gg n$), we impose equality constraints to prevent model bias toward specific features. This is achieved by incorporating lasso and ridge regression terms as $\mathcal{Q}$ constraints, yielding the sparse tensor PCA model:

\begin{equation}
\label{a sparse tensor PCA model based on T-SVDM}
\begin{aligned}
    &\underset{\mathcal{U},\mathcal{Q}}{\min} \quad {\Vert \mathcal{X}_{D^o} - \mathcal{U} \star_\mathbf{M} \mathcal{Q}^H \star_\mathbf{M} \mathcal{X}_{D^o}\Vert^2_F} +\alpha \sum\limits^{pn}_{j=1} {\Vert \left( \mathbf{Q}_{(1)}\right)_{:j} \Vert^2_2} + \beta \sum\limits^{pn}_{j=1} {\Vert \left( \mathbf{Q}_{(1)}\right)_{:j} \Vert_1}\\
    &\textit{s.t.}  \quad  \mathcal{U}^H\star_\mathbf{M}\mathcal{U} = \mathcal{I},
\end{aligned}
\end{equation}
which can be viewed as an extension of SPCA to a tensor form. Solving for the optimal $\mathcal{Q}$ enables extraction of sparse tensor principal components from the data tensor.

\emph{Discussion:} To this end, the proposed model naturally extends SPCA applications to tensor-structured data, including traditional tasks such as image compression, visualization, and denoising, while preserving intrinsic structural relationships. However, the encoder tensor $\mathcal{Q}$ alone proves suboptimal for UFS due to insufficient feature discriminability. The following section reformulates the model specifically for UFS requirements.

\subsection{Convex version of STPCA-MP for UFS}
As previously established, the self-contained regression term embodies an encoder-decoder architecture. Consequently, feature importance is jointly encoded by $\mathcal{Q}$ (encoder) and $\mathcal{U}$ (decoder). Comprehensive integration of their relationship is essential for UFS adaptation.

\emph{Dual Problem:} Given the bijective mapping $\hat{\mathcal{A}} = \mathcal{A} \times_{3} \mathbf{M}$ between tensors, Problem (\ref{a sparse tensor PCA model based on T-SVDM}) admits a dual formulation in the transform domain. This duality enables simultaneous optimization. For computational efficiency, we recast the problem using matrix algebra:  

\begin{equation}
\label{STPCA-MP: transform domain}
\begin{aligned}
    &\underset{\hat{\mathcal{U}},\hat{\mathcal{Q}}}{\min} \quad \sum\limits^{p}_{i=1}
    {\Vert \left(\hat{\mathcal{X}}_{D^o}\right)_{::i} - \hat{\mathcal{U}}_{::i} \hat{\mathcal{Q}}^H_{::i}\left(\hat{\mathcal{X}}_{D^o}\right)_{::i}\Vert^2_F} + \lambda \sum\limits^{p}_{j=1} {\Vert  \hat{\mathcal{Q}}_{::i} \Vert_{2,1}} \\
    &\text{s.t.} \quad \hat{\mathcal{U}}_{::i}^H\left(\hat{\mathcal{U}}_{::i}\right) = \mathbf{I}, i\in[p].
    \end{aligned}
\end{equation}
The $\ell_{2,1}$ norm concurrently satisfies lasso and ridge regression objectives. Critically, Problem (\ref{STPCA-MP: transform domain}) demonstrates tensor reconstruction via matrix outer products. Compared to vector outer products in one-dimensional SPCA, this formulation more effectively incorporates structural priors into sparse weight optimization.

\begin{figure}
    \centering
    \includegraphics[width=1\linewidth]{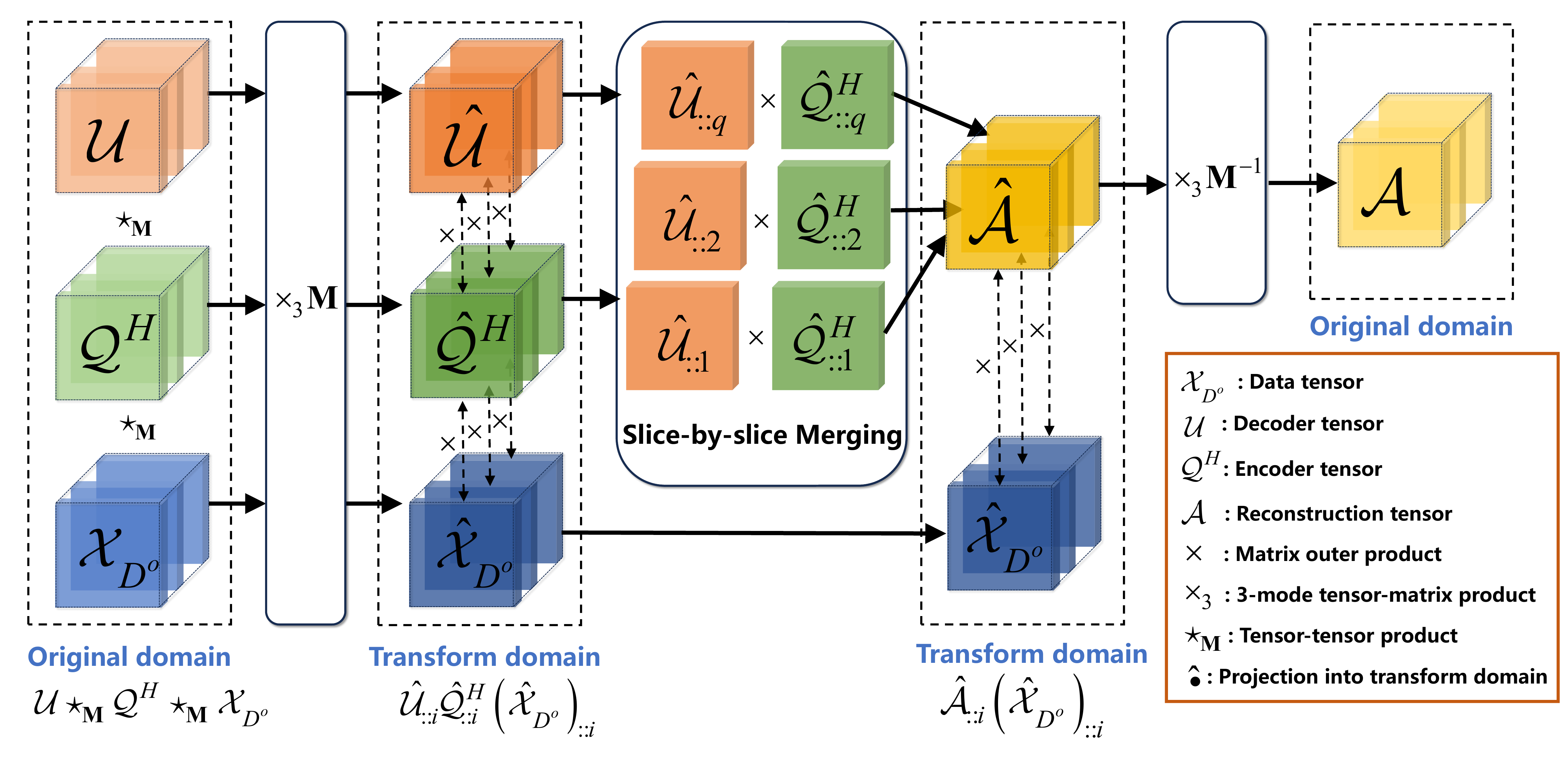}
    \caption{A schematic diagram of merging the encoder and decoder tensors.}
    \label{merging}
\end{figure}

\emph{Variable Merging: }As Problem (\ref{STPCA-MP: transform domain}) is a non-convex problem requiring alternating update of $\hat{\mathcal{U}}$ and $\hat{\mathcal{Q}}$, a convex relaxation is beneficial for optimization. Consequently, we developed a convex formulation strategy through variable merging. As illustrated in Figure \ref{merging}, tensor products in the transform domain are decomposed into multiple outer products of matrices. This enables pairwise multiplication of each frontal slice between the encoder and decoder, ultimately generating a novel reconstruction tensor. Each row vector in $\hat{\mathcal{A}}_{::i}$ integrates reconstruction weights from both $\hat{\mathcal{U}}_{::i}$ and $\hat{\mathcal{Q}}^H_{::i}$, establishing $\mathcal{A}$ as a superior feature selection tensor compared to $\mathcal{Q}$ alone. Guided by $D^o$, each frontal slice of $\hat{\mathcal{A}}$ reconstructs mode-specific fibers in the data tensor, thereby leveraging structural information inherent in the mode organization. 

Theorem \ref{theorem 1} demonstrates that the optimal $\hat{\mathcal{A}}$ constitutes a Hermitian positive semidefinite (HPSD) matrix.

\newtheorem{theorem}{Theorem}
\begin{theorem}
\label{theorem 1}
    Given the optimal solution $\hat{\mathcal{U}}^{opt}_{::i}$ and $\hat{\mathcal{Q}}^{opt}_{::i}$ to Problem (\ref{STPCA-MP: transform domain}), then $\hat{\mathcal{A}}^{opt}_{::i}=\hat{\mathcal{U}}^{opt}_{::i}(\hat{\mathcal{Q}}^{opt}_{::i})^H\in H^q_+$.
\end{theorem}

\newproof{pf}{Proof of Theorem \ref{theorem 1}}
\begin{pf}
    See Appendix.
\end{pf}

\emph{Norm Simplification: }Given $\hat{\mathcal{A}}_{::i}=\hat{\mathcal{U}}_{::i} \hat{\mathcal{Q}}^H_{::i}$ and $d\gg k$, each $\hat{\mathcal{A}}_{::i}$ is inherently low-rank. While the nuclear norm $\Vert \cdot  \Vert_*$ serves as the standard convex approximation for low-rank constraints, its computational demands present significant challenges. Fortunately, the HPSD constraint enables the equivalence $\Vert \hat{\mathcal{A}}_{::i} \Vert_* = {\rm Tr}\left(\hat{\mathcal{A}}_{::i}\right)$, dramatically reducing computational complexity.

Furthermore, leveraging the orthogonality of $\hat{\mathcal{U}}_{::i}$, we establish:
\begin{equation}
    {\Vert \hat{\mathcal{Q}}_{::i}\Vert}_{2,1}={\Vert \hat{\mathcal{U}}_{::i}\hat{\mathcal{Q}}^H_{::i}\Vert}_{2,1}={\Vert \hat{\mathcal{A}}_{::i}\Vert}_{2,1}={\Vert \hat{\mathcal{A}}^H_{::i}\Vert}_{2,1}.
\end{equation}
This enables substitution with ${\Vert \hat{\mathcal{A}}_{::i}\Vert}_{2,1}$ to streamline optimization. Moreover, when $\mathbf{M}$ is orthogonal, minimizing ${\Vert \hat{\mathcal{A}}_{::i}\Vert}_{2,1}$ and ${\rm Tr}\left(\hat{\mathcal{A}}_{::i}\right)$ is equivalent to minimizing ${\Vert\mathcal{A}_{::i}\Vert}_{2,1}$ and ${\rm Tr}\left(\mathcal{A}_{::i}\right)$. Consequently, column vectors in $\mathcal{A}_{::i}$ (representing weights for individual features) are collectively minimized toward zero for redundant features, thereby fulfilling feature selection.

Finally, a convex version of STPCA-MP for UFS is formulated as follows:

\begin{equation}
\label{STPCA-MP}
\begin{aligned}
    \underset{\mathcal{A}}{\min} \quad &{\Vert \mathcal{X}_{D^o} - \mathcal{A} \star_\mathbf{M} \mathcal{X}_{D^o}\Vert^2_F} + \lambda\sum\limits^{p}_{i=1} {\Vert  \mathcal{A}}_{::i} \Vert_{2,1} + \eta\sum\limits^{p}_{i=1}  {\rm Tr}\left( \mathcal{A}_{::i}\right)\\
    \text{s.t.} \quad &\hat{\mathcal{A}}_{::i} \in H^q_+, i\in[p].
\end{aligned}
\end{equation}
The HPSD constraint, while emerging naturally from variable merging, plays a dual role: it statistically guarantees positive contributions from all projection directions to the reconstruction process. This property intrinsically suppresses random noise features, leading to enhanced weight interpretability and ultimately more effective feature selection.

\emph{Discussion: } The convex STPCA-MP formulation achieves model compactness through variable merging. By unifying the sparse encoder and orthogonal decoder, the reconstruction tensor $\mathcal{A}$ simultaneously weights original features and their latent representations, providing a comprehensive feature importance measure.

\subsection{Optimization Algorithm}

Solving Problem (\ref{STPCA-MP}) efficiently requires leveraging the tensor structure to handle HPSD constraints with reduced computational complexity.

\emph{Multiple Dual Subproblems: } Due to the bijective mapping $\hat{\mathcal{A}} = \mathcal{A} \times_{3} \mathbf{M}$, Problem (\ref{STPCA-MP}) admits a dual formulation in the transform domain. This dual problem proves computationally tractable as shown below. We first recast it as:
\begin{equation}
\label{STPCA-MP: matrix form}
    \begin{aligned}
        \underset{\hat{\mathcal{A}}}{\min} \quad
        &\sum\limits^{p}_{i=1}
        \left({\Vert \left(\hat{\mathcal{X}}_{D^o}\right)_{::i} - \hat{\mathcal{A}}_{::i} \left(\hat{\mathcal{X}}_{D^o}\right)_{::i}\Vert^2_F} + \lambda{\Vert \hat{\mathcal{A}}_{::i} \Vert}_{2,1} 
        +  \eta {\rm Tr}\left( \hat{\mathcal{A}}_{::i}\right)
        \right)\\
        \text{s.t.} \quad &\hat{\mathcal{A}}_{::i} \in H^q_+, \quad i\in[p],
    \end{aligned}
\end{equation}
which decomposes into $p$ independent subproblems. This enables a divide-and-conquer strategy: solve each frontal slice optimization separately, then reconstruct $\mathcal{A}$ post-optimization.

\emph{Derivative Analysis:} For the $i$-th subproblem, setting the gradient with respect to $\hat{\mathcal{A}}_{::i}$ to zero yields:
\begin{equation}
\label{Derivative of A}
    \hat{\mathcal{A}}_{::i} = \left(\mathbf{S}_i-\frac{\eta_{i}}{2}\mathbf{I}_q\right)\left(\mathbf{S}_i+\lambda_i{\mathbf{W}_i}+\epsilon_1\mathbf{I}_q\right)^{-1},
\end{equation}
where $\mathbf{S}_i=\left(\hat{\mathcal{X}}_{D^o}\right)_{::i}\left(\hat{\mathcal{X}}_{D^o}\right)^T_{::i}$, $\epsilon_1>0$ is a small constant added for inversion stability, and $\mathbf{W}_i\in{\mathbb{R}^{q\times q}}$ denotes a diagonal matrix with $j$-th diagonal element:
\begin{equation}
\label{W}
\mathbf{w}_{jj}=\left(1/\left(2\sqrt{\hat{\mathbf{a}}_j^H\hat{\mathbf{a}}_j+\epsilon_2}\right)\right).
\end{equation}
$\epsilon_2>0$ is a small constant to prevent the denominator from approaching zero.

\emph{HPSD Projection:} Subsequently, $ \hat{\mathcal{A}}_{::i} $ is projected onto the HPSD cone using the operator $P_{H^d_+}(\cdot)$ \cite{PSD}. For an arbitrary square matrix $\mathbf{A}\in\mathbb{C}^{b\times b}$, this projection is defined through eigenvalue decomposition:
\begin{equation}
\begin{aligned}
    P_{H^b_+}(\mathbf{A}) &= \Pi_{H^b_{+}}\left(\Pi_{H^b}(\mathbf{A})\right) \\
    &= \Pi_{H^b_{+}}\left(\frac{1}{2}(\mathbf{A}+\mathbf{A}^H)\right)\\
    &=\sum\limits_{i=1}^{b}\max\left\{\sigma_i,0\right\}\mathbf{u}_i\mathbf{u}_i^H,   
\end{aligned}
\end{equation}
where $\frac{1}{2}(\mathbf{A}+\mathbf{A}^H) = \sum\nolimits_{i=1}^{b}\sigma_i\mathbf{u}_i\mathbf{u}_i^H$ is its eigenvalue decomposition.

Building on these techniques, we solve the $i$-th subproblem of (\ref{STPCA-MP: matrix form}) using a two-stage iterative algorithm:

\textbf{Step 1: Update $\hat{\mathcal{A}}_{::i}$ with $\mathbf{W}_i$ fixed.}
For fixed $\mathbf{W}_i$, the optimal $\hat{\mathcal{A}}_{::i}$ is given by:
\begin{equation}
\label{Update A for STPCA-MP}
    \hat{\mathcal{A}}_{::i} = P_{H^{q}_+}\left(\left(\mathbf{S}-\frac{\eta_{i}}{2}\mathbf{I}_q\right)\left(\mathbf{S}+\lambda_i{\mathbf{W}_i}+\epsilon_2\mathbf{I}_q\right)^{-1}\right).
\end{equation}

\textbf{Step 2: Update $\mathbf{W}_i$ with $\hat{\mathcal{A}}_{::i}$ fixed.}
For fixed $\hat{\mathcal{A}}_{::i}$, $\mathbf{W}_i$ is directly computed via (\ref{W}).

\emph{Discussion:} This convex subproblem formulation significantly simplifies optimization design. While HPSD projection requires eigenvalue decomposition with $O(x^3)$ complexity, tensor structure exploitation ensures $x \ll |\mathcal{F}|$ where $\mathcal{F}$ denotes the feature set. Experimental results demonstrate the method's computational efficiency.

\subsection{Feature selection via score map}
We quantify feature importance by its contribution to tensor reconstruction, accounting for the transform domain induced by $\mathbf{M}$. 

Suppose we reorient the data tensor to obtain $\mathcal{X}_{D^o}\in\mathbb{C}^{p\times n \times q}$. For the $ij$-th feature $f_{ij}$, its transform-domain $\hat{f}_{ij}$ counterpart and reconstructed version $\Tilde{f}_{ij}$ are expressed as:
\begin{equation}
\begin{aligned}
     \hat{f}_{ij} &= \sum\limits^{q}_{h=1}m_{jh}f_{ih}, \quad \text{and}\\
    \Tilde{f}_{ij} &= \sum\limits^{p}_{l=1}\hat{a}_{ilj}\hat{f}_{lj}=\sum\limits^{p}_{l=1}\left(\sum\limits^{p}_{k=1}m_{ik}a_{ilk}\right)\left(\sum\limits^{q}_{h=1}m_{jh}f_{lh}\right)=\sum\limits^{p}_{l=1}\sum\limits^{q}_{h=1}\left(\sum\limits^{p}_{k=1}m_{jh}m_{ik}a_{ilj}\right)f_{lh}, 
\end{aligned}
\end{equation}

The contribution of feature $f_{lh}$ to the reconstruction of all features is quantified by the matrix $\mathbf{C}^{lh} \in \mathbb{C}^{p \times q}$:
\begin{equation}
\mathbf{C}^{lh}_{ij} = \sum_{k=1}^{p} m_{jh} m_{ik} a_{ilj}.
\label{Contribution_matrix}
\end{equation}
The feature importance score is then defined as:
\begin{equation}
score(f_{lh}) = \|\mathbf{C}^{lh}\|_F^2 = \sum_{i=1}^{p} \sum_{j=1}^{q} \left( \sum_{k=1}^{p} m_{jh} m_{ik} a_{ilj} \right)^2, 
\label{feature_score}
\end{equation}
which comprehensively captures $f_{lh}$'s global influence on feature reconstruction. The complete procedure is summarized in \textbf{Algorithm \ref{Algorithm of STPCA-MP for one direction}}.

\emph{Discussion:} Equation (\ref{Contribution_matrix}) demonstrates that an appropriately chosen $\mathbf{M}$ effectively integrates cross-mode feature correlations into the slice-by-slice optimization process. Our scoring mechanism globally quantifies feature dependencies through tensor operations, eliminating the need for sample vectorization. This approach provides dual advantages: 
\begin{itemize}
    \item Preservation of intrinsic data organization structures
    \item Significant reduction in computational complexity
\end{itemize}
by maintaining the native tensor representation throughout the feature evaluation process.

\subsection{Implementation details}
The proposed optimization framework decomposes into multiple independent subproblems amenable to parallel computation. We establish a unified convergence criterion for all subproblems:
\begin{equation}
    |\text{obj}^{(h)} - \text{obj}^{(h-1)}| < 10^{-5}, \quad h=1,2,3,\cdots
\end{equation}
where $\text{obj}^{(h)}$ denotes the objective function value at iteration $h$.

For regularization parameters $\lambda$ and $\eta$, we optimized them via grid search per dataset in the following comparison experiments. In practice, $\lambda$ and $\eta$ can be set to $10$ for a rapid deployment. The stabilization constants are fixed at $\epsilon_1 = 10^{-3}$ and $\epsilon_2 = 10^{-2}$ throughout all experiments.

For data preprocessing, we apply centralization to $\left(\mathcal{X}_{D^o}\right)_{::i}$ for each subproblem individually to capture inter-dimension variance more effectively. To narrow down the grid search range, we normalize $\mathcal{X}$ by dividing each element by the maximum absolute value, scaling all values to the interval [-1, 1].

\begin{algorithm}[t] 
\caption{The optimization algorithm for STPCA-MP} 
\label{Algorithm of STPCA-MP for one direction}
\renewcommand{\algorithmicrequire}{\textbf{Input:}}
\renewcommand{\algorithmicensure}{\textbf{Output:}}
\begin{algorithmic}[1] \REQUIRE A third-order centralized data tensor $\mathcal{X}\in\mathbb{C}^{d_1\times d_2 \times n}$, the order set $D^o$ determining $q$, an invertible matrices $\mathbf{M}\in\mathbb{C}^{q\times q}$,  regularization parameters $\lambda,\eta$, and $\epsilon_1=10^{-3}$, $\epsilon_2=10^{-2}$.
\ENSURE $h$ selected features.
\STATE Randomly initialize $\left(\hat{\mathcal{A}}\right)_{::i}\in{H^{q}_+}$, transform $\mathcal{X}$ into $\hat{\mathcal{X}}$.
\FOR{$i=1,2,\cdots,q$}
\STATE revolve $\hat{\mathcal{X}}$ with $D^o$ and obtain $\hat{\mathcal{X}}_{D^o}$.
\STATE input centralized $\left(\hat{\mathcal{X}}_{D^o}\right)_{::i}$ and alternatively update $\hat{\mathcal{A}}_{::i}$ and $\mathbf{W}_i$ according to (\ref{Update A for STPCA-MP}) until $|\text{obj}^{(h)} - \text{obj}^{(h-1)}| < 10^{-5}$.
\ENDFOR
\STATE  Compute $\mathcal{A}=\hat{\mathcal{A}}\times_3 \mathbf{M}^{-1} $ and $\|\mathbf{C}^{lh}\|_F^2$ for each score according to (\ref{Contribution_matrix}). 
\STATE Output a feature score map $score$.
\STATE Sort $score$ in descending order, and select features of the $h$ greatest scores. For a slice-wise tensor, the score for each dimension will be obtained by summation.
\end{algorithmic}
\end{algorithm}

\section{Discussion}
\label{Discussion}
In this section, we discuss the convergence, computational complexity, and extension to higher-order tensor data of STPCA-MP.

\subsection{Convergence analysis}

First, we prove the convergence of the proposed algorithm. The proof leverages the following fundamental lemma \cite{SOGFS}:
\newtheorem{lemma}{Lemma}
\begin{lemma} \label{lemma1}
For any nonzero vectors $\mathbf{a}$, $\mathbf{b}\in{\mathbb{R}^{c\times{1}}}$, the following inequality holds:
\begin{equation}
    {\Vert \mathbf{a}\Vert_2}-\frac{{\Vert\mathbf{a}\Vert_2^2}}{2{\Vert \mathbf{b}\Vert_2}} \leq {\Vert \mathbf{b}\Vert_2}-\frac{{\Vert \mathbf{b}\Vert_2^2}}{2{\Vert \mathbf{b}\Vert_2}}.
\end{equation}
\end{lemma}

Based on \textbf{Lemma} \ref{lemma1}, we have the following theorem:

\begin{theorem}
\label{convergence}
      The objective function value of each subproblem in Problem (\ref{STPCA-MP: matrix form}) is non-increasing in each iteration of Algorithm \ref{Algorithm of STPCA-MP for one direction} and can converge to a global minimum.
\end{theorem}

\newproof{pf2}{Proof of \ref{convergence}}
\begin{pf2}
 For simplicity, we temporally use $\mathbf{A}$ and $\mathbf{X}$ to replace $\hat{\mathcal{A}}_{::i}$ and $\left(\hat{\mathcal{X}}_{D^o}\right)_{::i}$. Denote the updated $\mathbf{A}$ as $\mathbf{\Tilde{A}}$ and fix $\mathbf{W}$. By rewriting the $i$-th subproblem in (\ref{STPCA-MP: matrix form}) with the trace function, we have the following inequality
 \begin{equation}\label{Ineq 1}
    \begin{aligned}
        {\rm Tr}\left(\mathbf{\Tilde{A}}\mathbf{S}\mathbf{\Tilde{A}}^H\right)&-2Tr\left(\mathbf{S}\mathbf{\Tilde{A}}\right)+\lambda {\rm Tr}(\mathbf{\Tilde{A}}\mathbf{W}\mathbf{\Tilde{A}}^H)+\eta {\rm Tr}(\mathbf{\Tilde{A}})\\
        &\leq {\rm Tr}(\mathbf{A}\mathbf{S}\mathbf{A}^H)-2Tr(\mathbf{S}\mathbf{A})+\lambda {\rm Tr}(\mathbf{A}\mathbf{W}\mathbf{A}^H) +\eta {\rm Tr}(\mathbf{A}).
    \end{aligned}
 \end{equation}\par
 Denote $J(\mathbf{A})={\rm Tr}(\mathbf{A}\mathbf{S}\mathbf{A}^H)-2Tr(\mathbf{S}\mathbf{A})+\lambda {\rm Tr}(\mathbf{A}\mathbf{W}\mathbf{A}^H)+\eta {\rm Tr}(\mathbf{A})$ and $J(\mathbf{\Tilde{A}})$ as the updated $J(\mathbf{A})$. Inequality (\ref{Ineq 1}) can be rewritten as 
 \begin{equation}\label{Ineq 2}
    J(\mathbf{\Tilde{A}}) + \lambda {\rm Tr}(\mathbf{\Tilde{A}}\mathbf{W}\mathbf{\Tilde{A}}^H) \leq J(\mathbf{A}) + \lambda {\rm Tr}(\mathbf{A}\mathbf{W}\mathbf{A}^H).
 \end{equation}
 Adding the same item $\sum\limits_{j=1}^d{\lambda\epsilon}/{2\sqrt{\mathbf{a}_j^H\mathbf{a}_j+\epsilon}}$ to both sides of (\ref{Ineq 2}), we obtain
 \begin{equation}\label{Ineq 3}
 \begin{aligned}
    J(\mathbf{\Tilde{A}}) &+ \lambda {\rm Tr}(\mathbf{\Tilde{A}}\mathbf{W}\mathbf{\Tilde{A}}^H) + \sum\limits_{j=1}^d\frac{{\lambda\epsilon}}{{2\sqrt{\mathbf{a}_j^H\mathbf{a}_j+\epsilon}}}  \leq J(\mathbf{A}) + \lambda {\rm Tr}(\mathbf{A}\mathbf{W}\mathbf{A}^H)+\sum\limits_{j=1}^d\frac{{\lambda\epsilon}}{{2\sqrt{\mathbf{a}_j^H\mathbf{a}_j+\epsilon}}}.
\end{aligned}
\end{equation}
In the meantime, we have
\begin{equation}\label{equ 1}
    {\rm Tr}(\mathbf{A}\mathbf{W}\mathbf{A}^H)={\rm Tr}(\mathbf{A}^H\mathbf{W}\mathbf{A})=\sum\limits_{j=1}^d\frac{{\lambda{\mathbf{a}_j^H\mathbf{a}_j}}}{{2\sqrt{\mathbf{a}_j^H\mathbf{a}_j+\epsilon}}}.
\end{equation}
The same equality holds for ${\rm Tr}(\mathbf{\Tilde{A}}\mathbf{W}\mathbf{\Tilde{A}}^H)$. We substituted (\ref{equ 1}) into (\ref{Ineq 3}) and get
\begin{equation} \label{Ineq 4}
    J(\mathbf{\Tilde{A}})+\sum\limits_{j=1}^d\frac{{\lambda({\mathbf{\Tilde{a}}_j^H\mathbf{\Tilde{a}}_j}+\epsilon)}}{{2\sqrt{\mathbf{a}_j^H\mathbf{a}_j+\epsilon}}} \leq J(\mathbf{A})+\sum\limits_{j=1}^d\frac{{\lambda({\mathbf{a}_j^H\mathbf{a}_j}+\epsilon)}}{{2\sqrt{\mathbf{a}_j^H\mathbf{a}_j+\epsilon}}}.
\end{equation}
According to \textbf{Lemma} \ref{lemma1}, we have
\begin{equation}\label{Ineq 5}
    \sqrt{{\mathbf{\Tilde{a}}_j^H\mathbf{\Tilde{a}}_j}+\epsilon}-\frac{{{\mathbf{\Tilde{a}}_j^H\mathbf{\Tilde{a}}_j}+\epsilon}}{{2\sqrt{\mathbf{a}_j^H\mathbf{a}_j+\epsilon}}} \leq \sqrt{{\mathbf{a}_j^H\mathbf{a}_j}+\epsilon}-\frac{{{\mathbf{a}_j^H\mathbf{a}_j}+\epsilon}}{{2\sqrt{\mathbf{a}_j^H\mathbf{a}_j+\epsilon}}}.
\end{equation}
Further, we can accumulate (\ref{Ineq 4}) from $j=1$ to $j=d$. Considering $\lambda>0$, we can get
\begin{equation}\label{Ineq 6}
\begin{aligned}
        \lambda\sum\limits_{j=1}^d\sqrt{{\mathbf{\Tilde{a}}_j^H\mathbf{\Tilde{a}}_j}+\epsilon}-\lambda\sum\limits_{j=1}^d\frac{{{\mathbf{\Tilde{a}}_j^H\mathbf{\Tilde{a}}_j}+\epsilon}}{{2\sqrt{\mathbf{a}_j^H\mathbf{a}_j+\epsilon}}} \leq \lambda\sum\limits_{j=1}^d\sqrt{{\mathbf{a}_j^H\mathbf{a}_j}+\epsilon}-\lambda\sum\limits_{j=1}^d\frac{{{\mathbf{a}_j^H\mathbf{a}_j}+\epsilon}}{{2\sqrt{\mathbf{a}_j^H\mathbf{a}_j+\epsilon}}}.
\end{aligned}
\end{equation}
By summing (\ref{Ineq 4}) and (\ref{Ineq 6}), we have
\begin{equation}
    J(\mathbf{\Tilde{A}})+ \lambda\sum\limits_{j=1}^d\sqrt{{\mathbf{\Tilde{a}}_j^H\mathbf{\Tilde{a}}_j}+\epsilon} \leq J(\mathbf{A}) +  \lambda\sum\limits_{j=1}^d\sqrt{{\mathbf{a}_j^H\mathbf{a}_j}+\epsilon}.
\end{equation}
Finally, we can get the following inequality
\begin{equation}
    J(\mathbf{\Tilde{A}})+\lambda{\Vert {\mathbf{\Tilde{A}}}\Vert_{2,1}} \leq J(\mathbf{A})+\lambda{\Vert {\mathbf{A}}\Vert_{2,1}}.
\end{equation}
Each subproblem in (\ref{STPCA-MP: matrix form}) is convex and bounded below by 
$0$, ensuring convergence to the global minimum. Since the subproblems are solved independently, Problem (\ref{STPCA-MP: matrix form}) is guaranteed to converge to a global solution. When the solution in the transform domain is obtained, the dual problem in the original domain is solved as well.
\end{pf2}

\subsection{Computational complexity analysis}

Consider a data tensor $\mathcal{X} \in \mathbb{C}^{d_1 \times d_2 \times n}$ reoriented as $\mathcal{X}_{D^o} \in \mathbb{C}^{q \times n \times p}$, where $n$ denotes the number of samples and $(p,q) \in \{(d_1,d_2),(d_2,d_1)\}$. The computational cost comprises three main components:

\begin{itemize}
    \item \textbf{Transform domain conversion}: $O(q^2)$ operations for data transformation and covariance matrix initialization per frontal slice.
    \item \textbf{Iterative optimization}: $O(q^3)$ per iteration for reconstruction matrix updates (including matrix inversion and HPSD projection).
    \item \textbf{Feature scoring}: $O(p^2q)$ for final importance evaluation.
\end{itemize}
For $p$ independent subproblems, each requiring $t_i$ iterations, the total complexity is:
\begin{equation}
O\left(\sum_{i=1}^p (q^2 n + q^3 t_i) + p^2 q\right)
\label{eq:complexity}
\end{equation}

\emph{Computational Efficiency Advantages:} Table~\ref{Computational complexity} summarizes the computational complexity of comparable methods in subsequent experiments. The proposed STPCA-MP framework achieves significant computational efficiency through three core mechanisms: First, dimensionality reduction via tensor decomposition leverages the inherent structure by exploiting the inequality $q\times p \gg \max(q, p)$ through slice-by-slice operations, systematically replacing high-dimensional matrix computations with mode-decomposed equivalents and reducing per-iteration complexity from $\prod_i d_i^x$ to $d_i^x$. For $32 \times 32$ image data, this reduces complexity from $O(1024^3)$ (as in vectorized methods SOGFS \cite{SOGFS} and SPCAFS \cite{SPCAFS}) to $O(32\times(32^3))$, representing a 1024-fold reduction. Second, linear sample scalability ensures complexity grows as $O(n)$, maintaining large-scale dataset efficiency, in contrast to the quadratic sample complexity $O(n^2)$ incurred by graph-updating methods like CPUFS \cite{CPUFS_2023}. Third, the convex formulation with HPSD projection enables stable, rapid convergence, reducing iteration counts and further conserving runtime. Crucially, STPCA-MP uniquely maintains all advantages simultaneously, preserving the original tensor structure while achieving linear scaling in both feature dimensions ($d_i$) and sample size ($n$)—a dual efficiency unattained by approaches like CPUFS and CPUFSnn that sacrifice sample efficiency for dimensionality reduction.

\begin{table}[]
    \centering
    \caption{Computational complexity of competitive methods on third-order tensors. $d=d_1d_2$ is the number of features. $t_i$ stands for the number of iterations. $c$ stands for the number of classes.}
    \label{Computational complexity}
    \begin{tabular}{ccc}
    \toprule
        Method &Input &Computational complexity \\
    \midrule   
        SOGFS\tablefootnote{Structured Optimal Graph for Unsupervised Feature Selection} \cite{SOGFS} &matrix &$O\left(dn^2+t(d^3t_1+n^2c)\right)$ \\
        \specialrule{0em}{1pt}{1pt}
        RNE\tablefootnote{Robust Neighborhood Embedding for unsupervised feature selection} \cite{RNE} &matrix &$O\left(dn^2+t(t_1(n+s)d^2+t_2n))\right)$ \\
        \specialrule{0em}{1pt}{1pt}    
        SPCAFS\tablefootnote{Sparse PCA via $\ell_{2,p}$-Norm regularization for unsupervised Feature Selection} \cite{SPCA-PSD} &matrix &$O\left(d^2n+d^3t\right)$ \\
        \specialrule{0em}{1pt}{1pt}
        NLGMS\tablefootnote{Unsupervised group feature selection approach via non-convex regularized graph embedding and self-representation}  \cite{NLGMS} &matrix &$O\left(t\left( d^3+n^3+nd+n^2d+d\log{d}\right)\right)$ \\
        \specialrule{0em}{1pt}{1pt}
        CPUFS\tablefootnote{CP decomposition based Unsupervised Feature Selection} \cite{CPUFS_2023} &tensor  &$O\left(d_1nc^2+d_2nc^2+d_1d_2nc+n^2c)t\right)$  \\
        \specialrule{0em}{1pt}{1pt}
        CPUFSnn\tablefootnote{CP decomposition based Unsupervised Feature Selection with non-negativity} \cite{CPUFS_2023} &tensor  &$O\left((d_1nc^2+d_2nc^2+d_1d_2nc+(n_1c)^2+(n_2c)^2+n^2c)t\right)$  \\
        \specialrule{0em}{1pt}{1pt}        
        MSPCA\tablefootnote{Multilinear Sparse Principal Component Analysis} \cite{MSPCA} &tensor  &$O\left(t_2\sum_{i=1}^{2}(t_1+n)d_i^3\right)$ \\
        \specialrule{0em}{1pt}{1pt}
        STPCA-MP (ours) &tensor &$O\left(\sum^{p}_{i=1}\left(q^2n+q^3t_i\right)+p^2q\right)(q\in\{d_1,d_2\})$ \\
    \bottomrule
    \end{tabular}

\end{table}

\subsection{Extension to higher-order tensor data}

By conducting the $\star_{\mathbf{M}}$-product in a recursive fashion, the proposed method can be extended to higher-order tensors. Given two $n$-th order tensors $\mathcal{A}\in\mathbb{C}^{d_1\times d_2 \cdots \times d_{n-1} \times d_n}$ and $\mathcal{B}\in\mathbb{C}^{d_2 \times h \cdots \times d_{n}}$, along with a series of invertible matrices $\mathbf{M}_i\in\mathbb{C}^{d_{i+2}\times d_{i+2}}$, the $\star_{\mathbf{M}}$-product for higher-order tensors is defined as shown in Algorithm \ref{M product for higher-order data}. Using this technique, STPCA-MP can be formulated for higher-order settings. However, sequential $\star_{M}$-products with different orientations introduce significant optimization. This remains an open problem under active investigation. Therefore, this paper focuses on the third-order case.

 \begin{algorithm}[t] 
\caption{$\star_{\mathbf{M}}$-product for higher-order tensor data} 
\label{M product for higher-order data}
\renewcommand{\algorithmicrequire}{\textbf{Input:}}
\renewcommand{\algorithmicensure}{\textbf{Output:}}
\begin{algorithmic}[1] \REQUIRE $\mathcal{A}\in\mathbb{C}^{d_1\times d_2 \cdots \times d_{n-1} \times d_n}$, $\mathcal{B}\in\mathbb{C}^{d_2 \times h \cdots \times d_{n}}$, $\mathbf{M}_i\in\mathbb{C}^{d_{i+2}\times d_{i+2}}$
\ENSURE $\mathcal{C}=\mathcal{A}\star_{M}\mathcal{B}$
\STATE Project the input data into the transform domain: $\hat{\mathcal{A}} = \mathcal{A}\times_3 \mathbf{M}_1 \times_4 \mathbf{M}_2 \cdots  \times_n \mathbf{M}_{n-2}$, $\hat{\mathcal{B}} = \mathcal{B}\times_3 \mathbf{M}_1 \times_4 \mathbf{M}_2 \cdots  \times_n \mathbf{M}_{n-2}$ 
\FOR{$i=1,2,\cdots,\prod\limits^{n}_{j=3}d_{j}$}
\STATE $\hat{\mathcal{C}}=\hat{\mathcal{A}}_{::i_3\cdots i_n}\hat{\mathcal{B}}_{::i_3\cdots i_n}$
\ENDFOR
\STATE Project the result back to the original domain $\mathcal{C} = \hat{\mathcal{C}}\times_n \mathbf{M}^{-1}_{n-2}\times_{n-1}\mathbf{M}^{-1}_{n-1} \cdots \times_{3}\mathbf{M}^{-1}_1$.
\STATE Output $\mathcal{C}$
\end{algorithmic}
\end{algorithm}

\section{Experiments}
In this section, we conduct experiments on real-world datasets to empirically validate the effectiveness of our proposed method. Implementation codes for STPCA-MP and all experiments are publicly hosted at: \url{https://github.com/zjj20212035/STPCA.git}.

\subsection{Experimental Settings}
\label{Experimental Settings}
\subsubsection{Experimental environment}
All experiments are executed on a standard workstation equipped with a 2.20 GHz CPU and 64 GB RAM, utilizing the Windows 11 operating system environment.

\subsubsection{Comparative methods}
We evaluate our method against representative state-of-the-art UFS baselines:

\begin{enumerate}[\textbullet]
    \item \textbf{CPUFS and CPUFSnn} \cite{CPUFS_2023} (\emph{2023}): \textbf{CPUFS} is a tensor-based unsupervised feature selection method that integrates CP decomposition with spectral clustering. The method employs a sparse classifier to perform feature selection. \textbf{CPUFSnn} extends this framework by imposing non-negative constraints on the classifier matrix, enhancing interpretability through element-wise positivity requirements.
    \item \textbf{MSPCA} \cite{MSPCA} (\emph{2014}): MSPCA represents a sparse tensor PCA approach employing Tucker decomposition to leverage inherent tensor structures. Originally developed as a feature extraction technique, we adapt it for unsupervised feature selection (UFS) by utilizing the Kronecker product of its factor matrices, thereby demonstrating a Tucker-based UFS implementation.
    \item \textbf{NLGMS} \cite{NLGMS} (\emph{2024}): NLGMS constitutes a non-tensor-based approach that incorporates a self-representation mechanism within a graph-based framework to simultaneously maintain both global and local data structures. The method's self-representation component reconstructs individual samples through linear combinations of other data instances.
    \item \textbf{SPCAFS} \cite{SPCAFS} (\emph{2021}): SPCAFS is a non-tensor-based approach that incorporates the $\ell_{2,p}$-norm regularization into the standard one-dimensional PCA framework. This formulation yields an efficient and computationally tractable model.
    \item \textbf{RNE} \cite{RNE} (\emph{2020}): RNE represents a non-tensor-based approach that employs the locally linear embedding (LLE) algorithm to derive an optimal projection matrix. The method maintains neighborhood relationships by preserving proximity between samples and their local neighbors when projected onto the manifold space. 
    \item \textbf{SOGFS} \cite{SOGFS} (\emph{2019}): SOGFS is a non-tensor-based approach that optimizes an adaptive graph structure to maintain the local geometric relationships within the data. The method iteratively refines the similarity matrix to achieve progressively more flexible graph representations. 

\end{enumerate}
For baseline comparison, we additionally conduct clustering using the complete set of original features. All competing methods are implemented through their respective authors' original code releases.

\subsubsection{Parameter settings}

All methods employ consistent experimental protocols: (1) Regularization parameters are determined through grid search over $\{10^{-2},10^{-1},1,10,10^2\}$; (2) Uniform feature dimensionality is maintained across comparative methods. Specific configurations include: (i) Graph-based methods use $k=5$ nearest neighbors; (ii) $\ell_{2,p}$-norm methods fix $p=1$; (iii) STPCA-MP initializes $\mathbf{M}=\mathbf{I}$ for standard experiments (excluding the dedicated $\mathbf{M}$ analysis); (iv) Direction set $D^o$ undergoes grid search in $\{{1,3,2\},\{2,3,1\}}$. Remaining parameters follow default settings from authors' original implementations. All reported results reflect optimal parameter configurations.

\subsubsection{Evaluation methodology and metrics}
Following feature selection by each UFS method, we assess clustering performance through a standardized evaluation pipeline:
\begin{enumerate}[1)]
    \item \textbf{Cluster Generation}: Apply $K$-means clustering to the selected features.
    \item \textbf{Label Alignment}: Establish correspondence between pseudo-labels and ground truth using the Kuhn-Munkres algorithm \cite{Kuhn-Munkres}.
    \item \textbf{Stability Control}: Repeat the clustering process 30 independent times and average the result to account for $K$-means' initialization sensitivity.
\end{enumerate}
The following clustering metrics are employed for quantitative evaluation:

\begin{enumerate}[\textbullet]
\item \textbf{Accuracy (ACC)}: The clustering accuracy (ACC) measures the agreement between cluster assignments and ground truth labels, computed as:
\begin{equation}\label{ACC}
\mathrm{ACC} = \frac{1}{n}\sum_{i=1}^{n}\delta(\mathrm{map}(i),\mathrm{label}(i))
\end{equation}
where $\delta(p,q)$ is the indicator function defined in (\ref{indicator}), $n$ denotes the total sample size, $\mathrm{map}(i)$ represents the mapped cluster label of the $i$-th sample, and $\mathrm{label}(i)$ indicates the ground truth label. Higher ACC values correspond to better clustering performance. The indicator function is given by:
\begin{equation}\label{indicator}
\delta(p,q) = \begin{cases} 
1 & \text{if } p = q \\ 
0 & \text{otherwise}
\end{cases}
\end{equation}

\item \textbf{Normalized Mutual Information (NMI) \cite{NMI}}: NMI is utilized to describe the mutual dependence between mapped clustering labels and ground truth labels. Given the clustering results $\mathbf{m}$ and the ground truth labels $\mathbf{l}$ (both including all samples), $I(\mathbf{m},\mathbf{l})$ denotes the mutual information between $\mathbf{m}$ and $\mathbf{l}$. $H(\mathbf{m})$ and $H(\mathbf{l})$ are the entropy of $\mathbf{m}$ and $\mathbf{l}$ respectively. A larger NMI means better clustering performance.
\begin{equation}
    \text{NMI}(\mathbf{m},\mathbf{l}) = \frac{I(\mathbf{m},\mathbf{l})}{\sqrt{H(\mathbf{m})H(\mathbf{l})}}.
\end{equation}

\item \textbf{Silhouette score (SS) \cite{rousseeuw_silhouettes_1987}}: SS is a widely-used metric for evaluating clustering quality that measures both intra-cluster compactness and inter-cluster separation. For a given sample $i$, the silhouette score $s(i)$ is computed as:
\begin{equation}
s(i) = \frac{b(i) - a(i)}{\max\{a(i), b(i)\}},
\end{equation}
where $a(i)$ is the average distance between sample $i$ and all other points in the same cluster (intra-cluster distance), and $b(i)$ is the smallest average distance between $i$ and points in any other cluster (neighboring cluster distance). The global silhouette score $\text{SS}$ for the entire dataset ranges from $-1$ to $1$, with values closer to $1$ indicating better-defined clusters:
\begin{equation}
\text{SS} = \frac{1}{N}\sum_{i=1}^{N} s(i).
\end{equation}

\item \textbf{Training time}: For an algorithm, training time starts from the initialization and ends when the algorithm outputs the selected features. 
\end{enumerate}

For orientation-specific data (e.g., multi-channel time series), the temporal dependencies within channels violate the feature independence assumption underlying $K$-means clustering. This fundamental mismatch renders standard clustering metrics unreliable. We therefore propose the Proportion of Correctly selected features (POC) as an alternative evaluation metric:
\begin{enumerate}[\textbullet]
    \item \textbf{Proportion of correctly selected features (POC)}: We compute POC as:
    \begin{equation}
    \label{POC}
        \text{POC} = \frac{c}{h}, 
    \end{equation}
where $c$ denotes the number of features selected by both the unsupervised feature selection (UFS) method and the maximal between-class variance (BCV) criterion, and $h$ represents the total features selected per execution. 
Suppose data has been centralized, the BCV for the $(i,j)$-th element in a frontal slice is defined as:
\begin{equation}\label{BCV_formulation}
    \text{BCV}(i,j) = \frac{1}{K} \sum_{k=1}^{K} \left( \mu^{(k)}_{ij} - \mu_{ij} \right)^2, \quad i=1,\dots,d_1;\ j=1,\dots,d_2
\end{equation}
where $\mu_{ij}$ is the sample mean of the $(i,j)$-th element across all samples, $\mu^{(k)}_{ij}$ is the sample mean within class $k$, and $K$ is the number of classes. Here, $d_1$ and $d_2$ denote the dimensionalities of the first and second modes, respectively.
For slice-wise data where each channel is treated as a feature (See Table \ref{Two scenarios for third-order data tensors} and Section \ref{Datasets introduction}), the BCV for the $f$-th channel is obtained by aggregating along the temporal dimension:
\begin{equation}\label{BCV_feature}
    \text{BCV}_f = \sum_{j=1}^{d_2} \text{BCV}(i,j) \quad \text{for fixed channel index } i=f.
\end{equation}
This formulation quantifies the cumulative class-discriminative power of the entire time series in channel $f$. The POC metric thereby directly evaluates a method's capability to identify discriminative features where conventional clustering metrics fail.
\end{enumerate}

\subsection{Datasets}
\label{Datasets introduction}
Before delving into specific datasets, we first categorize tensor datasets based on their structural organization. As illustrated in Table \ref{Two scenarios for third-order data tensors}, third-order tensor datasets can be classified into two scenarios: tube-wise and slice-wise, according to their organizational patterns. Slice-wise data is orientation-specific, exhibiting feature correlations along the 1-mode, whereas tube-wise data demonstrates feature correlations along both the 1-mode and 2-mode. This distinction serves as a foundation for evaluating the capability of UFS methods to leverage tensor structural information. In our evaluation, we compare ACC, NMI, and SS for tube-wise data, and use the POC for slice-wise data. Experimental results validate the effectiveness of the proposed method in both scenarios.

\begin{table}
    \caption{Two scenarios for third-order data tensors. Each frontal slice corresponds to a distinct sample. Variables (represented by distinct colors) are treated as features. This study adopts two feature interpretation scenarios. \textbf{Tube-wise:} Elements within a single fiber constitute one feature variable. \textbf{Slice-wise:} Elements within a horizontal slice comprise one feature variable.}
    \centering
    \renewcommand\arraystretch{1.1}
        \begin{tabular}{|c|c|l|}
         \hline
         Scenario  &Visualization   & Example   \\
         \hline
         \multirow{1}{*}[5ex]{Tube-wise} & \centering{\includegraphics[width=0.8in]{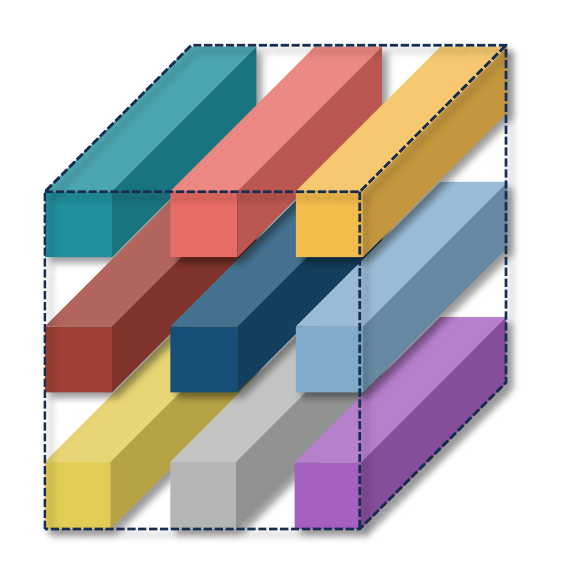}} &\multirow{1}{*}[6.5ex]{\makecell[l]{1. snapshots of array signal \cite{HOSVD_MIMO_Coarrays_AE, HOSVD_BlindDOA, gong_esprit-based_2023}.\\
         2. images and video\cite{HOSVD_Dynamic_Texture, HOSVD, 
PIE}.}}\\ 
         \hline
         \multirow{1}{*}[5ex]{\makecell{Slice-wise\\(orientation-specific)}} &\centering{\includegraphics[width=0.8in]{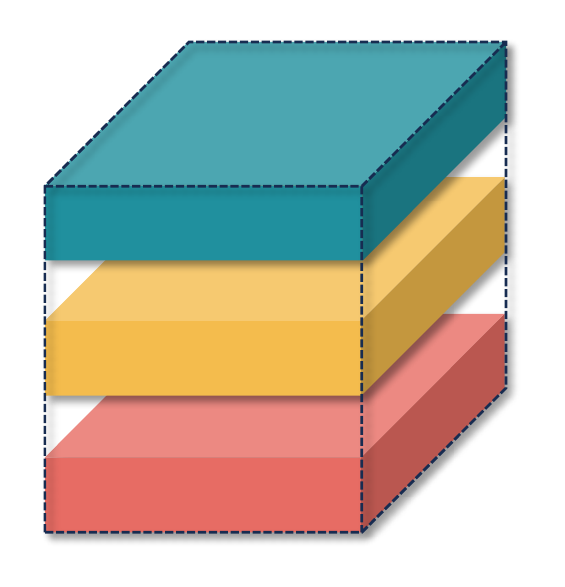}} &\multirow{1}{*}[6.5ex]{\makecell[l]{1. time series with different channels \cite{HOSVD_LCA_bio,zhang_automated_2023,cong_tensor_2015,HOSVD_Time_Series_Eco}.\\
         2. multi-way statistical tables \cite{levin_three-mode_1965}.}} \\
         \hline
    \end{tabular}

    \label{Two scenarios for third-order data tensors}  
\end{table}

To preliminarily evaluate the UFS ability of our proposed method in the two tensor scenarios, we design two synthetic tensor datasets: 

\begin{enumerate}[\textbullet]
    \item Orbit (Figure \ref{Synthetic experiments}(a)): As an example of slice-wise data, Orbit comprises 3-dimensional orbit samples with two distinct radii. Each sample includes three channels representing orbit coordinates, with the remaining channels containing random Gaussian noise.
    \item Array (Figure \ref{Synthetic experiments}(f)): As an example of tube-wise data, Array is constructed using a $10 \times 10$ synthetic radar unit array that receives a series of echoes from a stationary target. Each echo is represented as a matrix of unit sampling values, treated as a sample. Random measurement errors are introduced to half of the echo matrices to create two distinct classes. 
\end{enumerate}

We evaluate our method on nine real-world datasets, including: 1) six tube-wise datasets: COIL20 \cite{COIL20}, USPS \cite{USPS}, BreastMNIST \cite{yang_medmnist_2021}, Imm40 \cite{Imm40}, PIE \cite{PIE} and JAFFE \cite{jaffe}; 2) four slice-wise datasets: UCIHAR \cite{UCIHAR} and UCIDSA \cite{misc_daily_and_sports_activities_256}, UMCS\footnote{https://www.kaggle.com/c/seizure-detection/data\label{UMCS}}, and ASCH \cite{ASCH}. The detailed information of each dataset is shown in Table \ref{Statistics of real-world Dataset}. PIE, Imm40, and JAFFE contain frontal facial images, labeled according to individuals. COIL20 comprises images of various objects captured from multiple viewing angles. BreastMNIST is a medical dataset containing breast ultrasound images, from which we construct the data tensor using the test set. USPS is a dataset of handwritten digits. UCIHAR and UCIDSA are human activity datasets collected from multiple subjects using motion sensors. UCIDSA contains multi-channel signals, while UCIHAR comprises extracted statistical features (\emph{e.g.} mean, std, max, entropy). UMCS and ASCH are electroencephalogram (EEG) datasets collected for disease research, capturing brain activity signals. Due to the large volume of chosen slice-wise data, we retrieve a subset of samples to construct datasets. For UCIDSA, we extract the first $125$ sample points from all $45$ channels covering $19$ activities. For UCIHAR, we select $9$ individuals with all $561$ features sampled during $6$ activities. For UMCS, we collect the first $125$ sample points in all $65$ channels of the first human patient, classified into ictal and interictal segments. For ASCH, we collect the first $640$ sample points in all $16$ channels from healthy groups and groups with symptoms. All datasets are normalized and bounded within $[-1,1]$ for consistency. All the above data preprocessing is consistent across all comparative methods, including ours.

\begin{table}[!t]
\caption{Statistics of datasets.} 
\label{Statistics of real-world Dataset}
\centering
\renewcommand\arraystretch{1.2}
\setlength{\tabcolsep}{0.7mm}
\begin{tabular}{ccccc}

\toprule
  \textbf{Type} &\textbf{Dataset} & \textbf{\#Feature}  & \textbf{\#Samples} &\textbf{\#Classes}
  \\
 \midrule
   \multirow{7}{*}{\centering{Tube-wise}}
   &Array &$10\times10$ &100 &2 \\
   &USPS \cite{USPS} &$16\times16$  &9298 &10 \\
   &BreastMNIST \cite{yang_medmnist_2021}  &$28\times28$  &156 &2 \\
   &Imm40 \cite{Imm40} &$32\times32$  &240  &40\\
   &COIL20 \cite{COIL20}    &$32\times32$ &1440 &20 \\
   &PIE \cite{PIE}    &$32\times32$ &1166 &53 \\
    &JAFFE \cite{jaffe}     &$64\times64$  &213 &7 \\
 \midrule  
    \multirow{5}{*}{Slice-wise}
    &Orbit &$9\times41$  &100 &2 \\
    
    &UCIDSA \cite{misc_daily_and_sports_activities_256} & $45\times125$  &152 &19 \\
    
    &UCIHAR \cite{UCIHAR}   &$561\times9$  &273 &6 \\


   &UMCS \footref{UMCS}  &$65\times125$  &174 &2 \\
   

   &ASCH \cite{ASCH}&$16\times640$  &84 &2 \\
    
 \bottomrule
\end{tabular}
\end{table}

\subsection{Synthetic experiment}
\begin{figure}
    \centering
    \includegraphics[width=1\linewidth]{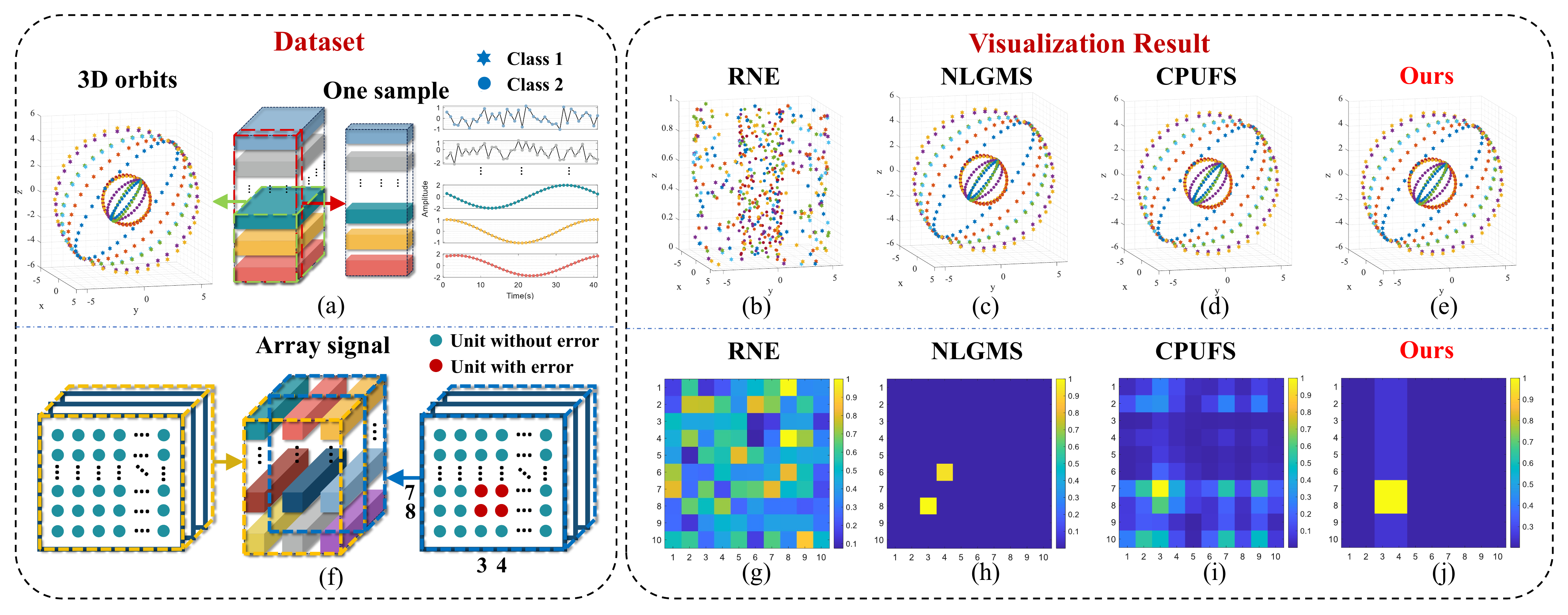}
    \caption{Synthetic experiments on Orbit and Array. The left part gives the composition of the two synthetic data tensors. The right part shows the feature selection results of comparative methods in a visualized manner.}
    \label{Synthetic experiments}
\end{figure}

In this experiment, we apply four typical UFS methods to synthetic datasets using the grid search strategy. The dataset construction and the optimal experimental results of the comparative methods are presented in Figure \ref{Synthetic experiments}.

For the Orbit dataset (Figure \ref{Synthetic experiments}(a)), we retrieve the three top-ranked features from the results of each method and plot their corresponding samples within a data subset. If a method successfully selects the correct features (i.e., the three orbit coordinates), the scatter plot will display rings with two distinct radii. Conversely, when the wrong features are chosen, distinguishing between the two classes becomes challenging. 

Regarding the Array dataset (Figure \ref{Synthetic experiments}(f)), we retrieve the feature score map associated with the best POC for each method. Then, we analyze the degree of match between the positions of the top-ranked features and the units with measurement errors. 

Considering the experimental results on both synthetic datasets, the key observations are summarized as follows:

\begin{enumerate}[\textbullet]
\item \emph{Tensor-based methods generally outperform non-tensor-based methods in analyzing data structure.} Both CPUFS and STPCA-MP correctly identify the orbit coordinates on the Orbit dataset (Figure \ref{Synthetic experiments}(d)(e)). Although CPUFS fails to precisely pinpoint the error units on the Array dataset, its score map exhibits a distinct grid-like pattern (Figure \ref{Synthetic experiments}(i)), which captures the row-column feature relationships and identifies an approximately correct region of interest. Similarly, STPCA-MP learns the column-wise relationships (Figure \ref{Synthetic experiments}(j)). In contrast, RNE and NLGMS treat features in isolation, thus losing the valuable tensor structure information. Notably, for RNE in Figure \ref{Synthetic experiments}(g), the scores are seemingly randomly distributed.

\item \emph{Our proposed method excels in selecting discriminative features on both synthetic datasets.} STPCA-MP selects the correct features on both datasets (Figure \ref{Synthetic experiments}(e)(j)). Specifically, the score map for the Array dataset (Figure \ref{Synthetic experiments}(j)) shows relatively sparse rows and columns. By leveraging the data's orientation, STPCA-MP can more effectively uncover latent feature relationships and identify discriminative features.
\end{enumerate}

\subsection{Clustering experiment on tube-wise data}
\begin{figure*}
    \centering\includegraphics[width=6.7in]{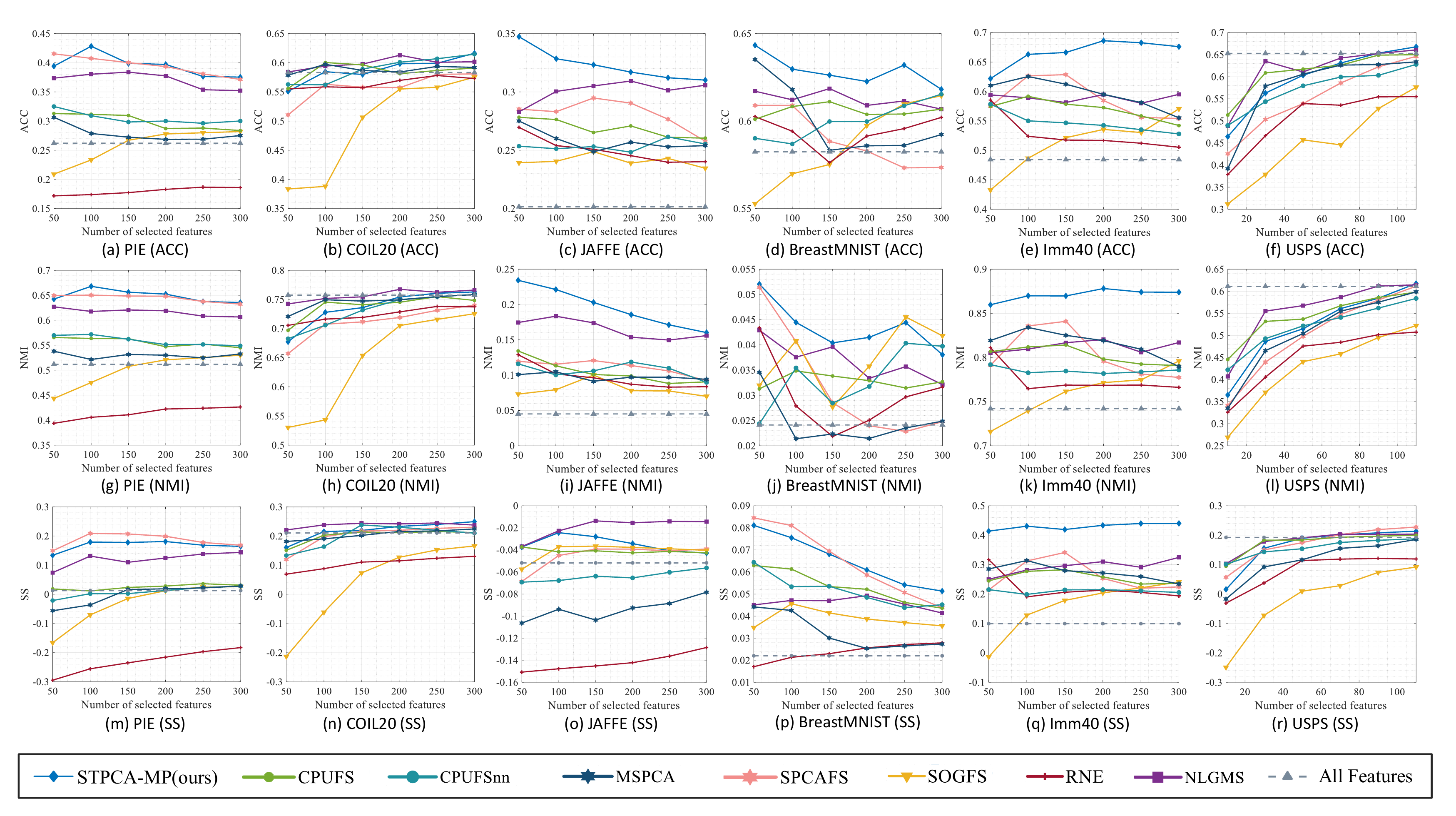}
    \caption{The ACC, NMI, and SS curves on tube-wise datasets. The upper, middle, and bottom row displays the ACC, NMI, and SS curves, respectively.}
\label{ACC_NMI_SS}
\end{figure*}

In this experiment, we perform clustering on five tube-wise real-world datasets. The regularization parameters of all methods go through a grid search in $\{10^{-2},10^{-1},1, 10^{1}, 10^{2}\}$. We set the selected features to $\{50,100,150,200,250,300\}$ on COIL20, JAFFE, BreastMNIST , Imm40, and $\{10,30,50,70,90,110\}$ on USPS. The best results and the corresponding training time of all methods are given in Table \ref{Metrics of the comparative methods on real-world data}. The detailed ACC, NMI, and SS curves are illustrated in Figure \ref{ACC_NMI_SS}. From these results, we draw three main conclusions: 
\begin{enumerate}[\textbullet]

    \item \emph{Features selection enhances data compactness}. Multiple methods achieve performance comparable to or superior to the baseline while utilizing only a subset of features. Notably, STPCA-MP significantly outperforms All Features on PIE (Figure \ref{ACC_NMI_SS}(a)(g)(m)), Jaffe (Figure \ref{ACC_NMI_SS}(c)(i)(o)), BreastMNIST (Figure \ref{ACC_NMI_SS}(d)(j)(p)), and Imm40 (Figure \ref{ACC_NMI_SS}(e)(k)(q)). 
    \item \emph{STPCA-MP demonstrates reliable and superior clustering performance}. The proposed method captures more global data information through its reconstruction term, outperforming graph-based approaches like SOGFS, CPUFS, and CPUFSnn. This is evidenced by: (i) smoother performance curves with smaller amplitude oscillations (\emph{e.g.} Figure \ref{ACC_NMI_SS}(e)(k)(q)), and (ii) more consistent trends across ACC, NMI, and SS metrics (Figure \ref{ACC_NMI_SS}(c)(i)(o)). Furthermore, STPCA-MP's T-SVDM-based framework effectively extracts tensor structural information, enabling more precise feature correlation modeling than: (a) CPD-based methods (CPUFS, CPUFSnn), (b) Tucker-based MSPCA, and (c) non-tensor methods (SPCAFS, RNE) on datasets with relatively strict feature spatial distributions (PIE, JAFFE, and Imm40). On datasets with irregular feature space distributions (COIL20 and USPS), STPCA-MP does not demonstrate a significant advantage in average performance, yet maintains top-ranked results across all tested feature selection dimensions (Figure \ref{ACC_NMI_SS}(b)(h)(n)) and ultimately attains the optimal results (Table \ref{Metrics of the comparative methods on real-world data}).
    \item \emph{STPCA-MP combines high performance with computational efficiency}. As shown in Table \ref{Metrics of the comparative methods on real-world data}, STPCA-MP completes training and feature selection in under 1 second per dataset through its slice-by-slice operations. This efficiency is achieved while maintaining proper data structure awareness. In contrast, NLGMS and SPCAFS incur substantially higher computational costs due to their vectorization approach, despite showing competitive accuracy.   

\end{enumerate}

\begin{table*}[!t]
\caption{Metrics of the comparative methods on real-world datasets. The best ACC ($\%$), best NMI ($\%$), best SS (between $-1$ and $1$), and mean POC ($\%$) regarding the grid-search range are shown. The training time corresponding to the best ACC (or POC) is recorded in seconds. The best result on each dataset is boldfaced, while the second-best one is underlined and italic.} \label{Metrics of the comparative methods on real-world data}
\centering
\resizebox{\linewidth}{!}{
\renewcommand\arraystretch{1.3}
\begin{tabular}{|c|c|c|c|c|c|c|c|c|c|c|c|c|c|c|c|c|c|c|c|c|}
\specialrule{0em}{1pt}{5pt}
\hline
\textbf{Method} &\multicolumn{2}{c|}{\textbf{PIE}}&\multicolumn{2}{c|}{\textbf{COIL20}} &\multicolumn{2}{c|}{\textbf{JAFFE}} &\multicolumn{2}{c|}{\textbf{BreastMNIST}} &\multicolumn{2}{c|}{\textbf{Imm40}} &\multicolumn{2}{c|}{\textbf{USPS}} &\multicolumn{2}{c|}{\textbf{UCIDSA}} &\multicolumn{2}{c|}{\textbf{UCIHAR}} &\multicolumn{2}{c|}{\textbf{UMCS}} &\multicolumn{2}{c|}{\textbf{ASCH}} \\
\hline
\multirow{3}{*}{All Features} &ACC &26.21  &ACC &58.34 &ACC &20.16  &ACC &58.25  &ACC &48.43  &ACC &65.29 &\multirow{3}{*}{POC} &\multirow{3}{*}{-} 
&\multirow{3}{*}{POC} &\multirow{3}{*}{-} 
&\multirow{3}{*}{POC} &\multirow{3}{*}{-}
&\multirow{3}{*}{POC} &\multirow{3}{*}{-} \\
&NMI &51.19 &NMI &75.74 &NMI&4.51  &NMI &2.41 &NMI &74.21 &NMI&61.11 & & & & & & & &\\
&SS &0.0127  &SS &0.2105  &SS &-0.0518  &SS &0.0220  &SS &0.0994  &SS &0.1924    & & & & & & & &\\
\hline
\multirow{4}{*}{RNE\cite{RNE}} &ACC &18.67  &ACC &57.83 &ACC &26.96  &ACC &60.26  &ACC &58.54 &ACC &55.51 &\multirow{4}{*}{\makecell{POC\\\specialrule{0em}{0.2pt}{0pt}\\TIME}}  
&\multirow{4}{*}{\makecell{40.00\\\specialrule{0em}{0.2pt}{0pt}\\83.57}} 
&\multirow{4}{*}{\makecell{POC\\\specialrule{0em}{0.2pt}{0pt}\\TIME}} 
&\multirow{4}{*}{\makecell{0\\\specialrule{0em}{0.2pt}{0pt}\\170.22}} 
&\multirow{4}{*}{\makecell{POC\\\specialrule{0em}{0.2pt}{0pt}\\TIME}} 
&\multirow{4}{*}{\makecell{\emph{\uline{40.00}}\\\specialrule{0em}{0.2pt}{0pt}\\193.22}}
&\multirow{4}{*}{\makecell{POC\\\specialrule{0em}{0.2pt}{0pt}\\TIME}}
&\multirow{4}{*}{\makecell{66.67\\\specialrule{0em}{0.2pt}{0pt}\\205.33}}\\
&NMI &42.64  &NMI &73.81 &NMI &12.89  &NMI &4.34 &NMI &81.11 &NMI&50.74 & & & & & & & &\\
&SS &-0.1828  &SS &0.1304  &SS &-0.1284  &SS &0.0279  &SS &0.3168  &SS &0.1211    & & & & & & & &\\
&TIME &5.56  &TIME &18.17 &TIME&63.42  &TIME &8.16 &TIME &15.44 &TIME&1.60 & & & & & & & & \\
\hline
\multirow{4}{*}{\makecell{SOGFS\cite{SOGFS}}} &ACC &28.22  &ACC &57.54 &ACC &24.88 &ACC &61.45 &ACC &57.06 &ACC &57.63 &\multirow{3}{*}{\makecell{POC\\\specialrule{0em}{0.2pt}{0pt}\\TIME}} 
&\multirow{4}{*}{\makecell{37.33\\\specialrule{0em}{0.2pt}{0pt}\\2409.95}} 
&\multirow{4}{*}{\makecell{POC\\\specialrule{0em}{0.2pt}{0pt}\\TIME}} 
&\multirow{4}{*}{\makecell{1.60\\\specialrule{0em}{0.2pt}{0pt}\\5569.15}}
&\multirow{4}{*}{\makecell{POC\\\specialrule{0em}{0.2pt}{0pt}\\TIME}} 
&\multirow{4}{*}{\makecell{17.33\\\specialrule{0em}{0.2pt}{0pt}\\5274.80}} 
&\multirow{4}{*}{\makecell{POC\\\specialrule{0em}{0.2pt}{0pt}\\TIME}} 
&\multirow{4}{*}{\makecell{23.33\\\specialrule{0em}{0.2pt}{0pt}\\8986.97}}\\
&NMI &53.01  &NMI &72.55 &NMI&10.01  &NMI &4.55 &NMI &79.60 &NMI&52.21 & & & & & & & &\\
&SS &0.0263  &SS &0.1661  &SS &-0.0366  &SS &0.0456  &SS &0.2411  &SS &0.0916    & & & & & & & &\\
&TIME &34.28  &TIME &55.56
&TIME &1794.41  &TIME &37.12 &TIME &83.77 &TIME&391.60 & & & & & & & & \\
\hline
\multirow{4}{*}{\makecell{SPCAFS\cite{SPCAFS}}} 
&ACC &\emph{\uline{41.55}}  &ACC &58.10&ACC &29.48  &ACC &60.90 &ACC &\emph{\uline{62.86}}  &ACC &64.63 &\multirow{4}{*}{\makecell{POC\\\specialrule{0em}{0.2pt}{0pt}\\TIME}}  
&\multirow{4}{*}{\makecell{70.67\\\specialrule{0em}{0.2pt}{0pt}\\140.61}} 
&\multirow{4}{*}{\makecell{POC\\\specialrule{0em}{0.2pt}{0pt}\\TIME}}  
&\multirow{4}{*}{\makecell{23.60\\\specialrule{0em}{0.2pt}{0pt}\\274.54}}
&\multirow{4}{*}{\makecell{POC\\\specialrule{0em}{0.2pt}{0pt}\\TIME}}  
&\multirow{4}{*}{\makecell{28.00\\\specialrule{0em}{0.2pt}{0pt}\\435.80}} 
&\multirow{4}{*}{\makecell{POC\\\specialrule{0em}{0.2pt}{0pt}\\TIME}}  
&\multirow{4}{*}{\makecell{\emph{\uline{70.00}}\\\specialrule{0em}{0.2pt}{0pt}\\6569.55}}\\
&NMI &\emph{\uline{65.04}}  &NMI &74.02 &NMI&12.07  &NMI &\emph{\uline{5.14}} &NMI &\emph{\uline{84.09}} &NMI &61.13 & & & & & & & &\\
&SS &\textbf{0.2091}  &SS &0.2303  &SS &-0.0392  &SS &\textbf{0.0845}  &SS &\emph{\uline{0.3413}}  &SS &\textbf{0.2275}    & & & & & & & &\\
&TIME &18.37  &TIME &3.11 &TIME &59.36  &TIME &0.61 &TIME &15.76 &TIME&0.34 & & & & & & & & \\
\hline
\multirow{4}{*}{\makecell{NLGMS\cite{NLGMS}}} 
&ACC &38.40  &ACC &61.30  &ACC &\emph{\uline{30.94}}  &ACC &61.86 &ACC &59.51  &ACC &\emph{\uline{66.08}}
&\multirow{4}{*}{\makecell{POC\\\specialrule{0em}{0.2pt}{0pt}\\TIME}}  
&\multirow{4}{*}{\makecell{40.53\\\specialrule{0em}{0.2pt}{0pt}\\222.98}} 
&\multirow{4}{*}{\makecell{POC\\\specialrule{0em}{0.2pt}{0pt}\\TIME}}  
&\multirow{4}{*}{\makecell{17.08\\\specialrule{0em}{0.2pt}{0pt}\\410.78}}
&\multirow{4}{*}{\makecell{POC\\\specialrule{0em}{0.2pt}{0pt}\\TIME}}  
&\multirow{4}{*}{\makecell{21.87\\\specialrule{0em}{0.2pt}{0pt}\\782.96}} 
&\multirow{4}{*}{\makecell{POC\\\specialrule{0em}{0.2pt}{0pt}\\TIME}}  
&\multirow{4}{*}{\makecell{46.00\\\specialrule{0em}{0.2pt}{0pt}\\3114.56}}\\
&NMI &62.69  &NMI &\textbf{76.74} &NMI&\emph{\uline{18.34}}  &NMI &4.29 &NMI &82.03 &NMI &\emph{\uline{61.45}} & & & & & & & &\\
&SS &0.1439  &SS &\emph{\uline{0.2449}}  &SS &\textbf{-0.0137}  &SS &0.0493  &SS &0.3245  &SS &0.2035    & & & & & & & &\\
&TIME &25.99  &TIME &25.71 &TIME &498.12  &TIME &2.98 &TIME &19.08 &TIME &511.68 & & & & & & & & \\
\hline
\multirow{4}{*}{\makecell{CPUFS\cite{CPUFS_2023}}} &ACC &31.30  &ACC &60.05 &ACC &27.82 &ACC &61.11 
 &ACC &59.18 &ACC &64.99  
 &\multirow{4}{*}{\makecell{POC\\\specialrule{0em}{0.2pt}{0pt}\\TIME}} 
 &\multirow{4}{*}{\makecell{34.35\\\specialrule{0em}{0.2pt}{0pt}\\51.01}} 
 &\multirow{4}{*}{\makecell{POC\\\specialrule{0em}{0.2pt}{0pt}\\TIME}}  
 &\multirow{4}{*}{\makecell{19.16\\\specialrule{0em}{0.2pt}{0pt}\\71.78}} 
 &\multirow{4}{*}{\makecell{POC\\\specialrule{0em}{0.2pt}{0pt}\\TIME}} 
 &\multirow{4}{*}{\makecell{21.49\\\specialrule{0em}{0.2pt}{0pt}\\11.47}} 
 &\multirow{4}{*}{\makecell{POC\\\specialrule{0em}{0.2pt}{0pt}\\TIME}}  
 &\multirow{4}{*}{\makecell{34.93\\\specialrule{0em}{0.2pt}{0pt}\\5.91}}\\
&NMI &56.56  &NMI &75.48 &NMI&13.42 &NMI &3.48 &NMI &81.41 &NMI&59.79 & & & & & & & &\\
&SS &0.0364  &SS &0.2159  &SS &-0.0374  &SS &0.0629  &SS &0.2825  &SS &0.1999    & & & & & & & &\\
&TIME &290.69  &TIME &69.89 &TIME &11.20  &TIME &6.87 &TIME &14.01 &TIME&1559.53 & & & & & & & & \\
\hline
\multirow{4}{*}{\makecell{CPUFSnn\cite{CPUFS_2023}}} &ACC &32.51  &ACC &\emph{\uline{61.47}} &ACC &26.15 &ACC &61.54 
 &ACC &57.85 &ACC &62.79 
 &\multirow{4}{*}{\makecell{POC\\\specialrule{0em}{0.2pt}{0pt}\\TIME}} 
 &\multirow{4}{*}{\makecell{34.40\\\specialrule{0em}{0.2pt}{0pt}\\48.77}} 
 &\multirow{4}{*}{\makecell{POC\\\specialrule{0em}{0.2pt}{0pt}\\TIME}} 
 &\multirow{4}{*}{\makecell{23.41\\\specialrule{0em}{0.2pt}{0pt}\\17.79}}
 &\multirow{4}{*}{\makecell{POC\\\specialrule{0em}{0.2pt}{0pt}\\TIME}} 
 &\multirow{4}{*}{\makecell{23.75\\\specialrule{0em}{0.2pt}{0pt}\\12.66}} 
 &\multirow{4}{*}{\makecell{POC\\\specialrule{0em}{0.2pt}{0pt}\\TIME}}  
 &\multirow{4}{*}{\makecell{34.27\\\specialrule{0em}{0.2pt}{0pt}\\6.67}}\\
&NMI &32.51  &NMI &75.85 &NMI&11.86  &NMI &4.03 &NMI &79.18 &NMI&58.37 & & & & & & & &\\
&SS &0.0279  &SS &0.2380  &SS &-0.0563  &SS &0.0644  &SS &0.2150  &SS &0.1879    & & & & & & & &\\
&TIME &101.76  &TIME &20.72 &TIME&11.68  &TIME &7.11 &TIME &13.35 &TIME&485.50 & & & & & & & & \\
\hline
\multirow{4}{*}{\makecell{MSPCA\cite{MSPCA}}} &ACC &30.66  &ACC &59.74 &ACC &27.51 &ACC &\emph{\uline{63.53}} &ACC &62.54 &ACC &63.37  
&\multirow{4}{*}{\makecell{POC\\\specialrule{0em}{0.2pt}{0pt}\\TIME}}  
&\multirow{4}{*}{\makecell{\emph{\uline{86.67}}\\\specialrule{0em}{0.2pt}{0pt}\\0.37}} 
&\multirow{4}{*}{\makecell{POC\\\specialrule{0em}{0.2pt}{0pt}\\TIME}}  
&\multirow{4}{*}{\makecell{\emph{\uline{24.20}}\\\specialrule{0em}{0.2pt}{0pt}\\3.22}}
&\multirow{4}{*}{\makecell{POC\\\specialrule{0em}{0.2pt}{0pt}\\TIME}}  
&\multirow{4}{*}{\makecell{30.80\\\specialrule{0em}{0.2pt}{0pt}\\0.02}} 
&\multirow{4}{*}{\makecell{POC\\\specialrule{0em}{0.2pt}{0pt}\\TIME}} 
&\multirow{4}{*}{\makecell{69.00\\\specialrule{0em}{0.2pt}{0pt}\\0.15}}\\
&NMI &53.81  &NMI &75.84 &NMI&10.47  &NMI &3.46 &NMI &83.40 &NMI&59.86 & & & & & & & &\\
&SS &0.0284  &SS &0.2244  &SS &-0.0784  &SS &0.0442  &SS &0.3139  &SS &0.1853    & & & & & & & &\\
&TIME &0.33  &TIME &0.02 &TIME&0.01  &TIME &0.01 &TIME &0.07 &TIME&0.05 & & & & & & & & \\
\hline
\multirow{4}{*}{\makecell{STPCA-MP (ours)}} 
&ACC &\textbf{42.84}  &ACC &\textbf{61.64} &ACC &\textbf{34.74}  &ACC &\bf{64.34}  &ACC &\textbf{68.92} &ACC &\textbf{66.77} 
&\multirow{4}{*}{\makecell{POC\\\specialrule{0em}{0.2pt}{0pt}\\TIME}}  
&\multirow{4}{*}{\makecell{\textbf{92.27}\\\specialrule{0em}{0.2pt}{0pt}\\0.08}} 
&\multirow{4}{*}{\makecell{POC\\\specialrule{0em}{0.2pt}{0pt}\\TIME}}  
&\multirow{4}{*}{\makecell{\textbf{35.20}\\\specialrule{0em}{0.2pt}{0pt}\\0.17}} 
&\multirow{4}{*}{\makecell{POC\\\specialrule{0em}{0.2pt}{0pt}\\TIME}}  
&\multirow{4}{*}{\makecell{\textbf{41.07}\\\specialrule{0em}{0.2pt}{0pt}\\0.04}} 
&\multirow{4}{*}{\makecell{POC\\\specialrule{0em}{0.2pt}{0pt}\\TIME}}  
&\multirow{4}{*}
{\makecell{\textbf{83.33}\\\specialrule{0em}{0.2pt}{0pt}\\0.94}}\\
&NMI &\textbf{66.80}  &NMI &\emph{\uline{76.29}} &NMI&\bf{23.42}  &NMI &\bf{5.20} &NMI &\textbf{87.81} &NMI&\textbf{61.75} & & & & & & & &\\
&SS &\emph{\uline{0.1812}}  &SS &\textbf{0.2494}  &SS &\emph{\uline{-0.0244}}  &SS &\emph{\uline{0.0811}}  &SS &\textbf{0.4400}  &SS &\emph{\uline{0.2141}}    & & & & & & & &\\
&TIME &0.03  &TIME &0.04 &TIME&0.05  &TIME &0.02 &TIME &0.02 &TIME&0.06 & & & & & & & & \\
\hline

\end{tabular}
}
\end{table*}

\subsection{POC comparison experiment on slice-wise data}
For slice-wise datasets where each channel (row vector) is treated as a feature, the experiment proceeds through two key phases: (i) Feature Scoring: executing the UFS algorithm to generate element-wise importance scores, followed by (ii) Feature Evaluation: aggregating scores row-wise for channel-level rankings, computing BCV on zero-centered data, and computing POC by comparing top-ranked features with maximal BCV features. The number of selected features is set to $15$, $100$, $15$, and $6$ on UCIDSA, UCIHAR, UMCS, and ASCH, respectively.

\begin{figure*}
    \hspace{-5em}\includegraphics[width=6.8in]{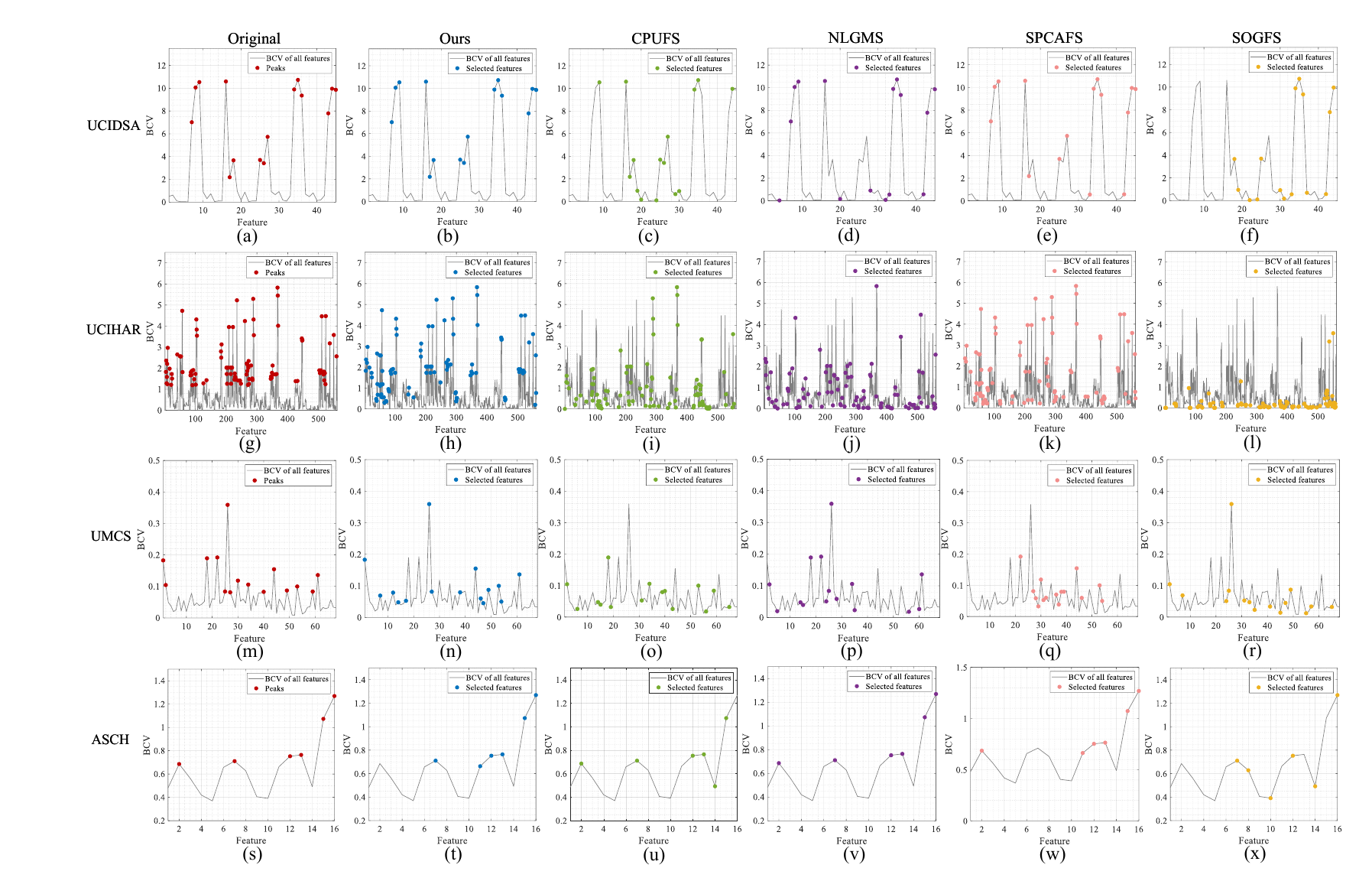}
    \caption{The BCV of selected features on slice-wise datasets.}   
\label{BCV}
\end{figure*}

The BCV distribution of selected features across four slice-wise datasets is visualized in Figure \ref{BCV}, displaying results corresponding to each method's maximal POC performance. Mean POC values are quantitatively compared in Table \ref{Metrics of the comparative methods on real-world data}. We have the following observations:

\begin{enumerate}[\textbullet]

    \item  Each dataset exhibits a subset of features with significantly higher BCV, as evidenced by distinct peaks in Figure \ref{BCV}(a)(g)(m)(s). These peaks correlate strongly with inter-class discriminative power.
    \item \emph{SPCA-based methods demonstrate superior BCV feature selection capability}. SPCAFS, MSPCA, and STPCA-MP maintain stable POC performance through their shared encoder-decoder framework, which simultaneously minimizes reconstruction error and maximizes global sample variance. Notably, MSPCA's sparse tensor decomposition contributes to its competitive performance on slice-wise data, as confirmed by its mean POC in Table \ref{Metrics of the comparative methods on real-world data}.
    \item \emph{STPCA-MP outperforms competitors in slice-wise data processing}. By explicitly modeling data orientation, STPCA-MP effectively captures cross-channel relationships within slice-wise tensors, consistently selecting BCV peaks, as shown in Figure \ref{BCV}(b)(h)(n)(t). While NLGMS achieves comparable peak performance on certain datasets, STPCA-MP shows superior stability in discriminative feature selection across the full parameter range, as illustrated in Table \ref{Metrics of the comparative methods on real-world data}, establishing it as the most effective method for slice-wise data analysis.
    
\end{enumerate}

\subsection{Visualization}
\begin{figure}
    \hspace{-1em}\centering\includegraphics[width=6.6in]{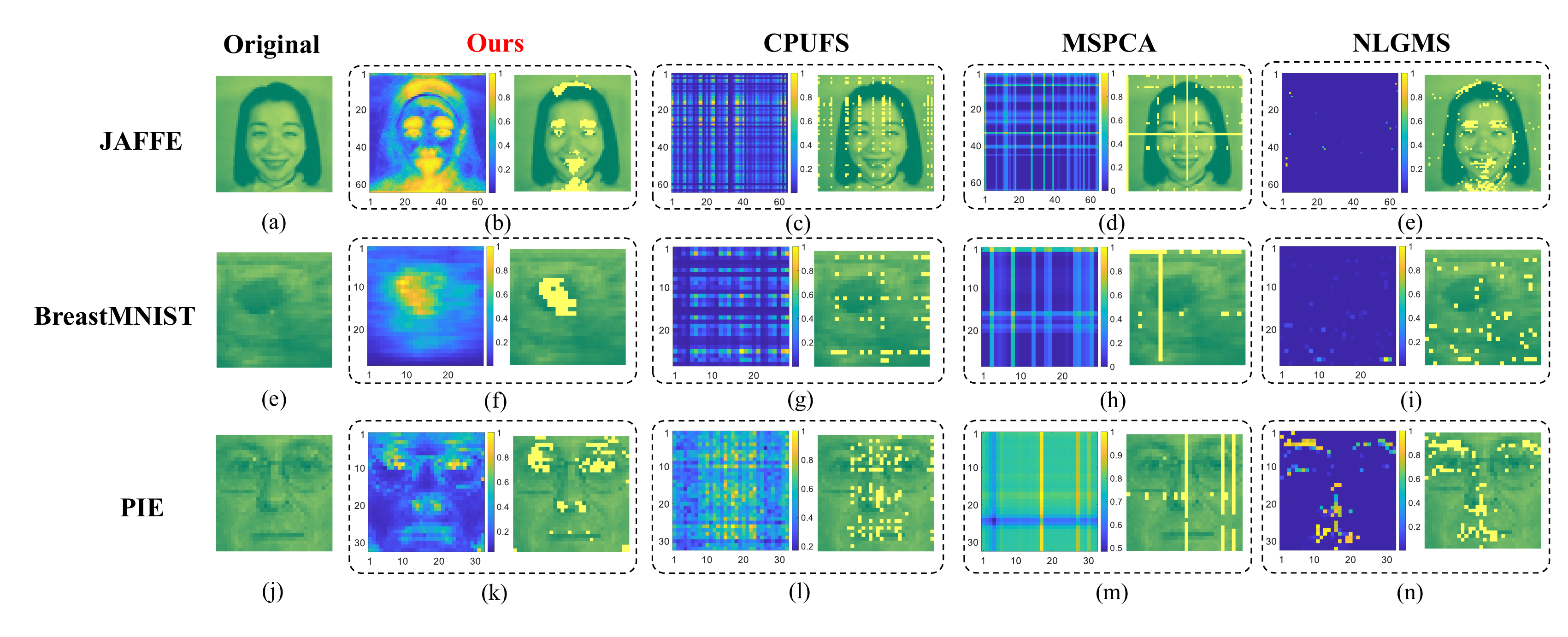}
    \caption{Visualization on PIE, JAFFE, and BreastMNIST. We show the score maps and highlight the positions of the top-ranked features in the samples. We set the number of selected features to $300$ on JAFFE, $50$ on BreastMNIST, and $100$ on PIE.}
    \label{Visualization}
\end{figure}

In this experiment, we conduct feature visualization to empirically compare the interpretability of selected features. Figure \ref{Visualization} displays the best NMI results from three tensor-based methods (STPCA-MP, CPUFS, MSPCA) and one non-tensor method (NLGMS) on three image datasets: JAFFE, BreastMNIST, and PIE, with selected features highlighted to reveal method attention patterns. We have the following observations:
\begin{enumerate}[\textbullet]
    \item \emph{STPCA-MP demonstrates superior discriminative feature selection}. By leveraging data orientation, STPCA-MP effectively captures essential spatial patterns: (i) On PIE and JAFFE, it consistently selects horizontal features corresponding to facial components (eyes, eyebrows, mouth, hair) crucial for facial and expression recognition (Figure \ref{Visualization}(b)(k)), while CPUFS, NLGMS and MSPCA misallocate attention to background pixels (Figure \ref{Visualization}(c)(l)(d)(m)); (ii) On BreastMNIST, it concentrates on relevant shadowed regions with both horizontal and vertical features (Figure~\ref{Visualization}(f)), unlike the scattered selections of CPUFS, MSPCA, and NLGMS (Figure~\ref{Visualization}(g)(h)(I)). Notably, MSPCA's sparse tensor decomposition provides moderate improvement over CPUFS by reducing background interference (Figure~\ref{Visualization}(d)(h)(m)).
    \item  \emph{STPCA-MP produces more interpretable score maps}. The slice-by-slice operation yields precise feature correlation modeling on: (i) Facial datasets, where STPCA-MP exhibits anatomically plausible attention maps resembling face contours (Figure~\ref{Visualization}(b)(k)); (ii) BreastMNIST, where STPCA-MP shows smooth score gradients radiating from pathological regions (Figure~\ref{Visualization}(f)). In contrast, CPUFS and MSPCA generate less coherent patterns due to inherent limitations of CPD/Tucker decompositions in modeling mode relations (Figure \ref{Visualization}(c)(d)(g)(h)(l)(m)), while NLGMS overweight background noise (Figure~\ref{Visualization}(e)(i)), likely due to graph construction sensitivity.
\end{enumerate}

\subsection{Training time comparison}
\begin{figure*}[h]
    \centering
    \subfloat[\centering{JAFFE (\#Feature=$4096$, \#Sample$=213$)}]{\includegraphics[width=1.4in]{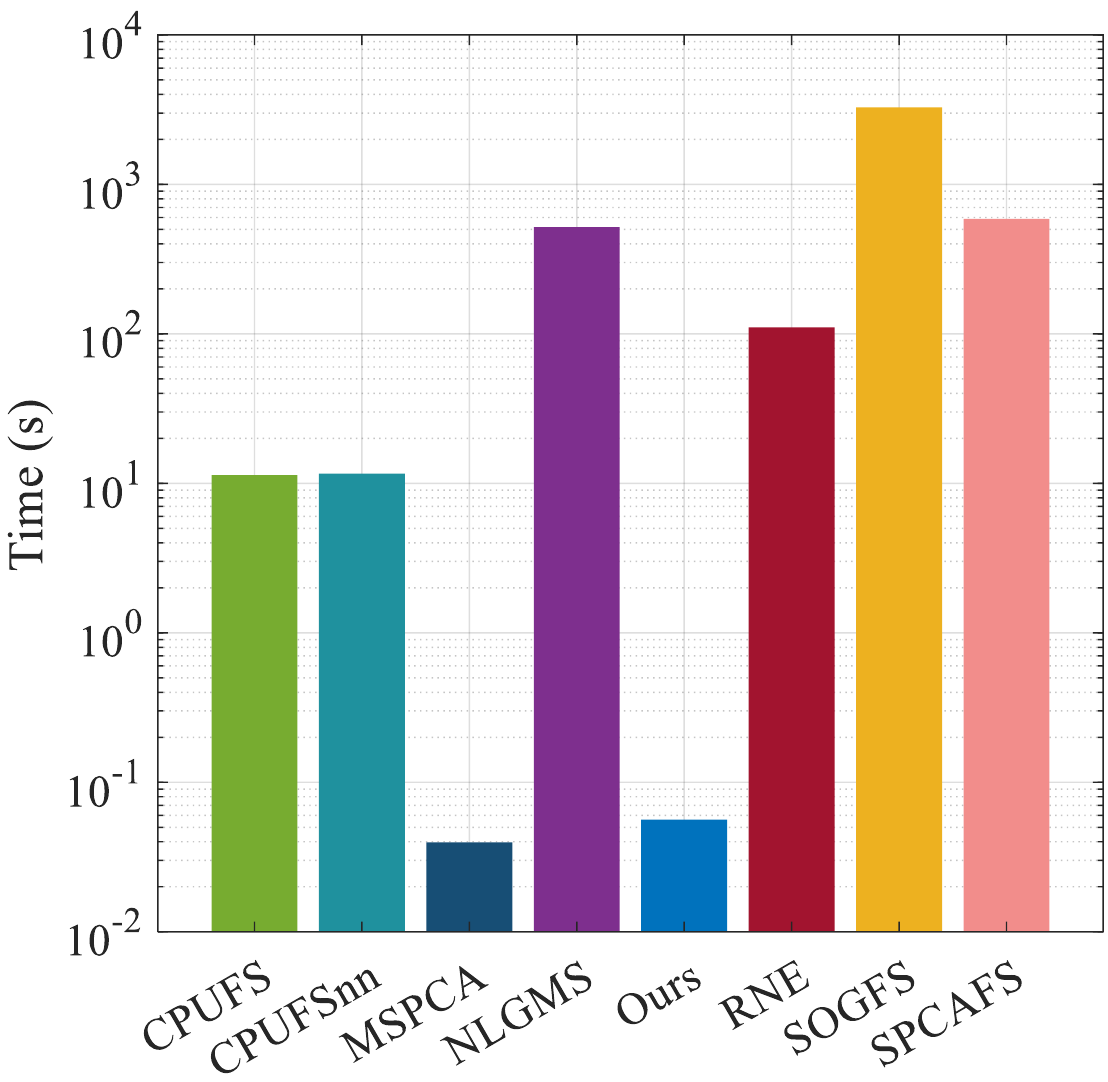}}
    \hspace{1em}
    \subfloat[\centering{UMCS (\#Feature=$8125$, \#Sample$=174$)}]{\includegraphics[width=1.4in]{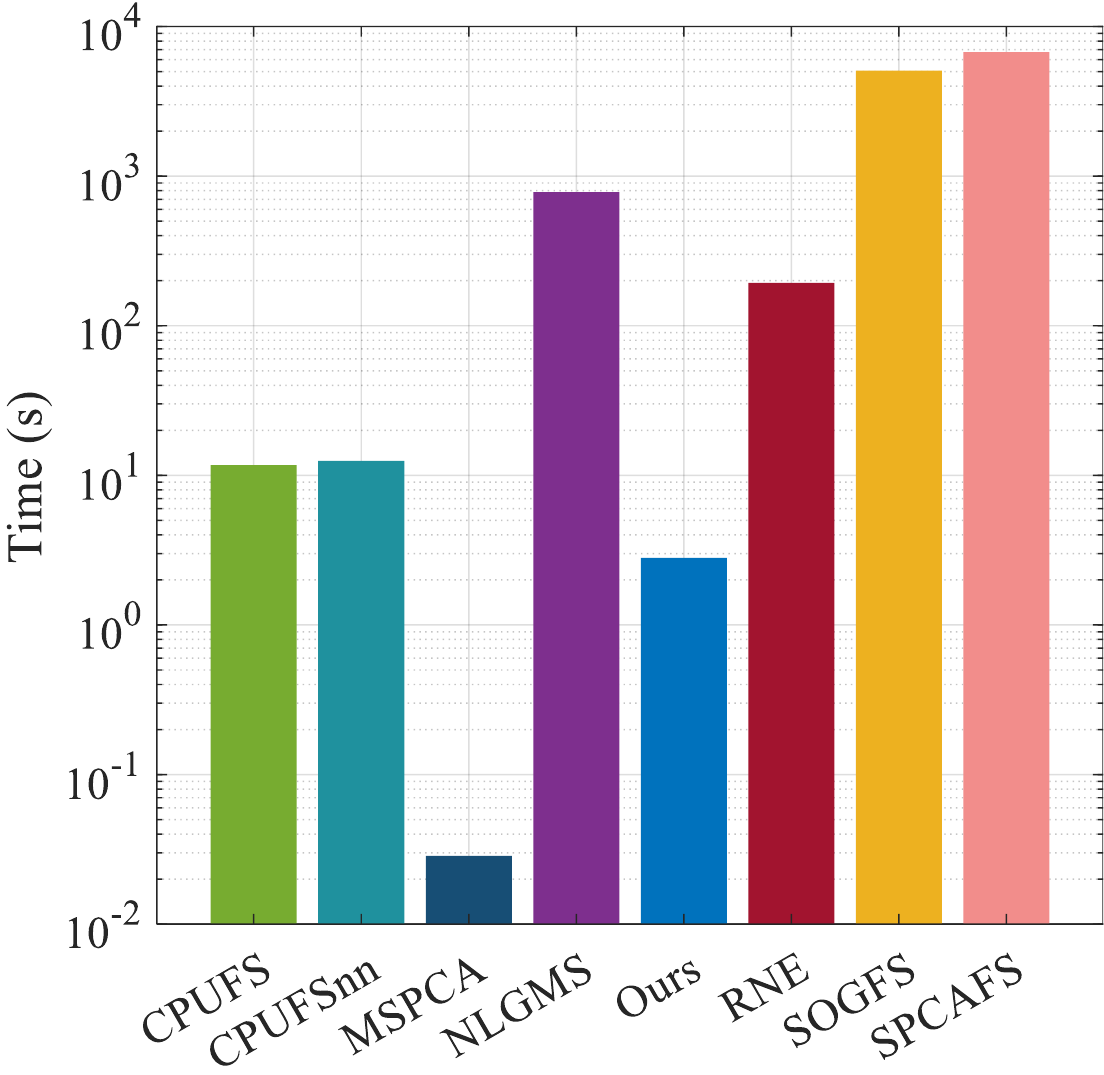}}
    \hspace{1em}
    \subfloat[\centering{ASCH (\#Feature=$10240$, \#Sample$=84$)}]{\includegraphics[width=1.4in]{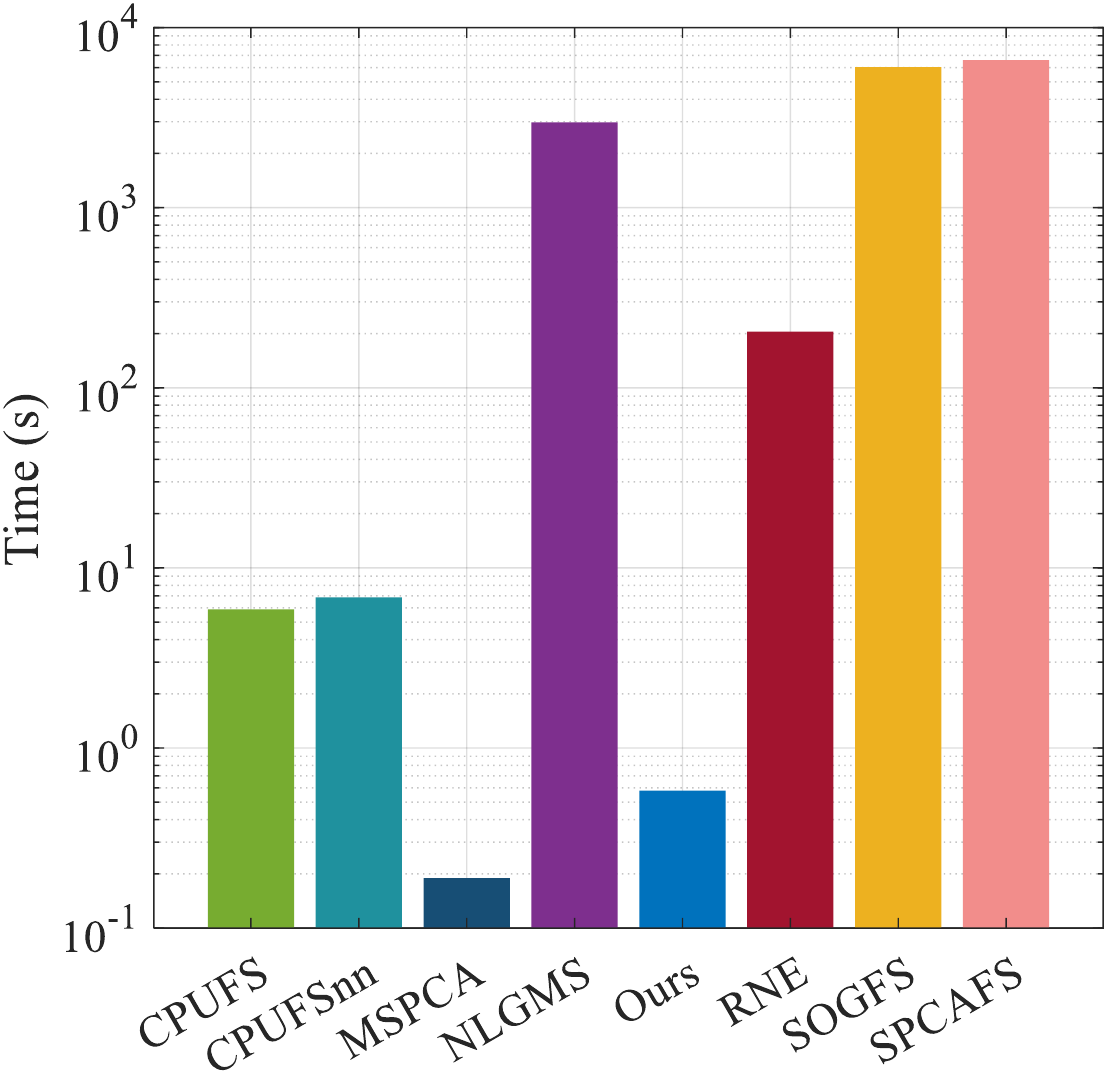}}
    \hspace{1em}
    \subfloat[\centering{USPS (\#Feature=$784$, \#Sample$=9298$)}]    {\includegraphics[width=1.4in]{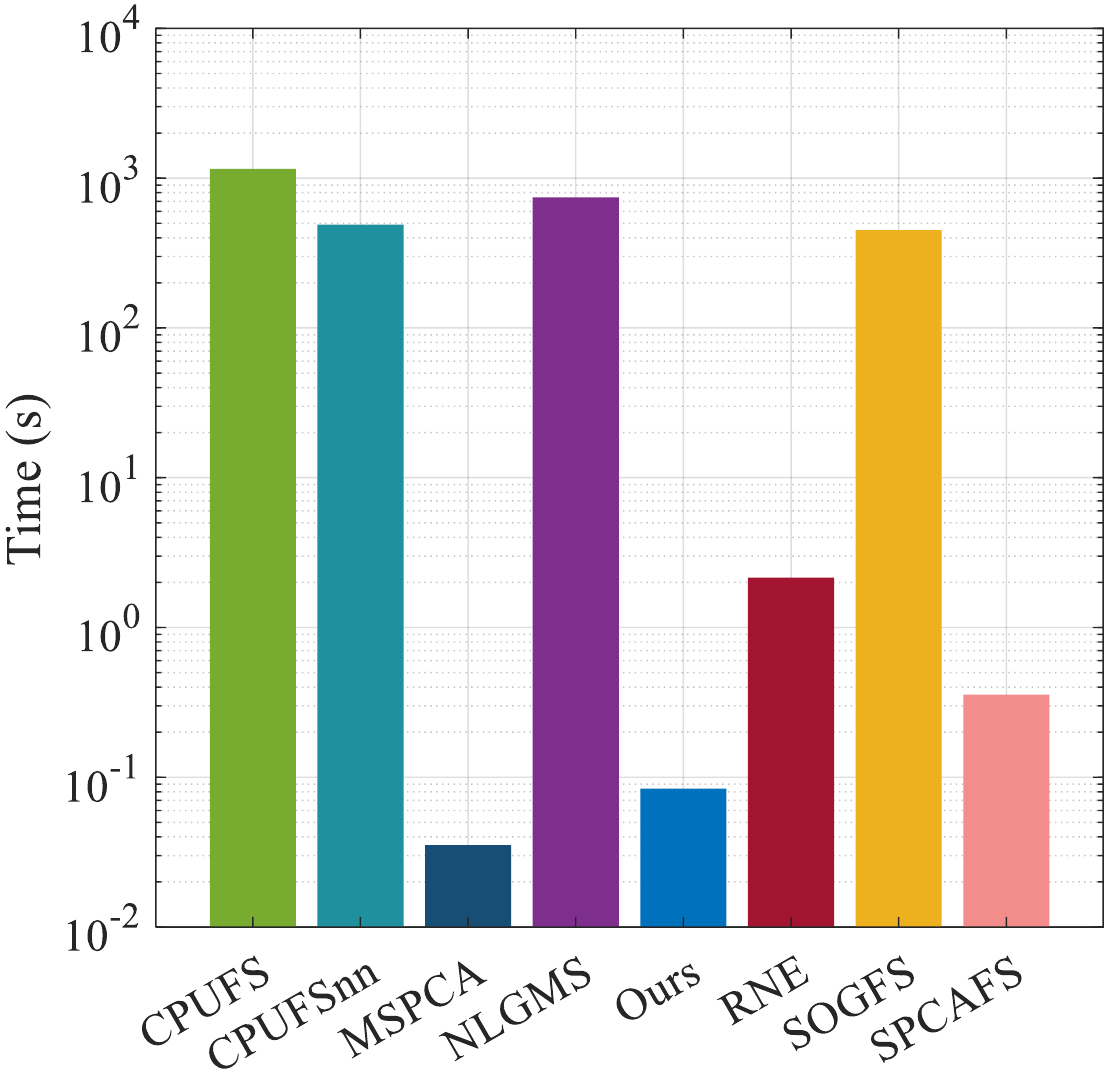}} 
    \caption{Average training time during grid search.}
    \label{training time}
\end{figure*}

In this experiment, we analyze the computational efficiency of comparative methods. The training time corresponding to optimal clustering performance is reported in Table~\ref{Metrics of the comparative methods on real-world data}, with average training times across parameter combinations shown in Figure~\ref{training time}. We have the following observations:
\begin{enumerate}[\textbullet]
    \item \emph{Tensor-based methods demonstrate superior efficiency on high-dimensional data}. By exploiting tensor operations, these methods reduce computational complexity from quadratic/cubic in full feature space to quadratic/cubic within single-mode dimensions. This explains why CPUFS, CPUFSnn, MSPCA, and our method consistently achieve lower training times on JAFFE, UMCS, and ASCH datasets compared to non-tensor alternatives (Figure \ref{training time}(a)(b)(c)). 
    \item \emph{STPCA-MP maintains efficiency across both high-dimensional and large-scale datasets}.While CPUFS and CPUFSnn require hundreds of iterations (due to non-convexity) with quadratic sample complexity $\mathcal{O}(n^2)$, STPCA-MP achieves linear complexity $\mathcal{O}(n)$. This efficiency advantage becomes particularly pronounced on large-scale datasets like USPS (Figure \ref{training time}(d)). Although MSPCA's Tucker decomposition with smaller weight parameters yields marginally faster average training times, it fails to outperform STPCA-MP in either clustering metrics or POC (Table \ref{Metrics of the comparative methods on real-world data}). Notably, the runtime difference between MSPCA and our method at optimal performance is negligible (Table~\ref{Metrics of the comparative methods on real-world data}), confirming STPCA-MP's balanced effectiveness-efficiency trade-off.
\end{enumerate}

\subsection{Convergence analysis}
\label{Convergence analysis}
\begin{figure}[h]
    \centering
    \subfloat[PIE]{\includegraphics[width=1.5in]{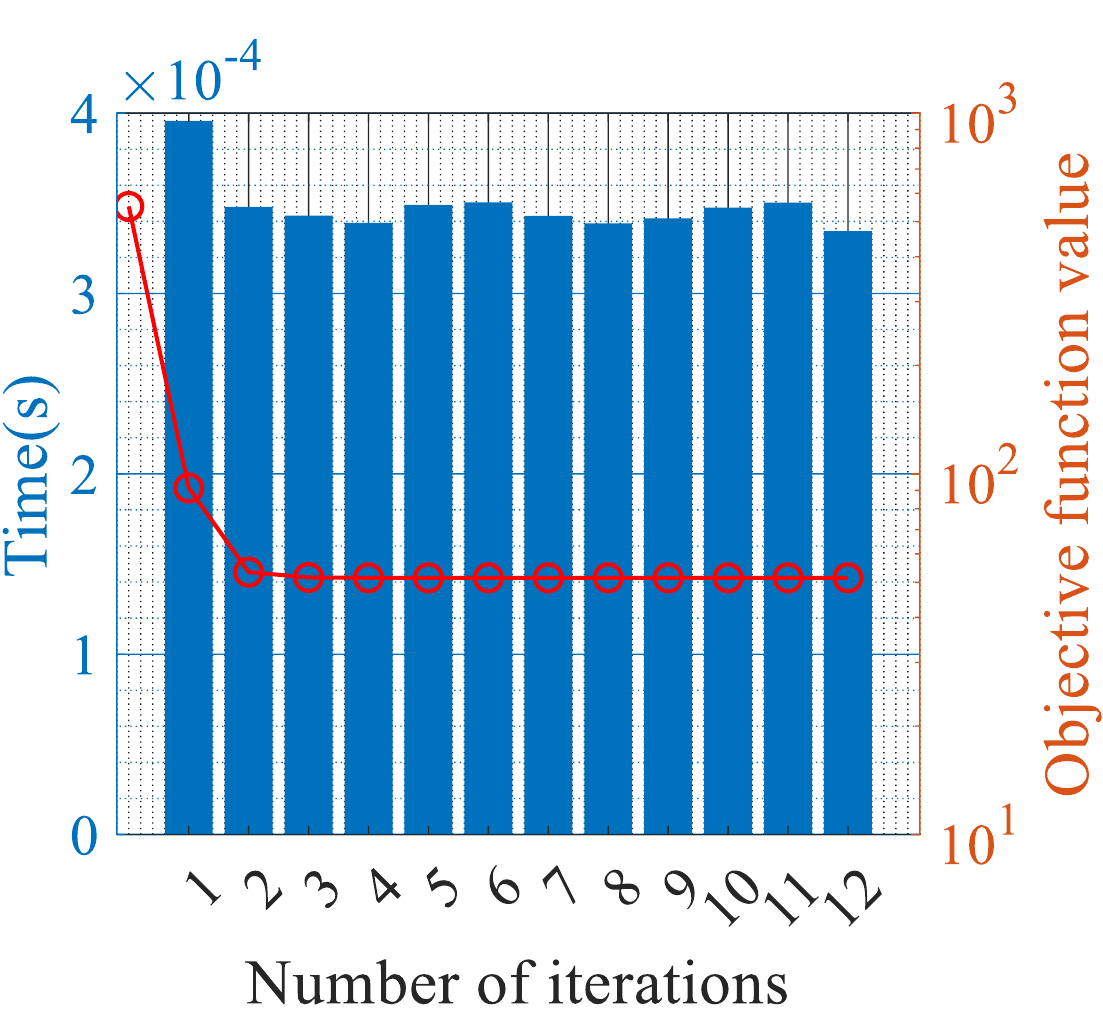}}\hspace{0.5em}
    \subfloat[JAFFE]{\includegraphics[width=1.5in]{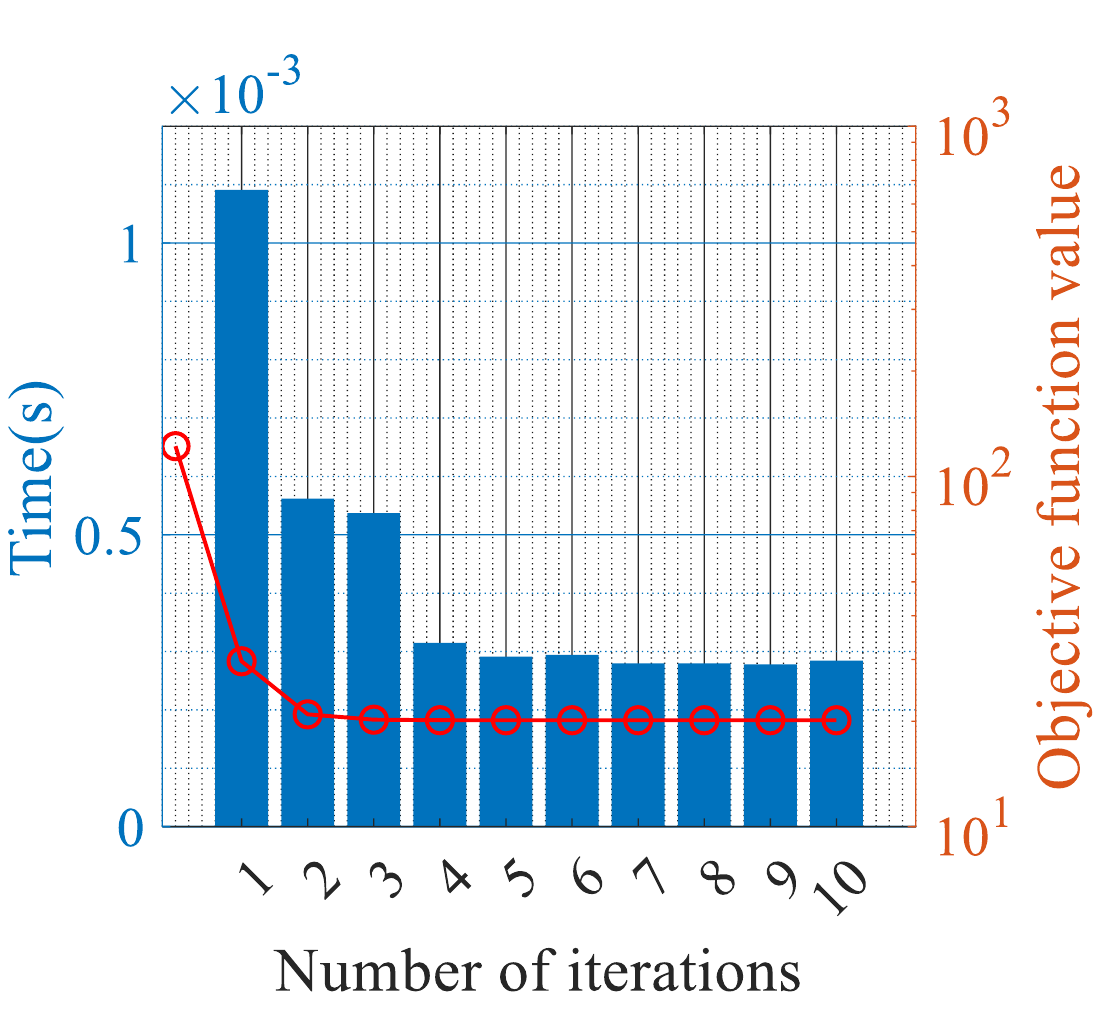}}\hspace{0.5em}
    \subfloat[BreastMNIST]{\includegraphics[width=1.5in]{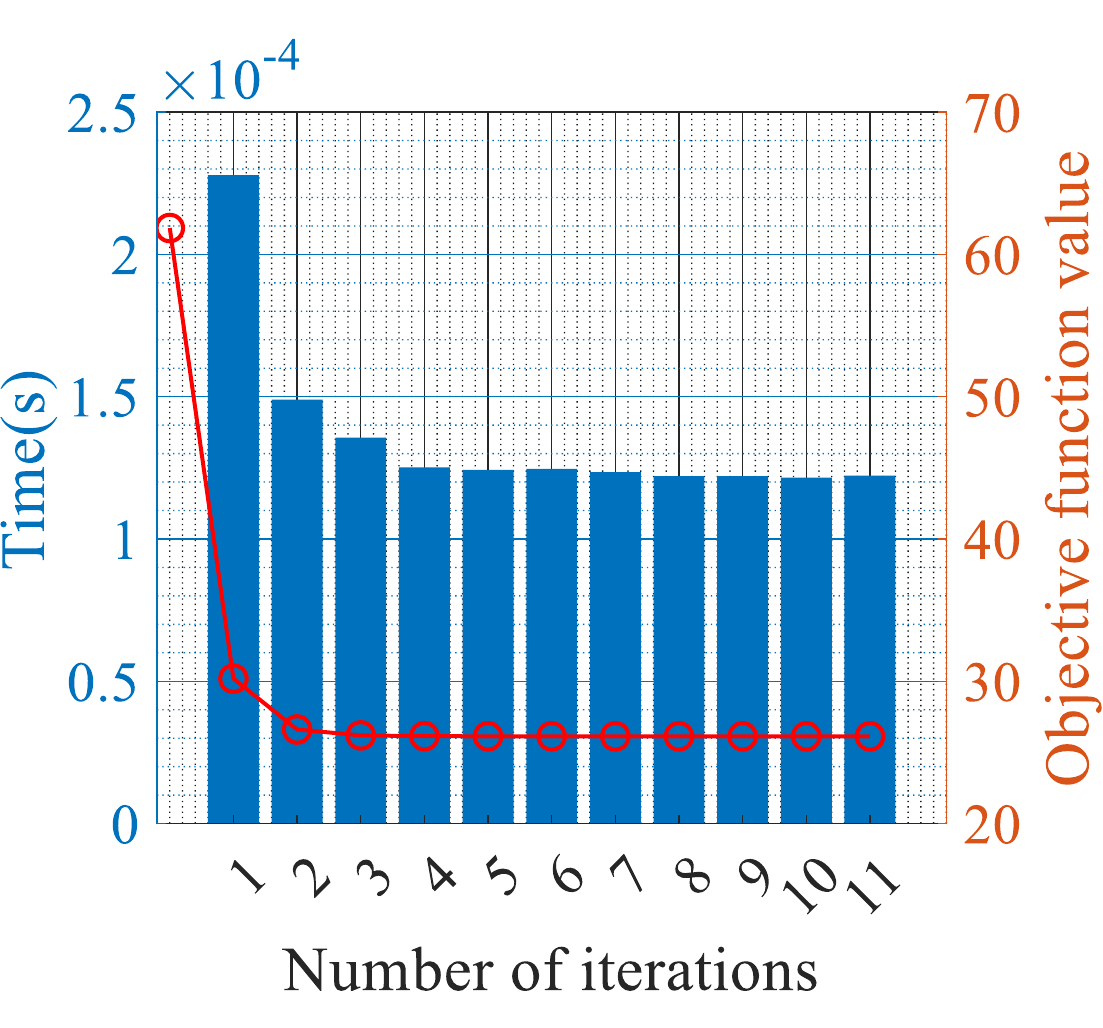}}\hspace{0.5em}
    \subfloat[UCIDSA]{\includegraphics[width=1.5in]{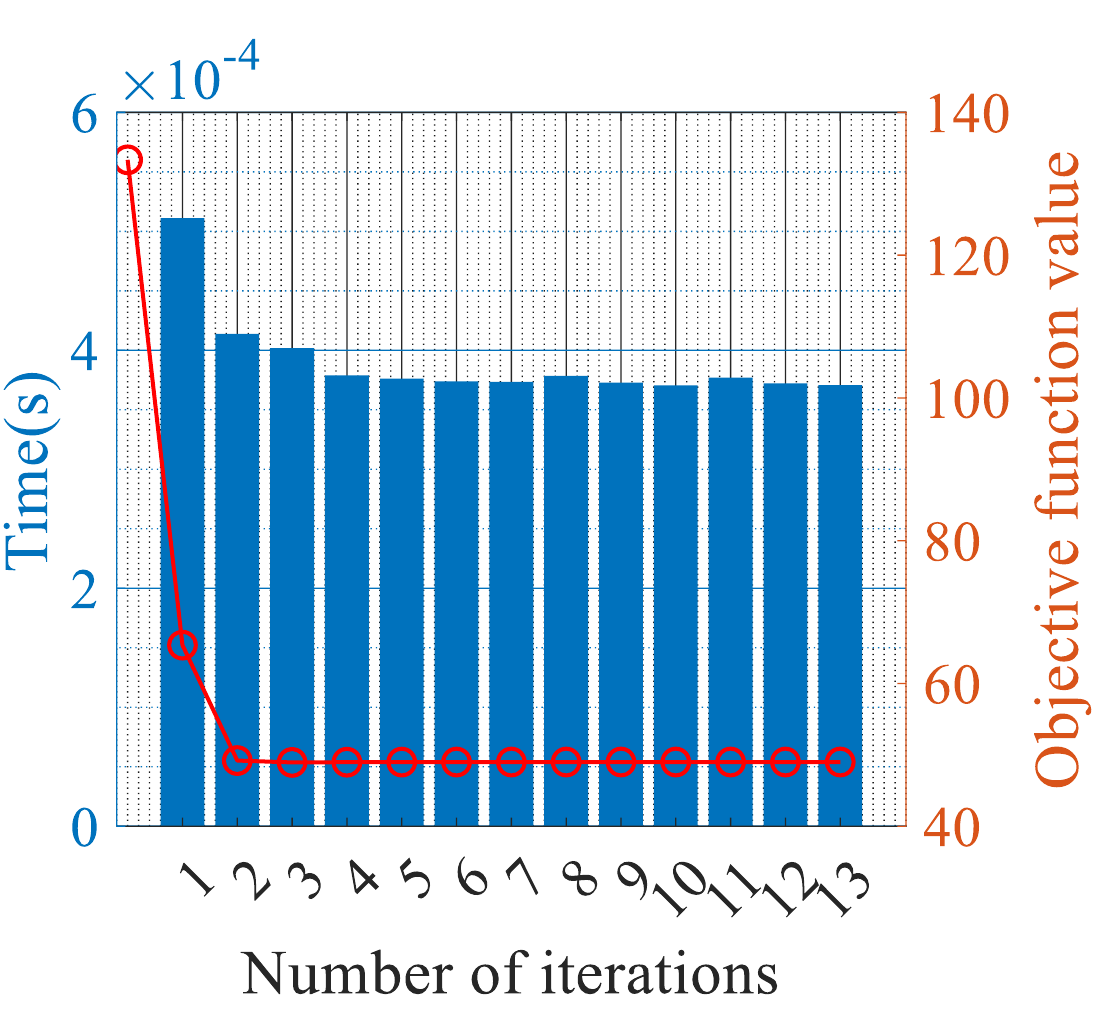}}
    \caption{Convergence curves and runtime per iteration of STPCA-MP on four real-world datasets. All regularization parameters are set to 10.}
    \label{Converge Curves}
\end{figure}

In this experiment, we investigate the convergence properties of the proposed method through a representative subproblem analysis due to space constraints. Figure~\ref{Converge Curves} demonstrates the convergence behavior by plotting both the objective function values and runtime per iteration across four real-world datasets. The two-step algorithm achieves rapid convergence to the solution space within dozens of iterations, facilitated by the HPSD projector that ensures stable optimization. The slice-by-slice operation significantly reduces computational complexity, enabling efficient execution of each iteration while maintaining solution quality.

\subsection{Parameter sensitivity analysis}
\begin{figure}
    \centering
    \subfloat[JAFFE (ACC)]{\includegraphics[width=1.5in]{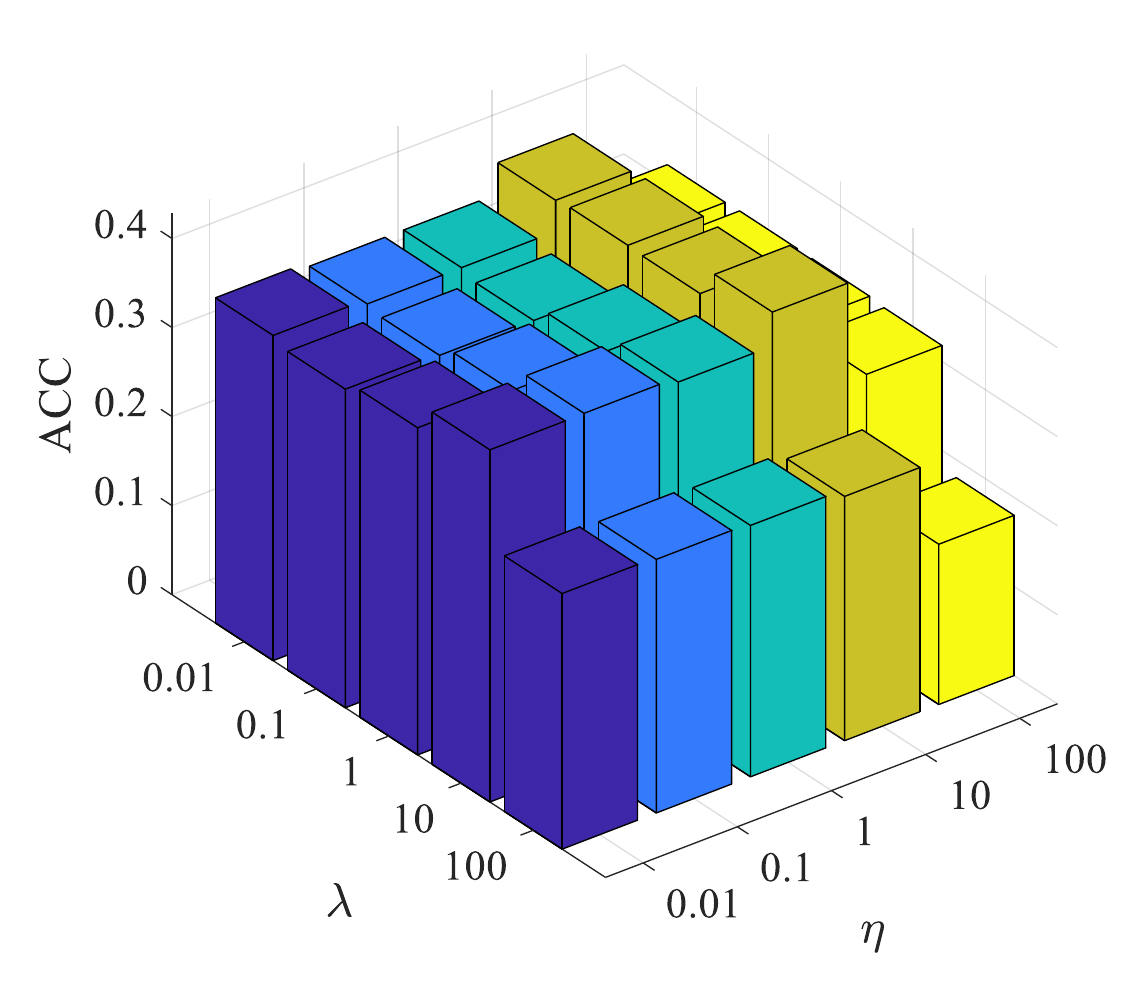}}
     \subfloat[JAFFE (NMI)]{\includegraphics[width=1.5in]{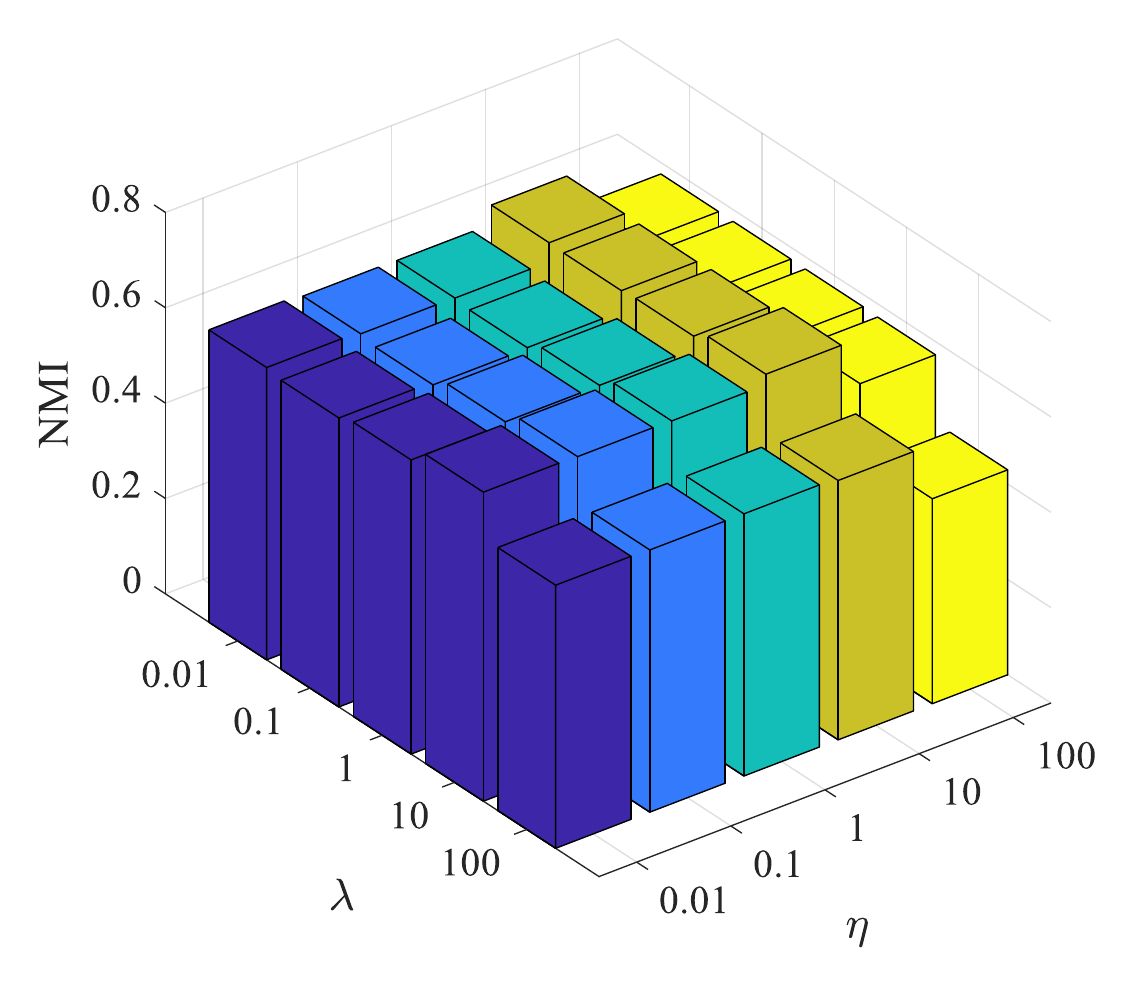}}
    \subfloat[USPS (ACC)]{\includegraphics[width=1.5in]{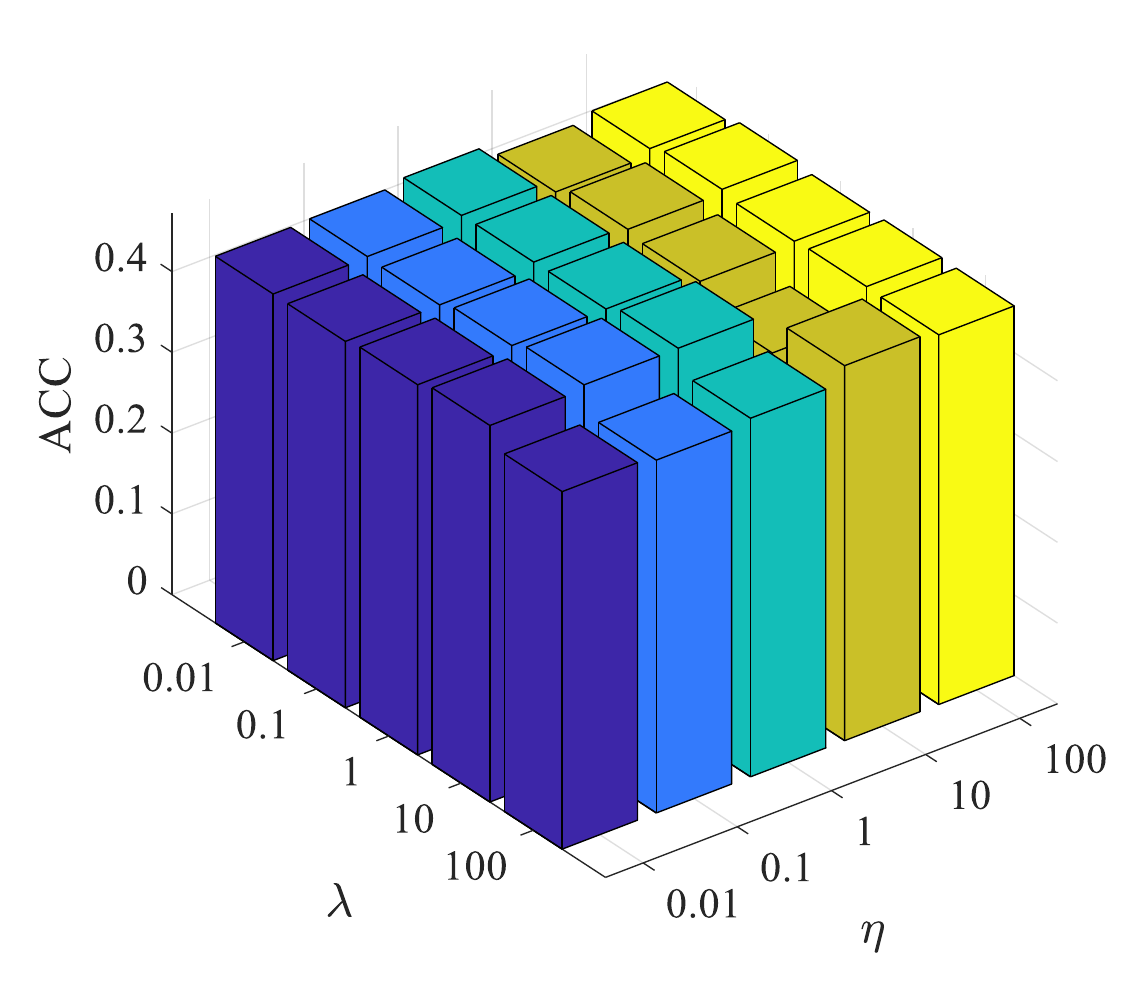}}
     \subfloat[USPS (NMI)]{\includegraphics[width=1.5in]{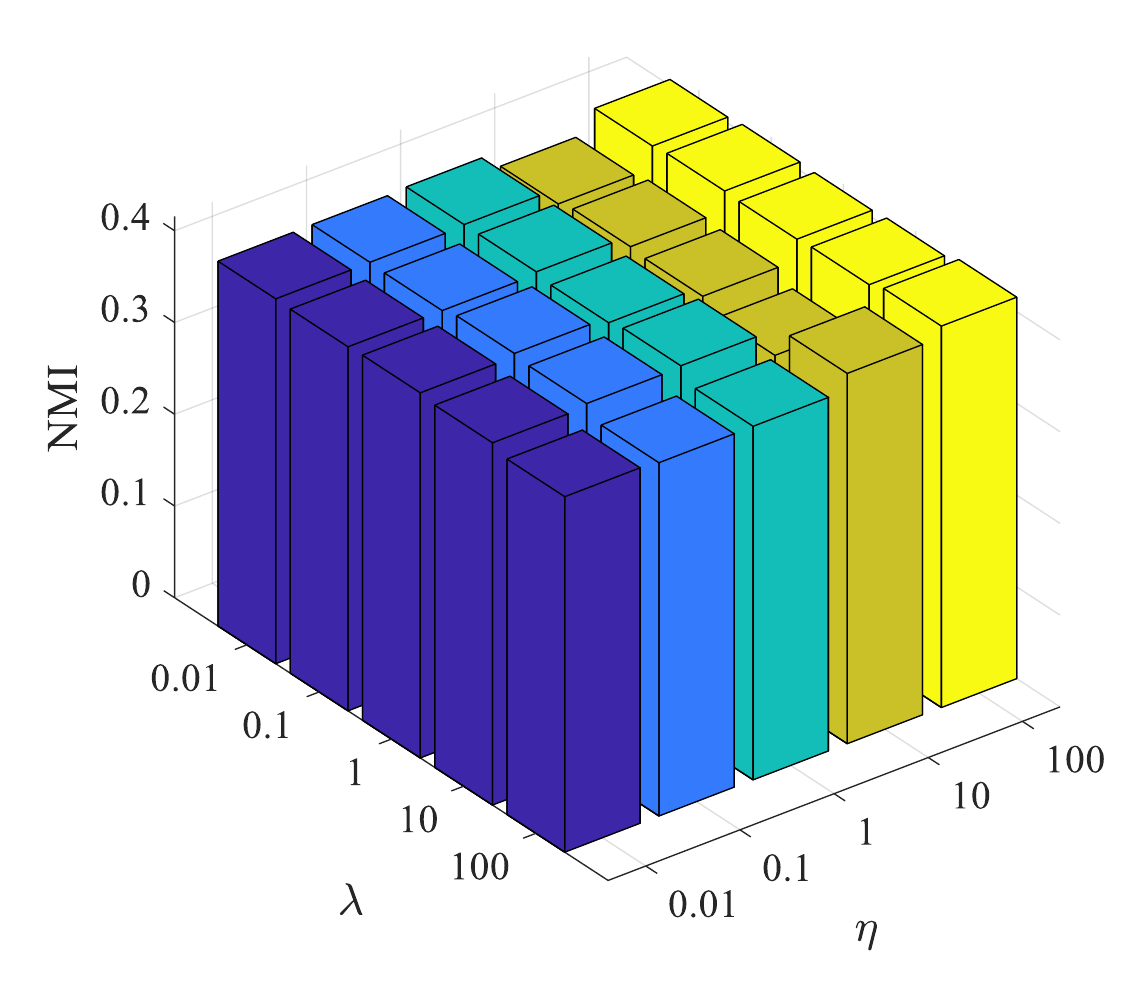}}     
    
    \caption{Parameter sensitivity analysis of STPCA-MP on JAFFE and USPS.}
    \label{Parameter sensitivity}
\end{figure}

In this section, we evaluate the impact of regularization parameters $\lambda$ and $\eta$ on the feature selection performance of STPCA-MP. By varying both parameters within $\{10^{-2},10^{-1},1,10^1,10^2\}$ while fixing the selected features at 100, we record the ACC and NMI values across all parameter combinations. Figure~\ref{Parameter sensitivity} presents the results on JAFFE and USPS datasets, revealing two key findings: 


\begin{enumerate}[\textbullet]
    \item For JAFFE, STPCA-MP maintains stable performance when $\lambda$ and $\eta$ remain within a moderate range, with noticeable degradation occurring only at excessively large parameter values - a trend consistently observed in both ACC and NMI metrics.
    \item For USPS, the method demonstrates robust performance across the tested parameter range, showing minimal sensitivity to $\lambda$ and $\eta$ variations in terms of both ACC and NMI measurements.
\end{enumerate}

\subsection{Impact of Data Orientation and Transform Domain}

\begin{figure}[t]
    \centering
    
    \subfloat[{JAFFE (different orientations)}]{\includegraphics[width=1.5in]{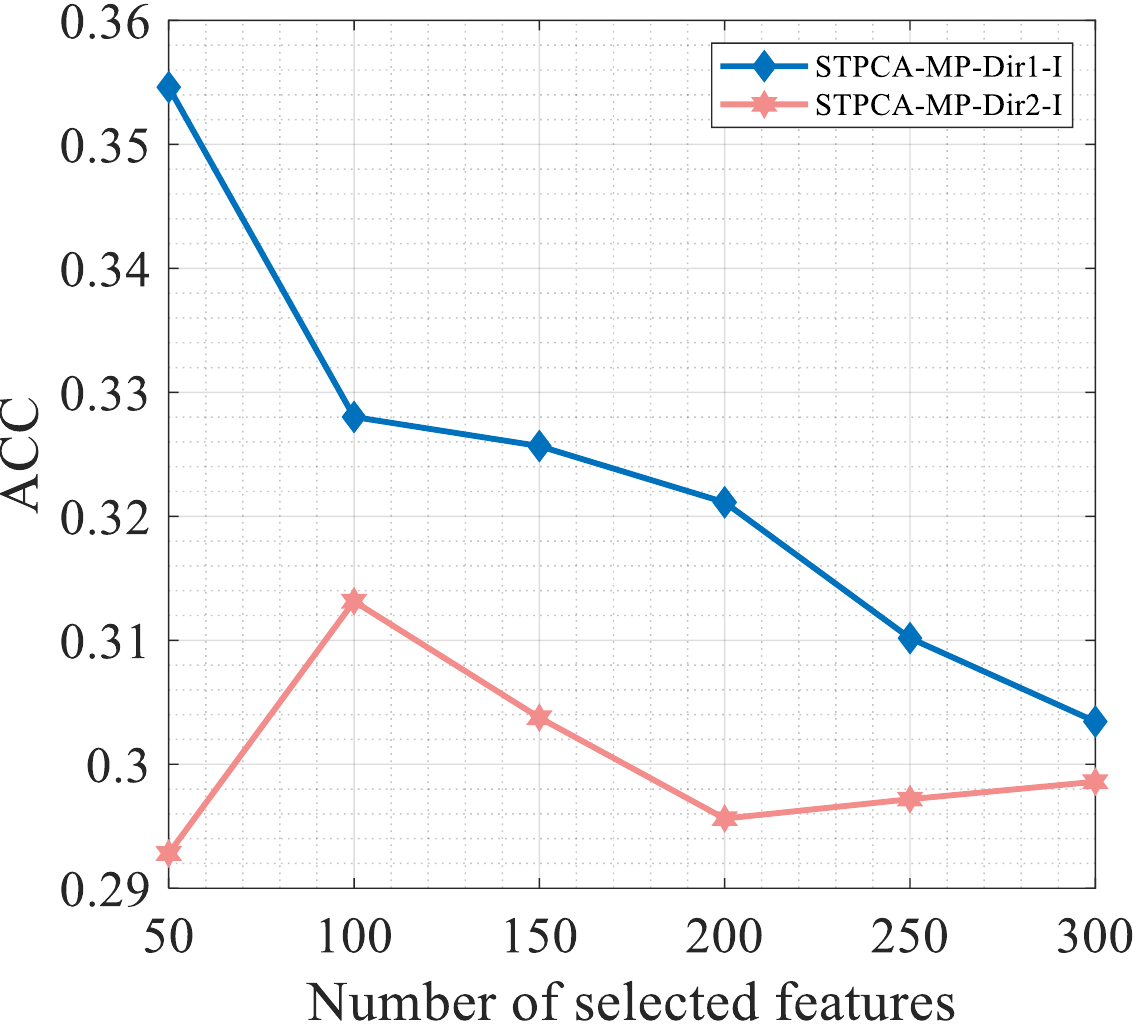}}
    \hspace{0.5em}
     \subfloat[{USPS (different orientations)}]{\includegraphics[width=1.5in]{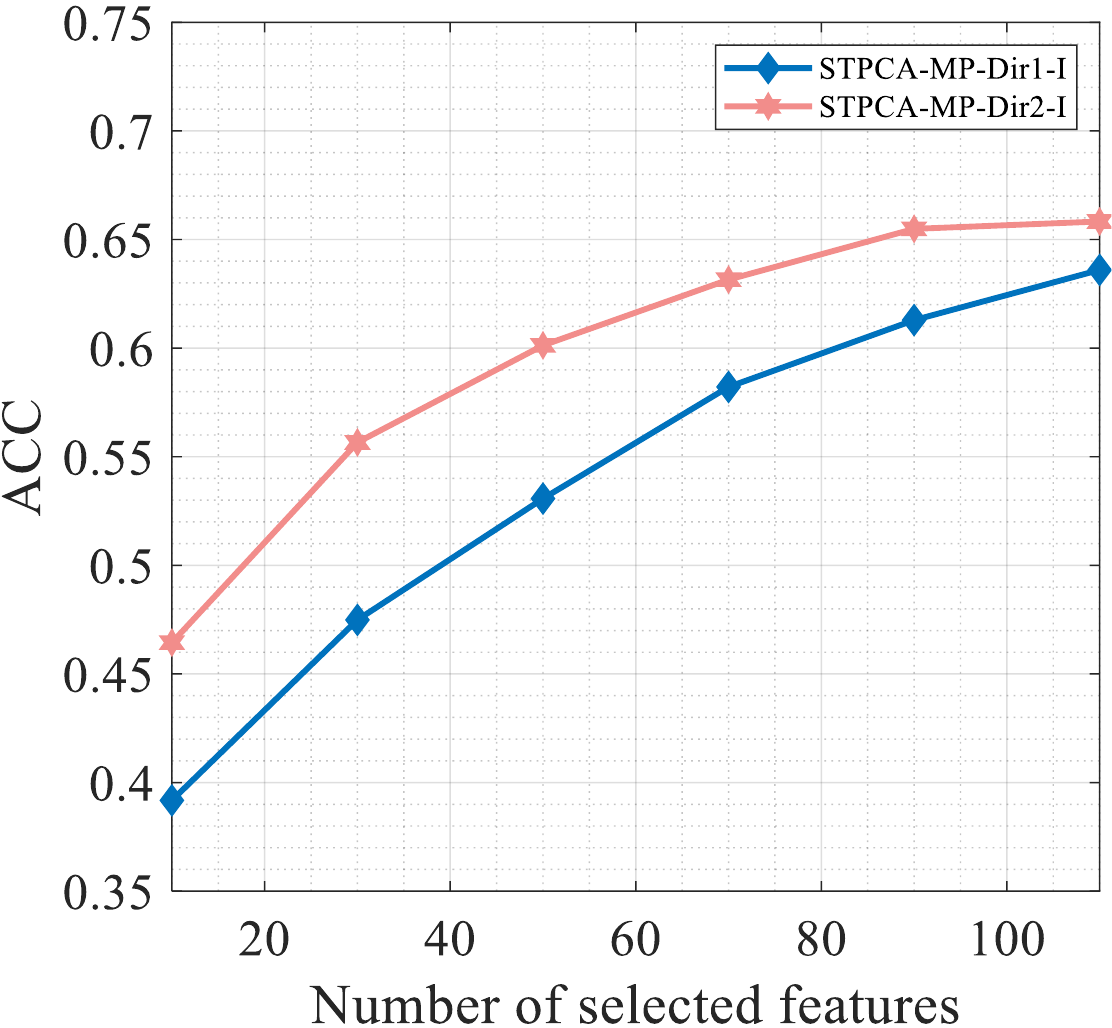}}
     \hspace{0.5em}    
     \subfloat[{JAFFE (different $\mathbf{M}$)}]{\includegraphics[width=1.5in]{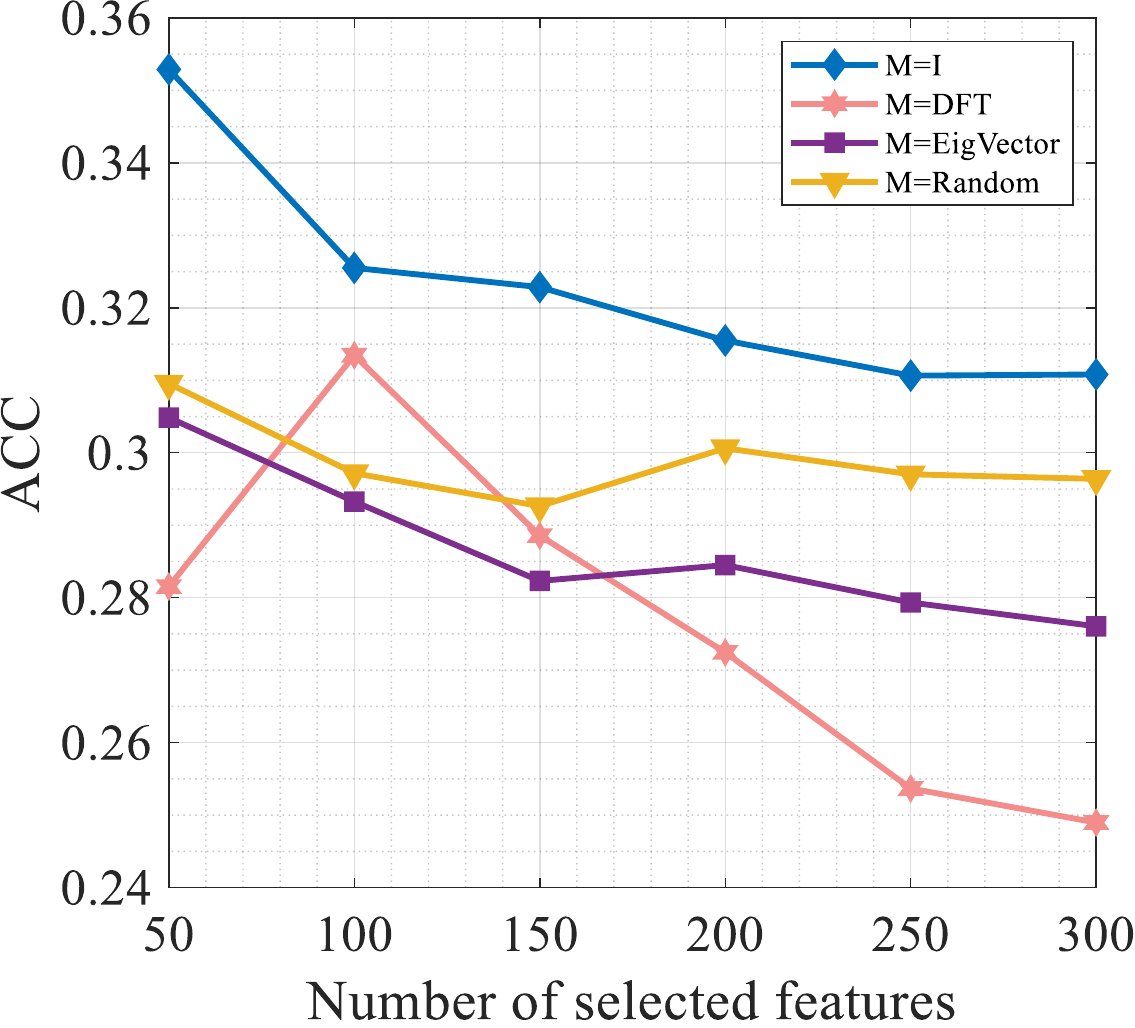}}
    \hspace{0.5em}     
    \subfloat[{USPS (different $\mathbf{M}$)}]{\includegraphics[width=1.5in]{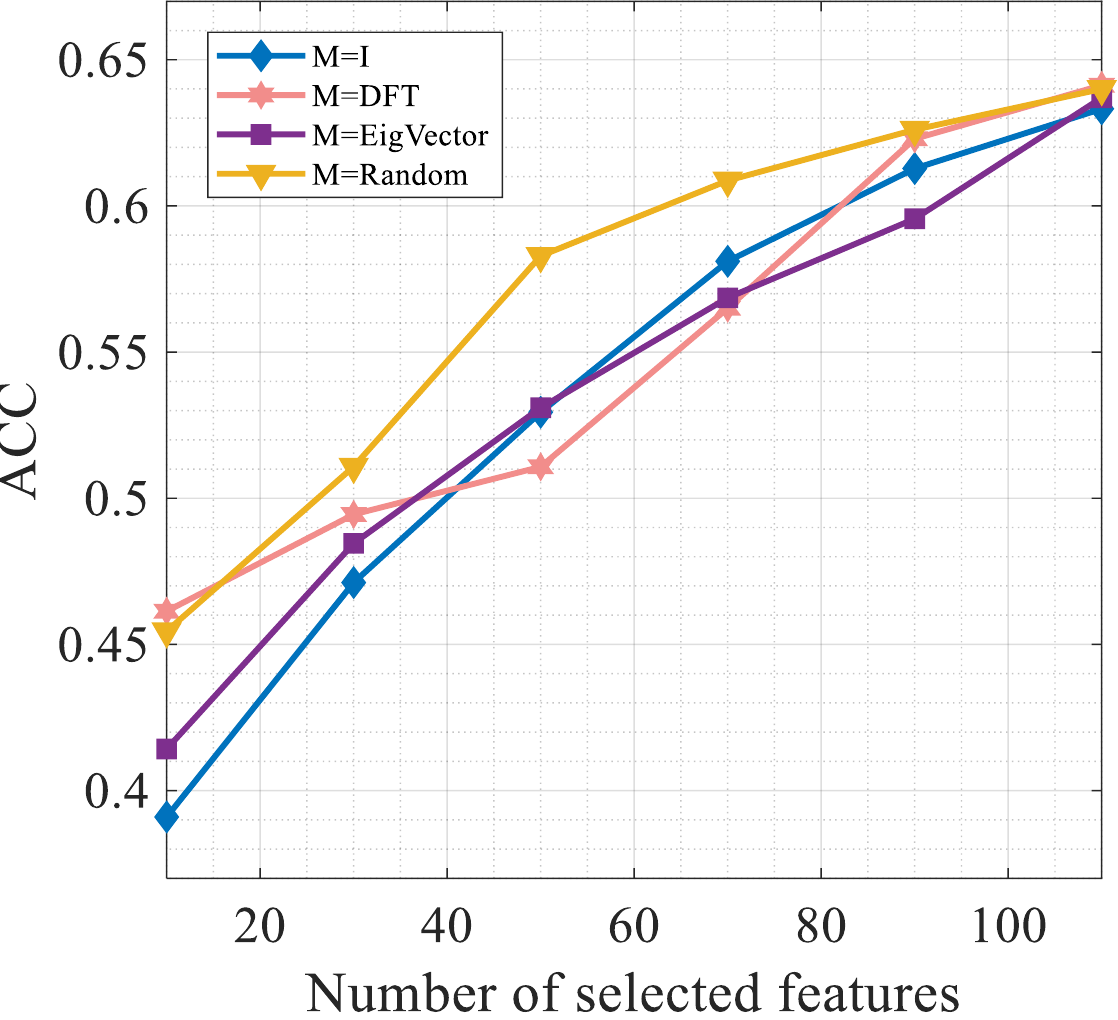}}
    \caption{The clustering ACC of STPCA-MP with different data tensor orientations and $\mathbf{M}$. Different data tensor orientations. 'Dir1' means $D^o=\{1,3,2\}$. 'Dir2' means $D^o=\{2,3,1\}$. 'I' means $\mathbf{M}=\mathbf{I}$. 'DFT' means that $\mathbf{M}$ is the discrete Fourier matrix (we conduct fast Fourier transform in practice). 'EigVector' means $\mathbf{M}$ is the eigenvectors decomposed from the self-correlation matrix of the other feature mode. 'Random' means $\mathbf{M}$ is a random orthogonal matrix. (c) and (d) are obtained by running STPCA-MP-Dir1.} 
    \label{Different M and orientation}
\end{figure}

\begin{figure}
    \centering
    \includegraphics[width=6in]{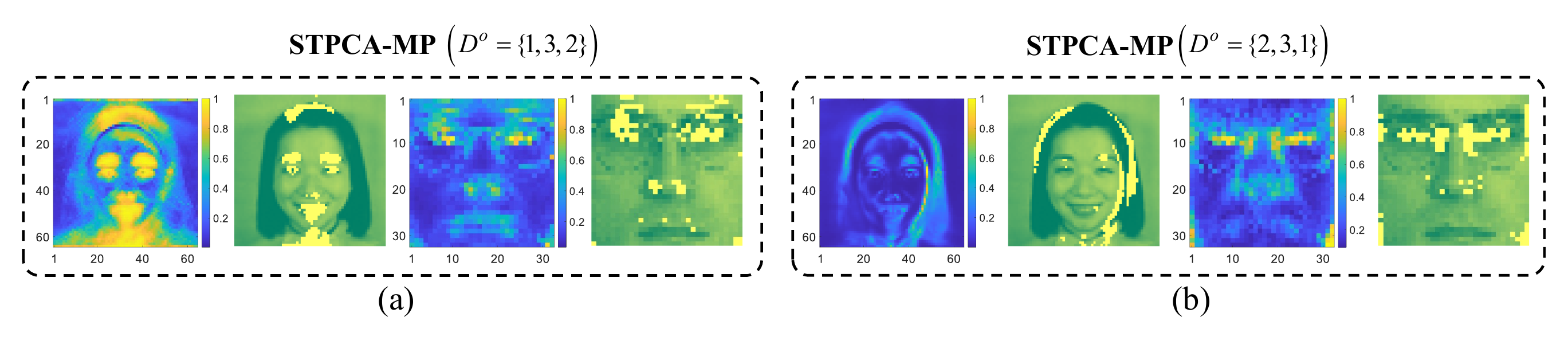}
    \caption{Visualized feature selection results of our proposed method with different data orientations. The experimental condition is the same as in Figure \ref{Visualization}.}
    \label{Visualization_different_direction}
\end{figure}

In this experiment, investigate how data tensor orientation and transform matrix $\mathbf{M}$ affect STPCA-MP performance. Experiments evaluate different tensor rotations and transform matrices across multiple datasets, with detailed ACC curves for JAFFE and USPS shown in Figure~\ref{Different M and orientation}. Table~\ref{ACC_POC_different_direction_M} reports the optimal ACC and mean POC values, while Figure~\ref{Visualization_different_direction} visualizes results using $\mathbf{M} = \mathbf{I}$ on PIE and JAFFE. We have the following observations:
\begin{enumerate}[\textbullet]
    \item \emph{Data orientation critically influences clustering performance}. Significant ACC variations in Figure~\ref{Different M and orientation}(a) and (b) demonstrate that, similar to viewing objects from different perspectives, tensor decomposition along distinct directions captures varying information. Figure~\ref{Visualization_different_direction} confirms STPCA-MP's ability to identify orientation-dependent spatial feature distributions. Optimal orientation selection enhances feature correlation mining, particularly for multi-channel time-series data like UCIDSA and UMCS, where 1-mode operations outperform due to inherent feature relationships along this dimension.
    \item \emph{Transform domain selection impacts results}. A well-selected transform matrix extracts information from one feature mode while slice-by-slice operations process the other, enabling prior knowledge incorporation for optimization. Table~\ref{ACC_POC_different_direction_M} shows Dir1-EigVector occasionally surpassing Dir-I, especially for time-series data.  However, no transform configuration universally dominates - both Figure~\ref{Different M and orientation}(a)(b)(c)(d) and Table~\ref{ACC_POC_different_direction_M} demonstrate performance varies substantially across datasets. Therefore, optimal orientation and transform choices require dataset-specific tuning.

\end{enumerate}

\begin{table}[t]
    \caption{The best ACC and mean POC of our proposed method with different data orientations and transform domains. The best result on each dataset is boldfaced, while the second best one is underlined and italic.}
    \centering
    \begin{tabular}{cccccc|ccc}
    \toprule 
    \textbf{Setting} & \textbf{Metric} & \textbf{JAFFE} & \textbf{BreastMNIST} & \textbf{USPS} & \textbf{PIE} & \textbf{Metric} & \textbf{UCIDSA} & \textbf{UMCS} \\
    \midrule
    Dir1-I  & ACC & \textbf{35.29} & \emph{\uline{64.34}} & 63.32 & \textbf{42.84} & POC & \emph{\uline{92.27}} & \emph{\uline{41.07}}  \\
    Dir1-DFT  & ACC & \emph{\uline{31.35}} & 55.13 & \emph{\uline{64.13}} & 28.20 & POC & 9.33 & 38.67 \\
    Dir1-EigVector  & ACC & 30.49 & 57.56 & 63.7 & 30.98 & POC & \textbf{92.8} & \textbf{44.80} \\
    \midrule
    Dir2-I  & ACC & 30.74 & 62.44 & \textbf{66.77} & 36.87 & POC & 65.07 & 28.8  \\
    Dir2-DFT  & ACC & 28.48 & 62.29 & 57.34 & \emph{\uline{38.39}} & POC & 24 & 16.53 \\
    Dir2-EigVector  & ACC & 22.68 & \textbf{65.32} & 61.55 & 33.02 & POC & 45.87 & 18.13 \\

    \bottomrule
    \end{tabular}
    \label{ACC_POC_different_direction_M}
\end{table}

\section{Conclusion}
In this paper, we propose a novel Sparse Tensor PCA method named STPCA-MP by combining the encoder-decoder structure of SPCA and the orientation-dependent framework of T-SVDM. We deduce a convex version of STPCA-MP that utilizes the sparse reconstruction tensor to perform UFS. An efficient slice-by-slice algorithm to conduct STPCA-MP is proposed to guarantee fast convergence. The experimental results on real-world data show that our method outperforms state-of-the-art methods on average in terms of clustering metrics (ACC, NMI, and SS) in different data organization scenarios at a smaller training time cost. In particular, STPCA-MP has the advantage of selecting discriminative features with the greatest BCV on slice-wise data. In addition, we also test the parameter sensitivity, examine the efficient convergence, and discuss the impacts of choosing different data orientations and transform domains. The experimental results show that when the transform axes align with the feature distribution pattern, our method can capture abundant discriminative information and achieve remarkable performance.  

There are two valuable research points concerning STPCA-MP for the future. First, the best choice of transform domain is possibly related to the data background knowledge. More experiments should be carried out for further applications. Second, the extension to higher-order scenarios of STPCA-MP still lacks an effective and efficient optimization algorithm. We will be working on addressing the above two problems in the future.

\appendix
\section{Proof of Theorem \ref{theorem 1}}
\newproof{pf3}{Proof}

\begin{pf3}
        To simplify the deduction, we first rewrite the $i$-th problem with compact symbols:
    \begin{equation}
    \label{simplified problem}
    \begin{aligned}
    \min\limits_{\mathbf{U}, \mathbf{Q}}\quad & {\Vert \mathbf{X}-\mathbf{U}\mathbf{Q}^H\mathbf{X} \Vert}^2_F + \lambda{\Vert \mathbf{Q}\Vert}_{2,1},\\
     s.t. \quad & \mathbf{U}^H\mathbf{U}=\mathbf{I}_k.
    \end{aligned}
    \end{equation}
     Thus, we can rewrite the objective function as: 
    \begin{equation}
        \begin{aligned} 
        &{\rm Tr}\left(\mathbf{X}\mathbf{X}^H\right)-\sum\limits^{k}_{j=1}\bigg[2Tr\left(\mathbf{u}_j^H\mathbf{X}\mathbf{X}^H\mathbf{q}_j\right)-\frac{\lambda}{\Vert\mathbf{q}_j\Vert_2}\mathbf{q}_j^H\mathbf{q}_j-{\rm Tr}\left(\mathbf{q}_j^H\mathbf{X}\mathbf{X}^H\mathbf{q}_j\right) \bigg]\\
        &={\rm Tr}\left(\mathbf{X}\mathbf{X}^H\right)-\sum\limits^{k}_{j=1}\bigg(2\mathbf{u}_j^H\mathbf{X}\mathbf{X}^H\mathbf{q}_j-\frac{\lambda}{\Vert\mathbf{q}_j\Vert_2}\mathbf{q}_j^H\mathbf{q}_j-\mathbf{q}_j^H\mathbf{X}\mathbf{X}^H\mathbf{q}_j\bigg)
        \end{aligned}
    \end{equation}
    if we view the above problem as a sum of $k$ subproblems with respect to $\mathbf{u}_j$ and $\mathbf{q}_j$. Then in each iteration, given a fixed $\mathbf{u}_j$, we can have each subproblem minimized at
    \begin{equation}
    \label{the optimal q}
        \mathbf{q}^*_{opt} = \left(\mathbf{X}^*_{L_e}\mathbf{X}^T_{L_e}+\frac{\lambda}{\Vert \mathbf{q} _j\Vert_2}\mathbf{I}_{q}\right)^{-1}\mathbf{X}^*\mathbf{X}^T\mathbf{u}_j^*.
    \end{equation}
    
    By substituting (\ref{the optimal q}) back to (\ref{simplified problem}) we can obtain
    \begin{equation}
    \begin{aligned}
           &\mathbf{u}_{opt} = \\&{\underset{\mathbf{u}^H\mathbf{u}=1}{\arg\min}}-\mathbf{u}\mathbf{X}\mathbf{X}^H\left(\mathbf{X}\mathbf{X}^H+
           \frac{\lambda}{\Vert \mathbf{q} \Vert_2}\mathbf{I}_{q}\right)^{-1}\mathbf{X}\mathbf{X}^H\mathbf{u}^H.
    \end{aligned}
     \end{equation}
       Given the singular decomposition $\mathbf{X}=\mathbf{V}\mathbf{D}\mathbf{M}^H$, we have $\mathbf{u}^{opt}_j=s_j\mathbf{v}_j$ where $s_j=1$ or $-1$. Then, we obtain $\mathbf{q}^{opt}_j=s_jd_{jj}/(\Vert d_{jj} \Vert ^2_2+{\lambda}/{\Vert \mathbf{q}_j\Vert_2})\mathbf{v}_j$.

     Finally, we have
     \begin{equation}
     \label{HPSD on STPCA-MP}
         \begin{aligned}
             \mathbf{U}_{opt}\mathbf{Q}^H_{opt} &= \sum\limits^k_{j=1}\mathbf{u}_j^{opt}(\mathbf{q}_j^{opt})^H = \mathbf{V}\mathbf{\Sigma}\mathbf{V}^H\in H^q_+,
         \end{aligned}
     \end{equation}
     where$\sigma=d_{jj}/(\Vert d_{jj} \Vert ^2_2+{\lambda}/{\Vert \mathbf{q}_j\Vert_2})\geq0$. In each iteration, $\mathbf{U}_{opt}\mathbf{Q}^H_{opt}$ satisfies (\ref{HPSD on STPCA-MP}). When Problem (\ref{simplified problem}) is solved, (\ref{HPSD on STPCA-MP}) still holds. 

     Eventually, after each subproblem of (\ref{STPCA-MP: transform domain}) is solved, the optimal solution satisfies $\hat{\mathcal{A}}^{opt}_{::i}=\hat{\mathcal{U}}^{opt}_{::i}(\hat{\mathcal{Q}}^{opt}_{::i})^H\in H^q_+$.
\end{pf3}

\section{Reproducibility Statement}
To facilitate the reproducibility of our research, we provide the following resources:

\subsection*{Code Implementation}
\begin{itemize}
    \item \textbf{Code Repository}: Available at \url{https://github.com/zjj20212035/STPCA.git} under the \texttt{main} branch (\texttt{git checkout v1.1}).
    \item \textbf{Dependencies}: 
    \begin{itemize}
        \item MATLAB R2023b
        \item Tensor Toolbox v3.5 (included in \texttt{utils/})
    \end{itemize}
    \item \textbf{Parallel Computing}: All experiments utilize MATLAB's Parallel Computing Toolbox.
\end{itemize}

\subsection*{Datasets}
\begin{itemize}
    \item All datasets are provided in \texttt{data/} folder as \texttt{.mat} files, each is unified to contain the following common attributes:
     \begin{itemize}
        \item \texttt{data}: $d_1 \times d_2 \times n$ tensor ($n$=samples, $d_i$=$i$-mode dimensionality)
        \item \texttt{label}: $n \times 1$ class vector
        \item \texttt{tensor\_type}: \texttt{tube-wise} or \texttt{slice-wise}
        \item \texttt{tensor\_size}: dimensionality of the feature space
    \end{itemize}
    \item No manual data preprocessing needed - all preprocessing is handled automatically by:
    \begin{itemize}
        \item \texttt{config\_Table7.m}
        \item \texttt{config\_Figure7.m}
    \end{itemize}
where each dataset shares the same normalization that is carried out by dividing each element by the maximum absolute value, scaling all values to the interval $[-1,1]$.
\end{itemize}

\subsection*{Experiment Reproduction}
We have streamlined the code to allow one-click reproduction of the paper's key results. Open the project in MATLAB and
\begin{itemize}
    \item To replicate Table~\ref{Metrics of the comparative methods on real-world data}:
    \begin{verbatim}
    cd code
    run('exp_Table7_clustering_POC.m')
    \end{verbatim}
    \item To replicate Figure~\ref{Visualization}:
    \begin{verbatim}
    cd code
    run('exp_Figure7_visualization.m')
    \end{verbatim}
\end{itemize}

Hyperparameters are specified in \texttt{config\_Table7.m} and \texttt{config\_Figure7.m}, where: 
\begin{itemize}
\item Regularization parameters $\lambda$ and $\eta$ are tuned by grid search in the range of $\{10^{-2},10^{-1},1,10,10^2\}$,
\item Orientation is tuned from $D^o=\{\{1,3,2\}, \{2,3,1\}\}$ and the best result is shown,
\item The stabilization constants are fixed at $\epsilon_1 = 10^{-3}$ and $\epsilon_2 = 10^{-2}$ throughout all experiments.
\end{itemize}

\printcredits







\end{document}